\documentclass{article}
\usepackage{arxiv}
\usepackage[utf8]{inputenc} % allow utf-8 input
\usepackage[T1]{fontenc}    % use 8-bit T1 fonts
\usepackage{hyperref}       % hyperlinks
\usepackage{url}            % simple URL typesetting
\usepackage{booktabs}       % professional-quality tables
\usepackage{amsfonts}       % blackboard math symbols
\usepackage{nicefrac}       % compact symbols for 1/2, etc.
\usepackage{microtype}      % microtypography
\usepackage{lipsum}
\usepackage{amsmath,amsthm,amsfonts,amssymb,amsbsy,bm,mathtools,textcomp}
\usepackage{graphicx}
\usepackage[rflt]{floatflt}
\usepackage{hyperref}
\usepackage{xcolor}
\usepackage{subcaption}
\usepackage{adjustbox}
\usepackage{tabularx}
\usepackage{array}
\usepackage{siunitx}
\usepackage{comment}
\usepackage[numbers]{natbib}
\usepackage[textsize=tiny]{todonotes}
\usepackage[listings]{tcolorbox}
\usepackage{url}
\usepackage{breqn}
\usepackage[normalem]{ulem}
\usepackage{wrapfig}
\usepackage{comment}
\usepackage[final]{pdfpages}  % Add pdfs
\usepackage{xcolor}
\usepackage{soul}
\usepackage{tabularx}
\usepackage{enumitem}
\usepackage{multirow}
\usepackage{lipsum}
\usepackage{algorithm}
\usepackage{algpseudocode}
\newcommand{\norm}[1]{\left\lVert#1\right\rVert}

\title{Mathematics of Digital Twins and Transfer Learning for PDE Models}

\author{
Yifei Zong \\
Civil Engineering Department\
  University of Illinois Urbana Champaign\\
  Urbana, IL 61801 \\
  \texttt{yzong6@illinois.edu} \\
   \And
  Alexandre Tartakovsky \\
Civil Engineering Department\
  University of Illinois Urbana Champaign\\
  Urbana, IL 61801 \\
  Pacific Northwest National Laboratory,\\
  Richland, WA 99352 \\
  \texttt{amt1998@illinois.edu} \\
}

\begin{document}
\maketitle
\begin{abstract}

We define a digital twin (DT) of a physical system governed by partial differential equations (PDEs) as a model for real-time simulations and control of the system behavior under changing conditions. 
We construct DTs using the Karhunen–Loève Neural Network (KL-NN) surrogate model and transfer learning (TL). The surrogate model allows fast inference and differentiability with respect to control parameters for control and optimization. TL is used to retrain the model for new conditions with minimal additional data. 
We employ the moment equations to analyze TL and  
identify parameters that can be transferred to new conditions. The proposed analysis also guides the control variable selection in DT to facilitate efficient TL. 

For linear PDE problems, the non-transferable parameters in the KL-NN surrogate model can be exactly estimated from a single solution of the PDE corresponding to the mean values of the control variables under new target conditions. Retraining an ML model with a single solution sample is known as one-shot learning, and our analysis shows that the one-shot TL is exact for linear PDEs. 

For nonlinear PDE problems, transferring of any parameters introduces errors. %all components of the KL-NN model change and, in general, must be updated for the target conditions. 
For a nonlinear diffusion PDE model, we find that for a relatively small range of control variables, some surrogate model parameters can be transferred without introducing a significant error, some can be approximately estimated from the mean-field equation, and the rest can be found using a linear residual least square problem or an ordinary linear least square problem if a small labeled dataset for new conditions is available. The former approach results in a one-shot TL while the latter approach is an example of a few-shot TL. Both methods are approximate for the nonlinear PDEs. 

\end{abstract}


\section{Introduction}

Digital twins (DTs) \cite{wagg2020digital} for systems governed by partial differential equations (PDEs) is an active area of research, driven by improvements in computational power, advances in machine learning (ML), and data assimilation \cite{kaur2020convergence}. Examples include DTs for structural damage detection combining a physics-based model and ML algorithms \cite{ritto2021digital}, aircraft response based on  reduced-order models (ROMs)  \cite{kapteyn2020toward}, managing alpine basins \cite{morlot2024hydrological}, and flood discharge predictions based on deep neural networks (DNNs) \cite{alperen2021hydrological}. 
The DT definition varies based on the intended applications of DTs. We define a digital twin as a model that can describe the system in near real-time under all relevant conditions and can be used to control and optimize the system behavior. Many applications of DT do not require real-time control of the system, and instead focus on system optimization and design. In such applications, the ``near real-time'' requirement is replaced with a requirement that the DT prediction time is significantly less than the computational time of standard numerical models, and one can use ROMs as a core of DT. In this work, we consider DTs based on surrogate models, which allow near-real-time predictions but could be less accurate than ROMs, especially outside the parameters' training range. Surrogate models approximate the relation between variables describing the system properties, external conditions, and system responses. Because system control is one of the main applications of DTs, we will refer to this set of variables as control variables. Examples of ML surrogate models are DeepONets \cite{lu2021learning}, Fourier neural operators \cite{li2020fourier}, latent space deep neural network (DNN) models based on Karhunen–Loève expansion (KLE) \cite{wang2024bayesian} and principal component analysis (PCA) \cite{bhattacharya2020model}, and convolutional neural network (CNN)-based models \cite{tang2020deep}. 

ML surrogate models are trained using labeled datasets. However, obtaining a dataset that captures the complex system behaviors under all relevant conditions is often impractical. Instead, data is collected or synthetically generated under certain (\emph{source}) conditions. A surrogate model trained with such a dataset might not be able to accurately predict the system response under other (\emph{target}) conditions if these conditions are sufficiently different. Transfer learning (TL) methods aim to improve the model performance by retraining the model for the target problem with relatively little additional data. The idea of TL is to leverage knowledge gained from a source dataset to learn and adapt to system behavior under target conditions. 

Many ML surrogates use architectures similar to those in computer vision, where the model training for a source problem is performed using a (larger) source dataset. When applying this model to a target problem, a relatively small number of model parameters are retrained using the target dataset while the other parameters remain unchanged. The choice of \emph{frozen} and retrained parameters might greatly affect the TL performance. The frozen (transferrable) parameters are typically selected based on empirical observations.  For example,  convolutional layers and top layers in fully connected networks were found to be general to the problem, while low fully connected layers are task-specific \cite{yosinski2014transferable,liu2021deep}.
In \cite{liu2021deep}, a similar TL approach was used for the DeepONet model, where the parameters in convolutional layers of the branch network and some of the parameters in the fully connected layers of the trunk network were transferred from the source problem and the remaining parameters were retrained using the target dataset. This approach produced biased predictions for problems where target conditions were sufficiently different from the source conditions \cite{liu2021deep}. 

We propose an analysis of TL for DT to answer the following questions: 
Which surrogate model components and parameters can be transferred without incurring errors? When are the errors due to TL acceptable? What is the required dataset size to achieve accurate predictions under target conditions? What ML architecture is needed for a PDE in question? The answers to these questions are PDE-specific. In this work, we focus on linear and nonlinear diffusion equations.

The variability of control variables within a specified range is modeled using probability distributions. We employ the Karhunen–Loève neural network (KL-NN) surrogate model to construct DT.  The state and control variables are represented using KL expansions (KLEs) \cite{TartakovskyJCP2024}. The KLE of a stochastic process is expressed as the sum of its mean function and a series expansion, including the eigenvalues and eigenfunctions derived from its covariance function. A fully connected DNN or another regression model maps the KLE coefficients of the control variables to the KLE coefficients of the state variables. This approach renders the governing PDEs stochastic and allows using the moment equation method for TL analysis. The moment equation method provides the closed-form equations for the mean and covariance of the state variables and is used to analyze how the mean and covariance of the state variables change from the source to target conditions. Thus, it guides the transferability of the mean function, the eigenfunctions/covariance, and the parameters in the regression model.   

We find that the parameters in the KLEs and the neural network can be fully transferred for linear PDEs. The mean state function must be “re-learned,” which can be done by solving the PDE for the target conditions and the mean values of the control variables. For the nonlinear diffusion equation, we obtain conditions when the eigenfunctions can be transferred, and the other parameters can be retrained with a few samples without incurring a significant error.  Other ML methods, including multi-fidelity \cite{chakraborty2021transfer, liao2021multi}, multi-resolution \cite{donahue2014decaf, feng2019structural, iraniparast2023surface}, and few-shot learning \cite{qiao2018few, yu2020transmatch}, similarly transfer information from one dataset to another and can potentially benefit from the proposed analysis. 

This paper is organized as follows. In Section \ref{sec:surrogate_model}, we introduce the KL-NN surrogate model and transfer learning in this context as a digital twin for a PDE system. Section \ref{sec:transfer_learning} analyzes the TL properties for both linear and nonlinear diffusion PDEs. Examples are provided in Section \ref{sec:examples}, and conclusions are given in Section \ref{sec:conclusion}.

\section{Digital twins based on the KL-NN surrogate model} \label{sec:surrogate_model}

We are interested in DTs for systems governed by a PDE of the form:
\begin{equation}
\label{eq:component_model}
\mathcal{L}(h(x,t), y(x),f(x,t)) = 0, \quad x \in \Omega, \quad t\in T,
\end{equation}
subject to the initial condition
\begin{equation}
h(x,t=0)=h_0(x) , \quad x \in \Omega,
\end{equation}
and the boundary condition  
\begin{equation}\label{eq:BC}
\mathcal{B}(h(x,t))=g(x,t) \quad x\in \Gamma,
\end{equation}
where $\mathcal{L}$ is a known differential operator, $\Omega$ is the spatial domain, $T$ is the time domain, $h(x,t)$ is the state (solution) of the PDE, $y(x)$ is the property of the system, $f(x,t)$ is the source term,  $h_0(x)$ is the initial condition, $\mathcal{B}$ is the known boundary condition operator, $g(x,t)$ is the boundary function, and $\Gamma$ is the domain boundary. 

We define a complete DT of this system as a surrogate model
$
\hat{h}\left (x,t| y,f, g, h_0;\bm\zeta \right)
$
that (i) can be evaluated in near real-time (or, at least, significantly faster than a numerical solution of Eqs \eqref{eq:component_model}--\eqref{eq:BC}), 
(ii) can accurately predict $h$ for the specified ranges of $f$, $g$, $y$, and $h_0$, and 
(iii) can be differentiated with respect to $f$, $g$, $y$, and $h_0$ analytically or using automatic differentiation. The latter is important for using  $\hat{h}$ to control the system, estimate $f$, $g$, $y$, and $h_0$ from measurements, and assimilate data. In this surrogate model, $x$, $t$, $f(x,t)$, $g(x,t)$, $y(x)$, and $h_0(x)$ are inputs,  $\hat{h}(x,t)$ is the output, and $\bm\zeta$ is the collection of the surrogate model parameters. Because one of the main applications of DTs is system control, we refer to the model inputs as control variables.  

A surrogate model is usually trained using labeled datasets $D_{\text{train}} = \left \{ (f^{(i)},g^{(i)},y^{(i)},h_0^{(i)}) \rightarrow h^{(i)} \right \}_{i=1}^{N_{\text{train}}}$. One approach for generating a dataset is to treat the control variables $f$, $g$, $y$, and $h_0$ as random processes $F$, $G$, $Y$, and $H_0$ with the distributions describing the variability of the corresponding control variables. Then, the dataset can be generated as in the Monte Carlo Simulation (MCS) method for stochastic PDEs; $f^{(i)}$, $g^{(i)}$, $y^{(i)}$, and $h_0^{(i)}$ are obtained as realizations of the respective random processes, and the $h^{(i)}$ samples are computed by solving the PDE for the $i^{\text{th}}$ realization of the control variables.   

The number of samples $N_{\text{train}}$ required for training the surrogate model to the desired accuracy increases with the number of variables and their dimensionality. The latter refers to the number of parameters (latent variables) needed to represent the control and state variables. In turn, the problem's dimensionality increases with the ranges of the control variables. Therefore, training the DT for a wide range of control variables might be computationally infeasible. Instead, one can train the surrogate mode for the expected (and more narrow) range of the control variables under current conditions and retrain the surrogate model when the conditions and the range of control variables change. 

We assume that $\hat{h}(x,t | y,f, g, h_0;\bm\zeta^s)$ is trained (i.e., $\bm\zeta^s$ is estimated) for the control variables in the ranges defined by distributions with the mean values $\overline{f}^s(x,t)$, $\overline{g}^s(x,t)$, $\overline{y}^s(x)$, and $\overline{h}^s_0(x)$ using a sufficiently large dataset $D^s_{\text{train}}$. Here, the subscript $s$ denotes the \emph{source} condition. The dataset size $N_{\text{train}}^s$ is assumed to be sufficiently large for training the DT to the desired accuracy under the source condition. Then, we define TL as a problem of constructing the surrogate model $\hat{h}(x,t | y,f, g, h_0;\bm\zeta^t)$ for the range of the control variables with the mean values $\overline{f}^t(x,t)$, $\overline{g}^t(x,t)$, $\overline{y}^t(x)$, and $\overline{h}^t_0(x)$ using a significantly smaller dataset $D_{\text{train}}^t$. Here, the superscript $t$ denotes the \emph{target} condition. TL is an essential feature for a surrogate model to satisfy condition (ii) in the definition of DT. Therefore, we can define DT as a differentiable surrogate model plus TL.

Random functions can be represented with the infinite-term KLE and approximated by truncating the KLE \cite{loeve1978probability,xiu2010numerical,TartakovskyJCP2024}. For example, the infinite-dimensional KLE of a random process $H(x,t,\omega)$ has the form
\begin{align}
  \label{eq:space-time-KL-stoch}
  H(x,t,\omega) &=
  \overline{h}(x,t) + \sum_{i=1}^{\infty} \sqrt{\lambda^{i}_h }\phi^i_h(x, t) \tilde{\eta}^i(\omega) \nonumber \\
  &\approx \overline{h}(x,t) + \sum_{i=1}^{N_h} \sqrt{\lambda^{i}_h}\phi^i_h(x, t) \tilde{\eta}^i(\omega) \nonumber  \\
  & = \overline{h}(x,t) + \bm\psi_h(x, t) \cdot \tilde{\bm\eta},
\end{align}
where $\omega$ is the coordinate in the outcome space, $\overline{h}$ is the ensemble mean (expectation) of $H$,  $\lambda^i_h$ and $ \phi_h^i(x,t)$ are the eigenvalues and eigenfunctions of the $H$ covariance ($\lambda^i_h$ organized in the descending order), $\tilde{\eta}^i$ are the uncorrelated zero-mean random variables 
%(do not have to be Gaussian)
, $\tilde{\bm\eta} = \left[ \tilde{\eta}^1,\dots,\tilde{\eta}^{N_h} \right]^{\text{T}}$,
$\bm\psi_h = \left[ \sqrt{\lambda^1_h}\phi_h^1,\dots,\sqrt{\lambda^{N_h}_h}\phi_h^{N_h} \right]^{\text{T}}$,
and $N_h$ is selected according to the desired tolerance $rtol$ as
\begin{equation}
rtol = \frac{\sum_{i=N_h+1}^{\infty}\lambda^{i}_{h} }{\sum_{i=1}^{\infty}\lambda^{i}_{h}}.
\end{equation}

In the KL-NN method, the state variable $h(x,t)$ is treated as a realization of the stochastic process $H$ and is approximated with the truncated KL decomposition (KLD), which is a deterministic counterpart of the truncated KLE and is similar to the functional PCA: 
\begin{equation}
  \label{eq:space-time-KL_general}
  h(x,t) \approx \mathcal{KL} \left[ \overline{h},\bm\psi_h, \bm{\eta} \right] \coloneq \overline{h}(x,t) + \bm\psi_h(x, t) \cdot \bm\eta.
\end{equation}
Here, $\mathcal{KL}$ denotes the KLD operator and $\bm\eta = \left[ \eta^1,\dots,\eta^{N_h} \right]^{\text{T}}$ is the vector of KLD coefficients.
Given $h^*(x,t)$, $\bm\eta$ is defined by the inverse KLD operator, 
 \begin{align}
   \bm{\eta}^* = \mathcal{KL}^{-1} \left[ h^*;\overline{h},\bm\psi_h \right] \coloneq \arg\min_{\bm{\eta} } \norm{  h^*(x,t) - \mathcal{KL} \left[ \overline{h},\bm\psi_h, \bm{\eta} \right]  }_2^2 + \gamma || \bm\xi||_2^2 \},
\end{align}
where $\gamma\ge 0$ is the regularization coefficient. We find that $\gamma>0$ improves the TL accuracy when the correlation length and time of $h^*$ are smaller than those of $H$.

Similarly, the control variables are represented with KLDs as 
\begin{align}
f(x,t) &\approx  \mathcal{KL} \left[ \overline{f}, \bm\psi_f, \bm\xi^{i}_f \right] = \overline{f}(x,t) + \bm\psi_f(x, t) \cdot \bm\xi_f , \\
y(x) &\approx  \mathcal{KL} \left[ \overline{y}, \bm\psi_y, \bm\xi^{i}_y \right] = \overline{y}(x) + \bm\psi_y(x) \cdot \bm\xi_y , \\
h_0(x) &\approx  \mathcal{KL} \left[ \overline{h}_0, \bm\psi_{h_0}, \bm\xi^{i}_{h_0} \right] = \overline{h}_0(x) + \bm\psi_{h_0}(x) \cdot \bm\xi_{h_0} , \\
g(x,t) &\approx  \mathcal{KL} \left[ \overline{g}, \bm\psi_g, \bm\xi^{i}_g \right] = \overline{g}(x,t) + \bm\psi_g(x, t) \cdot \bm\xi_g , 
\end{align}
where $\overline{f}$, $\overline{y}$, $\overline{h}_0$, and $\overline{g}$ are the mean functions of the corresponding stochastic processes $F$, $Y$, $H_0$, and $G$, and $\bm\psi_f = \left[ \sqrt{\lambda^1_f}\phi_f^1,\dots,\sqrt{\lambda^{N_f}_f}\phi_f^{N_f} \right]^{\text{T}}$, $\bm\psi_y = \left[ \sqrt{\lambda^1_y}\phi_y^1,\dots,\sqrt{\lambda^{N_y}_y}\phi_y^{N_y} \right]^{\text{T}}$, $\bm\psi_{h_0} = \left[ \sqrt{\lambda^1_{h_0}}\phi_{h_0}^1,\dots,\sqrt{\lambda^{N_{h_0}}_{h_0}}\phi_{h_0}^{N_{h_0}} \right]^{\text{T}}$, and $\bm\psi_g = \left[ \sqrt{\lambda^1_g}\phi_g^1,\dots,\sqrt{\lambda^{N_g}_g}\phi_g^{N_g} \right]^{\text{T}}$
are the basis functions of the $F$, $Y$, $H_0$, and $G$ covariance functions, respectively. The vectors of KLD coefficients $\bm{\xi}_f$, $\bm{\xi}_y$, $\bm{\xi}_{h_0}$, and $\bm{\xi}_g$ are given by the $\mathcal{KL}^{-1}$ operators acting on the corresponding control variables. 

Then, the KL-NN surrogate model for the PDE problem \eqref{eq:component_model}-\eqref{eq:BC} is constructed in the reduced (latent) space of KLD coefficients as  
\begin{equation}\label{eq:least_sqaure_fitting}
\bm{\eta}(\bm{\xi}) \approx
\hat{\bm{\eta}}(\bm{\xi};\bm\theta) = \mathcal{NN}(\bm{\xi}; \bm{\theta}),  
\end{equation}
where $\bm\xi^{\text{T}} = \left[ \bm{\xi}^{\text{T}}_y, \bm{\xi}^{\text{T}}_f, \bm{\xi}^{\text{T}}_{h_0}, \bm{\xi}^{\text{T}}_g \right] \in \mathbb{R}^\xi$ is the input vector, $\hat{\bm\eta}$ is the output vector, $\mathcal{NN}$ denotes a neural network, and $\bm{\theta}$ is the collection of the neural network parameters. 

As a general model, we employ a fully connected DNN with $N$ hidden layers defined as:
\begin{equation}
    \mathcal{NN}(\bm\xi; \bm\theta) := \rho_{N+1}(\rho_{N}(\rho_{N-1}(...(\rho_1(\bm{\xi}))))),
\end{equation}
where
\begin{equation}
    \rho_{i}(\bm{\xi}_i) := \rho_i \odot (\bm{W}_{i}\bm{\xi}_{i} + \bm{b}_{i}),
\end{equation}
$\bm\theta = (\bm{W}_{1:N+1}, \bm{b}_{1:N+1})$ denote the collection of weight and biases starting from the first layer to the last, $\rho_i$ is the activation function, $\bm{\xi}_i = \rho_{i-1}(\bm{\xi}_{i-1}) \in \mathbb{R}^{n_i}$, $n_i$ is the dimensionality of the $i^{\text{th}}$  layer, $\bm{b}_i \in \mathbb{R}^{n_{i+1}}$, $\bm{W}_i \in \mathbb{R}^{n_{i+1} \times n_{i}}$,  $\odot$ represents the element-wise operation, $\bm{\xi}_1 = \bm{\xi}_k$, and $\hat{\bm\eta} = \rho_{N+1}(\bm\xi_{N+1})$. Here, we use the hyperbolic tangent activation function $\rho_i(\cdot)= \tanh(\cdot)$ for $i \in \{1, ..., N \}$ and the identity function for $i=N+1$, i.e., $\hat{\bm\eta} = \rho_{N+1}(\bm\xi_{N+1}) = \bm\xi_{N+1}$.
We also consider a linear regression model $\mathcal{NN}(\bm{\xi}; \bm{\theta}) = \bm{W}\bm\xi$ 
($\bm{W} \in \mathbb{R}^{N_h \times N_\xi}$) 
as a special case of neural networks, where $\bm\theta=\bm{W}$.

In general, $\bm\theta$ is found by minimizing the loss function  
\begin{align}\label{eq:surrogate_loss}
L(\bm\theta) = \sum_{i=1}^{N_{\text{train}}} \norm{ \mathcal{NN}(\bm{\xi}^{(i)};\bm\theta) - \boldsymbol{\eta}^{(i)} }^2_2 
\end{align}
using the reduced-space labeled dataset $\{\bm\xi^{(i)} \rightarrow \bm\eta^{(i)}  \}_{i=1}^{N_{\text{train}}}$, which is obtained by applying the inverse KLD operator to elements of the $D_{\text{train}}$ dataset.  

The mean and covariance of control variables are selected based on the prior knowledge. The mean and covariance of the state solution $h$ are given by those of $H(x,t,\omega)$, found by solving the stochastic PDE problem defined by Eqs \eqref{eq:component_model}--\eqref{eq:BC} with the control variables replaced by their stochastic counterparts. The moment equation method, generalized Polynomial Chaos, or the MCS method can be used for computing the mean and covariance of $h$. In this work, the MCS-generated $D_{\text{train}}$ is used to approximate the mean and covariance of $h$ as  
\begin{equation}\label{eq:ens_mean}
  \overline{h}(x,t)\approx \frac{1}{N_{\text{train}}} \sum_{i=1}^{N_{\text{train}}} h^{(i)}(x,t)
\end{equation}
and
\begin{equation}\label{eq:ens_cov}
  C_h(x, x'; t, t') \approx \frac{1}{N_{\text{train}}-1} \sum_{i=1}^{N_{\text{train}}} \left[ (h^{(i)}(x,t)-\overline{h}(x,t)) (h^{(i)}(x',t') -\overline{h}(x',t')) \right].
\end{equation}
The eigenvalues and eigenfunctions are found by solving the Fredholm integral equation of the first kind
\begin{equation}
  \label{eq:space-time-covar}
  \int_T \int_\Omega C_h(x, x'; t, t') \phi^i_h(x', t') \, \mathrm{d} x' \mathrm{d} t' = \lambda^i_h \phi^i_h(x, t), \quad x, x' \in \Omega,\quad t, t'\in T, \quad i = 1, \dots, N_h. 
\end{equation}

Substituting the reduced-space mapping  \eqref{eq:least_sqaure_fitting} into Eq.~\eqref{eq:space-time-KL_general} yields the complete form of the surrogate model:
\begin{equation}\label{eq:surrogate_model}
\hat{h}(x,t|f,y,h_0,g; \bm\theta) = \mathcal{KL} \left[ \overline{h},\bm\psi_h, \mathcal{NN}(\bm{\xi};\bm\theta) \right].
\end{equation}

We note that $\mathcal{NN}(\bm{\xi}; \bm\theta)$ is a known function of  $\bm{\xi}$ and $\bm\theta$. Therefore, depending on the form of $\mathcal{NN}$, the KL-NN model can be differentiated with respect to $\bm{\xi}$ and $\bm\theta$ analytically or using automatic differentiation, i.e., the KL-NN model is a \emph{differentiable} model \cite{Shen2023NatureR}.

In most cases, the dataset $D_{\text{train}}$ is computed numerically on a $N_m = N_t \times N_x$ time-space mesh, where
$N_t$ is the number of time steps, and $N_x$ is the number of the grid points or elements. Then, $\overline{h}(x,t)$, $C_h(x,x',t,t')$, and $\bm\psi_h(x,t)$ eigenfunctions of the state and control variables must also be computed on the same mesh. Then, it is convenient to define discrete forward and inverse KLD transforms, e.g., 
\begin{equation}\label{eq:surrogate_model_discrete}
\bm{h}(\bm{f},\bm{y},\bm{h}_0,\bm{g}) \approx \mathcal{KL} \left[ \overline{\bm{h}}, \bm\Psi_h, \mathcal{NN}(\bm{\xi};\bm\theta) \right],
\end{equation}
where ${\bm{h}}\in \mathbb{R}^{N_m}$, $\overline{\bm{h}}\in \mathbb{R}^{N_m}$, and ${\bm{\Psi}_h}\in \mathbb{R}^{N_m \times N_h}$. 

We define the relative surrogate model error $\varepsilon$ as, 
\begin{equation}\label{eq:total_error}
  \varepsilon = \frac{ \norm{ \bm{h} - \mathcal{KL} \left[ \overline{\bm{h}},\bm\Psi_h, \mathcal{NN}(\bm{\xi}^*; \bm{\theta}) \right] }_2}{ \norm { \bm{h} }_2}.  
\end{equation}
The $\varepsilon$ error is a combination of the representation errors $\varepsilon_{KL}$ (the accuracy of KLD approximations of the control and state variables) and the $\mathcal{NN}$ mapping error $\varepsilon_{NN}$ (the accuracy of the $\mathcal{NN}(\bm\xi;\bm\theta)$ approximation of the $\bm\eta$ versus $\bm\xi$ relationship). The relative mapping error can be defined as 
\begin{equation}\label{eq:NN_error}
\varepsilon_{NN} = \frac{ \norm{ \bm{\eta}(\bm{\xi}) - \mathcal{NN}(\bm{\xi}; \bm{\theta})}_2}{ \norm{ \bm{\eta}(\bm{\xi}) }_2}.
\end{equation}
The relative representation error is defined as
\begin{equation}\label{eq:kl_error}
\varepsilon_{KL} = \frac{
\norm{ \bm{h} - \mathcal{KL}\left[\overline{\bm{h}},\bm\Psi_h, 
 \mathcal{KL}^{-1} \{\bm{h};\overline{\bm{h}}, \bm\Psi_h \}
\right] }_2
}
{\norm{ \bm{h}}_2}.
\end{equation}

\section{Analysis of transfer learning} \label{sec:transfer_learning}

In this section, we investigate the relationship between the latent control and state variables, and the transferability of the KL-NN surrogate model components learned under source conditions $\overline{f}^s(x,t)$, $\overline{g}^s(x,t)$, $\overline{y}^s(x)$, and $\overline{h}^s_0(x)$ to a model for the target conditions $\overline{f}^t(x,t)$, $\overline{g}^t(x,t)$, $\overline{y}^t(x)$, and $\overline{h}^t_0(x)$.  
In the following analysis, we obtain the equations for $\overline{h}(x,t)$, $h'(x,t,\omega)$ (zero-mean fluctuation of $H(x,t,\omega)$), and $ C_h(x, x'; t, t') = \overline{h'(x,t,\omega)h'(x',t',\omega)}$ using the moment equation method by treating the control and state variable in the PDE problem \eqref{eq:component_model}-\eqref{eq:BC} as random variables and using the Reynolds decomposition of random variables: $H(x,t,\omega) = \overline{h}(x,t)+h'(x,t,\omega)$, $F(x,t,\omega) = \overline{f}(x,t)+f'(x,t,\omega)$, $G(x,t,\omega) = \overline{g}(x,t)+g'(x,t,\omega)$, $y(x,\omega) = \overline{y}(x)+y'(x,\omega)$, and $H_0(x,\omega) = \overline{h}_0(x)+h'_0(x,\omega)$, where $f'$, $g'$, $y'$, and $h'_0$ are zero-mean fluctuations. Then, we study the transferability of KL-NN components by analyzing the dependence of these equations on the source and target conditions. 
In the following, we omit the outcome space coordinate $\omega$ to simplify the notation. 

\begin{figure}
    \centering
    \includegraphics[width=0.95\linewidth]{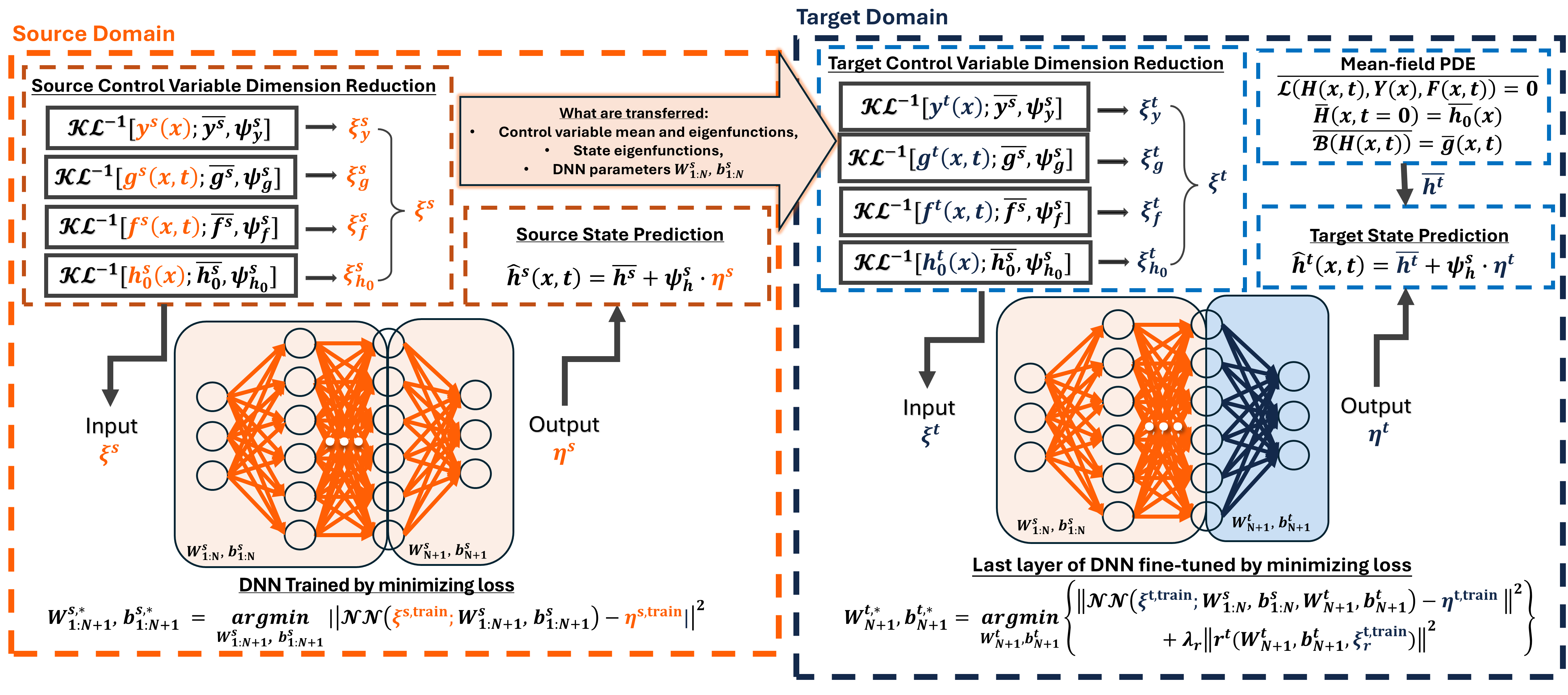}
    \caption{ Schematic description of TL for the KL-NN surrogate of the PDE \eqref{eq:component_model}-\eqref{eq:BC} model: 
    KL-NN is initially trained using a sufficiently large source dataset (the orange box). The control and state variables are represented with the truncated KLDs whose means and covariances describe the range of the variables under source conditions. The DNN parameters $\bm\theta^s = (\bm{W}_{1:N+1}^s,\bm{b}_{1:N+1}^s)$ are computed from the source training dataset based on Eq \eqref{eq:surrogate_loss}. In the inference step, the DNN inputs $\bm \xi^s$ are obtained with the inverse KLD operators acting on control variables. The DNN output $\bm \eta^s$ is transformed into the state variable using the (forward) KLD. 
    The KL-NN model is retrained for inference under target conditions 
    (blue box) 
    by transferring  the means and eigenfunctions of the source control variables and the state variable eigenfunctions. For nonlinear PDEs, the parameters of the last layer of the DNN, $\bm{W}_{N+1}^t$ and $\bm{b}_{N+1}^t$, are retrained using Eq. \eqref{eq:pi_loss_target}.  For linear PDEs, $\bm{W}_{1:N}^s=0$, $\bm{b}_{1:N+1}^s =0$, and $\bm{W}_{N+1}^t = \bm{W}_{N+1}^s$, i.e., the NN mapping is linear and its parameters are transferable. The target state is predicted as $\hat{\bm h}^t \approx \mathcal{KL}[\overline{\bm{h}}^t, \bm \Psi^s, \bm\eta^t]$ where $\overline{\bm{h}}^t$ is computed from the mean-filed equation subject to the target IBCs. 
    In summary, the proposed TL-KL-NN approach enables "one-shot" learning for linear PDEs and "few-shot" learning for nonlinear PDEs.
    }
    \label{fig:tl_schematics}
\end{figure}

First, we take the ensemble mean of Eqs. \eqref{eq:component_model}-
\eqref{eq:BC} as
\begin{equation}
\label{eq:component_model_mean}
\overline{\mathcal{L}(H(x,t), Y(x),F(x,t))} = 0, \quad x \in \Omega, \quad t\in T,
\end{equation}
subject to the initial condition
\begin{equation}
\overline{H}(x,t=0)= \overline{h}_0(x)
\end{equation}
and the boundary conditions  
\begin{equation}\label{eq:BC_mean}
\overline{\mathcal{B}(H(x,t))}= \overline{g}(x,t), \quad x \in \Gamma.
\end{equation}
The equation for fluctuations $h'(x,t)$ can be derived by subtracting Eqs \eqref{eq:component_model_mean}-\eqref{eq:BC_mean} from Eqs \eqref{eq:component_model}-\eqref{eq:BC}. The covariance equation is obtained by multiplying both sides of the fluctuation equation with $h'(x',t')$ and taking the ensemble average. 
The resulting TL framework for the PDE problem \eqref{eq:component_model}-\eqref{eq:BC} is shown in Figure \ref{fig:tl_schematics}. 
Next, we consider the moment equations and TL analysis for linear and nonlinear PDE problems.

\subsection{Linear PDE problems}\label{sec:linear_TL}
In the context of TL, we define a linear PDE problem, or, equivalently, a linear $\mathcal{L}$ operator acting on a combination of the control and state variables, as 
$\mathcal{L}({h}(x,t), {y}(x),{f}(x,t)) = \mathcal{L}_1({h}(x,t))+ \mathcal{L}_2( {y}(x)) + \mathcal{L}_3 ({f}(x,t))$, where $\mathcal{L}_1$,  $\mathcal{L}_2$, and  $\mathcal{L}_3$
are linear operators. 
Then, for linear operators $\mathcal{L}$ and $\mathcal{B}$, Eqs \eqref{eq:component_model_mean}-\eqref{eq:BC_mean} reduce to 
\begin{equation}
\label{eq:component_model_mean_linear}
\mathcal{L}(\overline{h}(x,t), \overline{y}(x),\overline{f}(x,t)) = 0, \quad x \in \Omega, \quad t\in T,
\end{equation}
subject to the initial condition
\begin{equation}
\overline{h}(x,t=0) = \overline{h}_0(x)
\end{equation}
and the boundary conditions  
\begin{equation}\label{eq:BC_mean_linear}
\mathcal{B}(\overline{h}(x,t)) = \overline{g}(x,t) , \quad x \in \Gamma.
\end{equation}
The fluctuation equation is obtained by subtracting Eqs \eqref{eq:component_model_mean_linear}-\eqref{eq:BC_mean_linear} from Eqs \eqref{eq:component_model}-\eqref{eq:BC} 
as
\begin{equation}
\label{eq:component_model_dev_linear}
\mathcal{L}(h'(x,t), y'(x),f'(x,t)) = 0, \quad x \in \Omega, \quad t\in T,
\end{equation}
subject to the initial condition
\begin{equation}
h'(x,t=0)= h'_0(x)
\end{equation}
and the boundary conditions  
\begin{equation}\label{eq:BC_dev_linear}
\mathcal{B}(h'(x,t)) = g'(x,t), \quad x \in \Gamma.
\end{equation}

The equation and IBCs for the covariance $C_h(x,x',t,t') \equiv \overline{h'(x,t) h'(x',t') } $ are obtained by multiplying Eqs \eqref{eq:component_model_dev_linear}-\eqref{eq:BC_dev_linear} with $h'(x',t')$ and taking the ensemble mean:
\begin{equation}
\label{eq:component_model_cov_linear}
\mathcal{L}(C_h(x,x',t,t'), C_{yh}(x,x',t'),C_{fh}(x,x',t,t')) = 0, \quad x \in \Omega, \quad t\in T,
\end{equation}
subject to the initial condition
\begin{equation}
C_h(x,x',t=0,t') = \overline{h'_0(x) h'(x',t')}
\end{equation}
and the boundary conditions  
\begin{equation}\label{eq:BC_cov_linear}
\mathcal{B}(C_h(x,x',t,t'))= C_{gh}(x,x',t,t'))   \quad x \in\Gamma.
\end{equation}
The equations for cross-covariances  $C_{yh}(x,x',t') \equiv \overline{y'(x) h'(x',t') } $, $C_{fh}(x,x',t,t') \equiv \overline{f'(x,t) h'(x',t') }$, and $C_{gh}(x,x',t,t') \equiv \overline{g'(x,t) h'(x',t') }$ are given in Appendix \ref{sec:crosscov}. These equations depend on the covariances $C_y$, $C_f$, and $C_g$, which stay the same under the source and target conditions in this analysis. In the numerical examples in Section \ref{sec:linear_diff_example}, we consider the effect of changing $C_y$, $C_f$, and $C_g$ on TL.

It is important to note that the equations for $C_h$ and $h'(x,t,\omega)$ are independent of the mean functions 
and only depend on the covariances of the control variables (which are assumed to be known). In turn, $\overline{h}$ is independent of the fluctuations 
%$h'$, $y'$, $f'$, and $g'$ 
and only depends on the mean values of the control variables. This leads to the following conclusions about TL for linear PDE problems:
\begin{enumerate}
    \item Exact ``one-shot” learning for $\overline{h}^t(x,t)$: From Eqs \eqref{eq:component_model_mean_linear}-\eqref{eq:BC_mean_linear}, we find that    
    $\overline{h}^s(x,t) \ne \overline{h}^t(x,t)$ and that $\overline{h}^t(x,t)$ can be obtained by solving Eqs. \eqref{eq:component_model_mean_linear}-\eqref{eq:BC_mean_linear} with $\overline{f}(x,t) = \overline{f}^t(x,t)$, $\overline{g}(x,t)=\overline{g}^t(x,t)$, $\overline{y}(x) = \overline{y}^t(x)$, and $\overline{h}_0(x) = \overline{h}^t_0(x)$, which amounts to collecting a single sample under the target conditions. 

 \item Exact transfer of eigenfunctions and eigenvalues $\left \{ \lambda^i_h, \phi_h^i(x,t) \right \}_{i = 1}^{N_h}$: Eqs \eqref{eq:component_model_cov_linear}-\eqref{eq:BC_cov_linear} indicate that $C_h$ is independent of the source and target conditions. Therefore, its eigenfunction and eigenvalues can be fully transferred from the source to the target problem. 
 
    \item Exact transfer of the NN parameters $\bm\theta$: Eqs \eqref{eq:component_model_dev_linear}-\eqref{eq:BC_dev_linear} imply that the relationship between $h'(x,t,\omega)$ and the fluctuations in the control variables is independent of the source and target conditions. From (2), the eigenvalues and eigenvalues of $h$ do not depend on the source and target conditions. Therefore,  the relationship between $\bm\eta$ and $\bm\xi$ is also independent of the source and target conditions, and the parameter $\theta$ in the neural network model in \eqref{eq:surrogate_model} can be completely transferred from the source to target problem.
 
\end{enumerate}
This process amounts to ``one-shot” learning for the entire KL-NN surrogate model for a linear PDE system. 
Next, we obtain the relationship between $\bm\eta$ and $\bm\xi$ by formulating the residual least-square (RLS) problem as: 
\begin{align}\label{eq:RLS}
\bm\eta^*  &= \min_{\bm\eta} L_{\text{RLS}}(\bm\eta,\bm\xi), \\
&= \min_{\bm\eta} \left [ \omega_r \left \| \bm{r} (\bm{\eta},\bm\xi_y,\bm\xi_f) \right \|_2^2 
+ \omega_b \left \| \bm{r}_b(\bm\eta, \bm{\xi}_g) \right \|_2^2 + 
\omega_0 \left \| \bm{r}_0(\bm\eta, \bm{\xi}_{h_0}) \right \|_2^2 \right ], \nonumber
\end{align}
where $\bm r \in \mathbb{R}^{N_m \times N_\eta}$ is the residual vector of Eq \eqref{eq:component_model_dev_linear}
evaluated on the $N_m=N_t \times N_x$ time-space grid:
\begin{equation}
  \label{eq:residual}
  \bm{r} (\bm\eta,\bm\xi_y,\bm\xi_f ) = \mathcal{L} \left[
  \hat{h}(\bm\eta), \hat{y}(\bm{\xi}_y), \hat{f}( \bm{\xi}_f) \right],
  \end{equation}
and the boundary condition residual vector is:
 \begin{equation} 
\bm{r}_b(\bm\eta, \bm\xi_g) = \mathcal{B}\left[ \hat{h}(\bm{\eta}) \right] - \hat{g}( \bm{\xi}_g) , 
\end{equation}
and the initial condition residual vector is:
\begin{equation}
   \bm{r}_0(\bm\eta,\bm\xi_{h_0}) = \hat{h}(x, t = 0; \bm{\eta}) - \hat{h}_0( \bm\xi_{h_0}) .
\end{equation}
A similar RLS problem was considered in the physics-informed KLE method for forward and inverse PDE problems \cite{tartakovsky2021physics,TartakovskyJCP2024}. 
For linear $\mathcal{L}$ and $\mathcal{B}$, the RLS problem can be rewritten as
\begin{equation}\label{eq:res_vector}
\bm\eta^* = \min_{\bm\eta} || \bm{A}\bm\eta - \bm{B}\bm\xi ||_2^2,
\end{equation}
where the matrices $\bm{A}$ and $\bm{B}$ depend on the $\mathcal{L}$ and $\mathcal{B}$ operators and are independent of $\bm\xi$ and $\bm\eta$. For the linear diffusion equation considered in Section \ref{sec:linear_diff_example}, $\bm A$ and $\bm B$ are given in Appendix \ref{sec:linear_diff}. 
The solution to this minimization problem has the form
\begin{equation}\label{eq:W_RLS}  
 \bm\eta^*  = (  \bm{A}^{\text{T}}  \bm{A} )^{-1}  \bm{A}^{\text{T}} \bm{B} \bm\xi = \bm{W} \bm\xi,
\end{equation}
i.e., $\bm\eta$ is a linear function of $\bm\xi$. In the following, we refer to the method of estimating $\bm{W}$ from Eq \eqref{eq:W_RLS} as the RLS method. 

Alternatively, $\bm{W}$ can be estimated by minimizing the loss function \eqref{eq:surrogate_loss}. For the linear model $\mathcal{NN}(\bm\xi,\bm{W}) = \bm{W} \bm\xi$, this results in the following minimization problem:
\begin{equation}\label{eq:regression_est}
    \bm{W}^* = \min_{\bm{W}} \sum_{i=1}^{N_{\text{train}}}
    ||\bm\eta^{(i)} - \bm{W} \bm\xi^{(i)} ||_2^2 = \min_{\bm{W}} || \bm H - \bm W \bm\Xi||_F^2,
\end{equation}   
where $\bm\Xi =  \left[ \bm\xi^{(1)}| \bm\xi^{(2)} |\dots| \bm\xi^{(N_{\text{train}})} \right] \in \mathbb{R}^{N_\xi \times N_{\text{train}}}$, $\bm H = \left[ \eta^{(1)}| \eta^{(2)}| \dots, \eta^{(N_{\text{train}})} \right] \in \mathbb{R}^{N_\eta \times N_{\text{train}}}$. The solution to this problem is (Appendix \ref{sec:OLS}):
\begin{equation}\label{eq:OLS_solution}
   \bm{W}^*  =   \bm H \bm \Xi^T (\bm \Xi\bm \Xi^T)^{-1}.
\end{equation} 
If $N_{\text{train}}$ is on the order of or significantly less than $N_\xi$, the total number of KLE dimensions needed to represent all control parameters, then the $l^2$ regularization must be added, yielding:
\begin{equation}\label{eq:regression_l2}
    \bm{W}^{\text{reg}} = \min_{\bm{W}} \{ || \bm H - \bm W \bm\Xi||_F^2 + \lambda || \bm W  ||_F^2 \} = \min_{\bm{W}} || \bm \tilde{H} - \bm W \bm \tilde{\Xi}||_F^2, 
\end{equation}   
where $\lambda$ is the regularization coefficient, $\bm\tilde{\Xi} = \left[ \bm \Xi | \sqrt{\lambda}\bm I \right] \in \mathbb{R}^{N_\xi \times (N_{\text{train}} + N_\xi)}$, and $\bm \tilde{H} = \left[ \bm H | \bm 0 \right] \in \mathbb{R}^{N_\eta \times (N_{\text{train}} + N_\xi)}$.  We refer to Eqs \eqref{eq:OLS_solution} and \eqref{eq:regression_l2} as the ordinary least-square (OLS) method of computing $\bm W$. 

\subsection{Transfer learning for a nonlinear PDE}\label{sec:nonlinear_TL}
For linear PDE problems, the moment equations for $\overline{h}$, $h'$, and $C_h$  are decoupled. However, the moment equations are coupled for nonlinear PDE problems that might affect the transferability of the parameters in the KL-NN model. As an example, we present a TL analysis for the one-dimensional diffusion equation with the space-dependent control variable $k(x)$: 
\begin{equation}\label{eq:nonlinear_pde} 
\frac {\partial h}{\partial t} =  \frac{\partial}{\partial x} \left[ k(x) \frac{\partial h}{\partial x} \right], \quad x \in [0, L], \quad t \in  [0,T]
\end{equation} 
subject to the initial condition 
\begin{equation} 
h(x,t=0)= h_0, \quad x \in [0, L],
\end{equation} 
and the boundary conditions 
\begin{equation}\label{eq:nonlinear_pde_bc}  
h(x=0,t)= h_l, \quad h(x=L,t)= h_r , \quad t \in [0,T]. 
\end{equation}
This equation, among other applications, describes the flow in a confined aquifer with a heterogeneous conductivity field.  We model the control variable $k(x)$ with the KLD: 
\begin{equation}
    k(x) \approx  %\overline{k}(x) + \sum_i^{N_k} \sqrt{\lambda^i_K} \phi^i_k(x) \xi^i_k = 
    \mathcal{KL} \left[\overline{k}, \bm\psi_k, \bm\xi_k \right], \nonumber
\end{equation}
and the surrogate model takes the form
\begin{equation} 
\hat{h}(x,t| k ; \bm{\theta}) =\mathcal{KL} \left[\overline{h}, \bm\psi_h, \mathcal{NN}(\bm\xi_k;\bm\theta) \right]. \nonumber
\end{equation}

To simplify the analysis, we further assume that the initial condition $h_0$ and boundary conditions $h_l$ and $h_r$ are deterministic and different for the source and target problems. 

We now employ the moment equation method for the PDE problem \eqref{eq:nonlinear_pde}--\eqref{eq:nonlinear_pde_bc}. We treat $k(x)$ and $h(x,t)$ as random fields $K(x)$ and $H(x,t)$  decomposed as $K(x) = \overline{k}(x) + k'(x)$ and $H(x,t) = \overline{h}(x,t) + h'(x,t)$. Then, taking the ensemble mean of the stochastic PDE yields 
\begin{equation}\label{eq:nonlinear_mean}
 \frac {\partial \overline{h}}{\partial t}  =  \frac{\partial}{\partial x} \left[ \overline{k}(x) \frac{\partial \overline{h}}{\partial x} \right] + \overline{ \frac{\partial}{\partial x} \left[ k'(x) \frac{\partial h'}{\partial x} \right] },
\end{equation}
subject to the initial condition
\begin{equation}\label{eq:non_linear_IC_mean}
\overline{h}(x,t=0)= h_0
\end{equation}
and the boundary conditions  
\begin{equation}\label{eq:nonlinear_bc_mean}
\overline{h}(x=0,t)= h_l, \quad \overline{h}(x=L,t)= h_r. 
\end{equation}
The equation for fluctuations $h'$ is obtained by subtracting Eq \eqref{eq:nonlinear_mean}--\eqref{eq:nonlinear_bc_mean} from \eqref{eq:nonlinear_pde}--\eqref{eq:nonlinear_pde_bc}: 
\begin{equation}\label{eq:nonlinear_fluctuations}
\frac {\partial h'}{\partial t} =  \frac{\partial}{\partial x} \left[ k'(x) \frac{\partial h'}{\partial x} \right]
+
\frac{\partial}{\partial x} \left[ \overline{k} (x) \frac{\partial h'}{\partial x} \right]
+
\frac{\partial}{\partial x} \left[ k'(x) \frac{\partial \overline{h}}{\partial x} \right]
 - \overline{ \frac{\partial}{\partial x} \left[ k'(x) \frac{\partial h'}{\partial x} \right] },  
\end{equation}
subject to the initial condition
\begin{equation}\label{eq:fluc_IC}
h'(x,t=0) =  0
\end{equation}
and the boundary conditions  
\begin{equation}\label{eq:nonlinear_bc_fluctuations}
h'(x=0,t)=  0, \quad h'(x=L,t)=  0. 
\end{equation}
The equation for the covariance $C_h$ can be found in \cite{Yeung2022WRR}.
Substituting the KLEs of $k'$ and $h'$ in Eq \eqref{eq:nonlinear_fluctuations} reveals that (i) $\bm\eta$ is a nonlinear function of $\bm\xi_k$, (ii) $\bm\eta$ depends on $\overline{h}$ and, therefore, on the IBC $h_0$, $h_l$, and $h_r$, and (iii) the parameters $\bm\theta$ in the $\hat{\bm\eta} = \mathcal{NN}(\bm\xi_k;\bm\theta)$ model are different for the source and target IBCs and cannot be fully transferred.     
For the same reason, the covariance $C_h$ and, therefore, the eigenfunctions and eigenvalues in the KLD of $h$ cannot be fully transferred.  Furthermore, the equations for $\overline{h}$, $h'$, and $C_h$ contain non-local terms such as  $\overline{ \frac{\partial}{\partial x} \left[ k'(x) \frac{\partial h'}{\partial x} \right] }$. Evaluation of these terms requires computationally expensive closures or MCS under both the source and target conditions. 

The equations for $\overline{h}$ and $h'$ can be simplified if the variance of $y=\ln K$ is less than 1, $\sigma^2_y < 1$ (or, $\sigma_k/\overline{k} < 1$, where $\sigma_k$ is the standard deviation of $K$). In many applications, $k(x)$ represents a system property, such as the hydraulic conductivity of porous media, which is typically determined by solving an inverse PDE problem. Inverse problems are ill-posed and their solutions are subject to uncertainty characterized by the \emph{posterior} variance, e.g., $\sigma^2_y$. The magnitude of the posterior variance depends on the number of the system state and parameter measurements and the quality of the prior assumptions. 

From the moment equation perturbation analysis \cite{Yeung2022WRR}, it follows that $\overline{h}$ is of the zero-order in  $k'$ (i.e., $\overline{h}$ can be expanded in a series using terms of zero power of $k'$ and higher) and the mixed terms are of the second order in $k'$. The IBCs for Eq \eqref{eq:nonlinear_mean} are deterministic and independent of $k'$. Therefore, for $\sigma_y^2<1$, Eq \eqref{eq:nonlinear_mean} can be approximated as 
\begin{equation}\label{eq:nonlinear_mean_sim}
 \frac {\partial \overline{h}}{\partial t}  =  \frac{\partial}{\partial x} \left[ \overline{k}(x) \frac{\partial \overline{h}}{\partial x} \right], 
\end{equation}
subject to the IBCs \eqref{eq:non_linear_IC_mean}
and \eqref{eq:nonlinear_bc_mean}. Furthermore, \cite{Yeung2022WRR} found the changes in $C_h$ due to $\epsilon$ changes in the BCs are on the order of $\epsilon/L$, where $L$ is the characteristic length of the domain. For most natural systems, the variations in the boundary conditions are much smaller than the domain size (i.e., $\epsilon/L \ll 1$), and the changes in the covariance and its eigenvalues and eigenfunctions due to changes in boundary conditions could be neglected.  This means that TL can be performed by estimating $\overline{h}$ from the mean-field equation \eqref{eq:nonlinear_mean_sim} subject to the target IBCs and transferring $\bm\psi_h$ from the source problem to the target problem. 

Finally, we analyze the transferability of $\bm\theta$ in the $\bm\eta=\mathcal{NN}(\bm{\xi}_k; \bm{\theta})$ model.
To model the nonlinear relationship between $\bm\eta$ and $\bm\xi$, we use a fully connected DNN with $N$ hidden layers.
The DNN is trained using the source dataset $\{\bm\xi^{s,(i)}_k \rightarrow \bm\eta^{s,(i)} \}_{i=1}^{N_{\text{train}}^s}$ by minimizing the loss function \eqref{eq:surrogate_loss}, yielding the DNN parameters $\bm\theta^s = \{ \bm{W}^s_{1:N+1}, \bm{b}^s_{1:N+1}\}$. For the target problem, we freeze DNN parameters except those in the last layer (i.e., we set $\bm{W}^t_{1:N} = \bm{W}^s_{1:N}$ and $\bm{b}^t_{1:N} = \bm{b}^s_{1:N}$) and use the target dataset $\{\bm\xi^{t,(i)}_k \rightarrow \bm\eta^{t,(i)} \}_{i=1}^{N_{\text{train}}^t}$ to retrain parameters in the last layer $\{ \bm{W}^t_{N+1}, \bm{b}^t_{N+1}\}$ by solving the linear LS problem:
\begin{align}\label{eq:surrogate_loss_target}
(\bm{W}^{t,*}_{N+1}, \bm{b}^{t,*}_{N+1}) = \underset{\bm{W}^t_{N+1}, \bm{b}^t_{N+1}}{\arg\min} \sum_{i=1}^{N_{\text{train}}^t} || \mathcal{NN}(\bm{\xi}^{t,(i)}_k; \bm{W}^s_{1:N},\bm{W}^t_{N+1}, \bm{b}^s_{1:N}, \bm{b}^t_{N+1} )- \boldsymbol{\eta}^{t,(i)}||^2_2. 
\end{align}
Given the dimension $n_{N+1}$ of the last layer, $N^t_{\text{train}} \ge n_{N+1}+1$ is needed to find a unique solution of this problem.  If the labeled target dataset is unavailable, the solution can be found using the RLS formulation. Recall that the DNN model of $\bm\eta$ for the target problem is 
\begin{equation}\label{eq:linear_DNN_eta_model}
\hat{\bm\eta} = \bm{W}^t_{N+1}\cdot\bm{\xi}_{N+1}(\bm\xi_k; \bm{W}^s_{1:N}, \bm{b}^s_{1:N}) + \bm{b}^t_{N+1}.
\end{equation}
The residual vector $\bm r^t \in \mathbb{R}^{N_m}$ of Eq \eqref{eq:nonlinear_pde} evaluated on the $N_m = N_t \times N_x$ time-space grid for the target IBC can be written as
\begin{equation}\label{eq:res_nonlinear_comp} 
\bm{r}^t(\bm{W}^t_{N+1},\bm{b}^t_{N+1},\bm\xi_k) =  
\bm{A}(\bm\xi_k) \cdot 
[\bm{W}^t_{N+1} \cdot \bm{\xi}_{N+1}(\bm\xi_k; \bm{W}^s_{1:N}, \bm{b}^s_{1:N}) + \bm{b}^t_{N+1}]
+\bm{b}(\bm\xi_k),
\end{equation} 
where the elements of the matrix $\bm A \in \mathbb{R}^{N_{m} \times N_\eta}$ are 
$$
{a}_{ij} =\left\{ \frac {\partial \psi_{h,j}}{\partial t}
-
\frac{\partial}{\partial x} \left[ \bm\psi_k \cdot \bm\xi_k \frac{\partial \psi_{h,j}}{\partial x} \right] 
-
\frac{\partial}{\partial x} \left[ \overline{k}(x) \frac{\partial \psi_{h,j}}{\partial x} \right] \right\}_{(x,t) = (x_i, t_i)}
$$
and the elements of the vector $\bm b \in \mathbb{R}^{N_m}$ are 
$$
{b}_i = \left\{ \frac {\partial \overline{h}^t}{\partial t} 
-
\frac{\partial}{\partial x} \left[ \overline{k}(x) \frac{\partial \overline{h}^t}{\partial x} \right]
-
\frac{\partial}{\partial x} \left[ \bm\psi_k \cdot \bm\xi_k \frac{\partial \overline{h}^t}{\partial x} \right] \right\}_{ (x,t) = (x_i, t_i)}.
$$
 $\bm{W}^t_{N+1}$ and $\bm{b}^t_{N+1}$ can be found by solving the linear RLS minimization problem 

\begin{equation}\label{eq:PI-KL-DNN}
   (\bm{W}^{t,*}_{N+1}, \bm{b}^{t,*}_{N+1}) = \underset{\bm{W}^t_{N+1}, \bm{b}^t_{N+1}}{\arg\min}  \sum_{i=1}^{N_r} 
    ||\bm{r}^t(\bm{W}^t_{N+1},\bm{b}^t_{N+1},\bm\xi_{k,r}^{(i)}) ||_2^2,
\end{equation}
where $\{\bm\xi^{(i)}_{k,r} \}_{i=1}^{N_r}$ are the random realizations of the $\bm\xi_k$ vector. Similarly, $N^t_{\text{train}} \ge n_{N+1}+1$ is needed to find a unique solution to this problem. The deterministic IBCs $h_0$, $h_l$, and $h_r$ are exactly satisfied by the $h$ KLE because $\overline{h}$ satisfies these IBCs and the eigenfunctions satisfy the homogeneous IBCs \cite{TartakovskyJCP2024}. Therefore, there is no need to add penalty terms in Eq \eqref{eq:PI-KL-DNN} to satisfy the IBCs. We refer to Eq \eqref{eq:PI-KL-DNN} as the physics-informed (PI) DNN method. 

If there are target data $\{\bm\xi^{t,(i)}_k,\bm\eta^{t,(i)} \}_{i=1}^{N_{\text{train}}^t}$ available, they can be assimilated in estimation of $\bm{W}_{N+1}$ and $\bm{b}_{N+1}$ as
\begin{equation}\label{eq:pi_loss_target}
    (\bm{W}^{t,*}_{N+1}, \bm{b}^{t,*}_{N+1}) = \underset{\bm{W}^t_{N+1}, \bm{b}^t_{N+1}}{\arg\min} 
    \Big[ 
    \lambda_r \sum_{i=1}^{N_r} 
    ||\bm{r}^t(\bm{W}^t_{N+1},\bm{b}^t_{N+1},\bm\xi_{k,r}^{(i)}) ||_2^2
+ \sum_{i=1}^{N_{\text{train}}^t} 
|| \mathcal{NN}(\bm\xi^{t,(i)}_k;\bm{W}^s_{1:N},\bm{W}^t_{N+1}, \bm{b}^s_{1:N}, \bm{b}^t_{N+1})- \bm{\eta}^{t,(i)}||^2_2
    \Big],
\end{equation}
where $\lambda_r$ is the weight.  

To further investigate the relationship between $\bm\eta$ and $\bm\xi_k$ for $\sigma_y^2 <1$, we simplify the fluctuation  equation by disregarding the terms $\nabla [k' \nabla h']$ and 
  $\overline{\nabla  k'(x) \nabla h'}$, which are of the $\sigma^2_y$ order (the rest of the terms in this equation are of the $\sigma_y$ order), as 
\begin{equation}\label{eq:fluctuations_sim}
  \frac {\partial h'}{\partial t}  = 
\frac{\partial}{\partial x} \left[ \overline{k} (x) \frac{\partial h'}{\partial x} \right]
+
\frac{\partial}{\partial x} \left[ k'(x) \frac{\partial \overline{h}}{\partial x} \right], 
\end{equation}
subject to the homogeneous IBCs \eqref{eq:fluc_IC} and \eqref{eq:nonlinear_bc_fluctuations}.
This equation is linear in $h'$ and $k'$, and  $\bm\eta$ can be found as a function $\bm\xi_k$ using the RLS method. 
The residual vector  $\bm r \in \mathbb{R}^{N_m}$  of Eq \eqref{eq:fluctuations_sim} can be written as 
\begin{align}
  \bm{r}(\bm\eta,\bm\xi_k) &=  \bm A \bm \eta - \bm B \bm \xi_k 
\end{align}
where $\bm A \in \mathbb{R}^{N_m \times N_\eta}$ with elements
\begin{equation}
a_{ij} = 
\left[
 \frac{\partial \phi_h^{j} (x,t)}{\partial t} - \frac{\partial} {\partial x} \left( \overline{k} (x)\frac{\partial \phi_h^{j}(x,t) }{\partial x} \right)
\right]_{x = x_i,t = t_i}
\end{equation}
and $\bm B \in \mathbb{R}^{N_m \times N_\xi}$ with elements
\begin{equation}
b_{ij} = 
\left[
\frac{\partial }{\partial x} 
\left(\phi_k^{j}(x)
\frac{\partial \overline{h}(x,t)}{\partial x}  \right)
\right]_{x = x_i,t = t_i}.
\end{equation}
For a given $\bm\xi_k$, $\bm\eta$ is obtained by minimizing $||\bm{\hat{r}} ||_2^2$:
\begin{align}
\bm{\eta}^* &= \min_{\bm{\eta}} ||\bm{r}(\bm{\eta},\bm{\xi}_k) ||_2^2.  
\end{align}
The solution of this (linear) residual least square problem is given by the normal equation as
\begin{equation}
\bm\eta^*  
=
(\bm{A}^{\text{T}} \bm{A} )^{-1}\bm{A}^{\text{T}} \bm{B} \cdot \bm\xi_k = \bm{W}\cdot \bm\xi_k. 
\end{equation}
In this expression, $\bm{A}$ is independent of $\overline{h}$ and can be transferred from the source to the target problem while  $\bm{B}$ depends on $\overline{h}$ and must be recomputed for the target problem. If some samples of the PDE solution under target IBCs are available, then $\bm W$ for the target problem can be computed using the OLS method from Eq \eqref{eq:regression_l2}. 

The proposed TL method for the KL-DNN surrogate of a PDE with the control variable $\bm{y}$ is schematically depicted in Figure \ref{fig:tl_schematics}.

\section{Numerical examples}\label{sec:examples}

\subsection{Linear diffusion problem}\label{sec:linear_diff_example}

We consider the one-dimensional diffusion equation with point and distributed source terms:

\begin{equation}\label{eq:linear_diffusion}
\frac {\partial h}{\partial t} = \frac{\partial}{\partial x} \left[ k(x) \frac{\partial h}{\partial x} \right]
+ f(x,t) + q(t)\delta(x-x^*) , \quad x \in (0, L), \quad t \in (0, T]
\end{equation}
subject to the initial condition
\begin{equation}\label{eq:linear_diffusion_ic}
    h(x,t=0)= h_0 , \quad x \in (0, L),
\end{equation}
and the boundary conditions
\begin{equation}\label{eq:linear_diffusion_bc}
    h(x=0,t)= h_l\quad h(x=L,t)= h_r , \quad t \in [0, T],
\end{equation}
where $L = 1$, $T=0.03$, and $\delta(\cdot)$ is the Dirac delta function. The location of the point source $x^*$ is known and fixed at $x^* = 0.25L$.  The space-dependent diffusion parameter $k(x) = \exp(y(x))$ is known, and $y(x)$ is generated as a realization of the zero-mean Gaussian process $\mathcal{GP} \{ 0, C_y(x,x')\}$, where $C_y(x,x') = \sigma^2_y \exp \left[ -\frac{(x-x')^2}{l_y^2}\right]$ is the squared exponential covariance function with the variance $\sigma^2_y=0.6$ and the correlation length $l_y=0.5L$.

The distributed source function $f(x,t)$, the point source rate $q(t)$, and the IBCs are chosen as control variables with the ranges defined by the stochastic processes $F(x,t) \sim \mathcal{GP}\{ \overline{f}(x,t), C_f(x,x',t,t') \}$, 
$Q(t) \sim \mathcal{GP}\{ \overline{q}(t),C_q(t,t') \}$ and random variables $H_0\sim \mathcal{U}(h_0^{\text{min}}, h_0^{\text{max}})$, $H_l\sim \mathcal{U}(h_l^{\text{min}}, h_l^{\text{max}})$, and $H_r\sim \mathcal{U}(h_r^{\text{min}}, h_r^{\text{max}})$. Here, $\mathcal{U}$ denotes a uniform distribution. The squared exponential covariances are assumed:  $C_f(x,x',t,t') = \sigma^2_f \exp \left[ -\frac{(x-x')^2}{l_f^2} - \frac{(t-t')^2}{\tau_f^2} \right]$ and  $C_q(t,t') = \sigma^2_q \exp \left[-\frac{(t-t')^2}{\tau_q^2} \right]$ with, in general,  the variances $\sigma^2_f$ and $\sigma^2_q$, the correlation lengths $l_f$, and correlation times $\tau_f$ and $\tau_q$ under source and target conditions. 

To generate the source training dataset, $f$ and $q$ are sampled using the truncated $F$ and $Q$ KLEs 
with $N^\xi_f = 200$ and $N^\xi_q = 40$ terms, respectively, corresponding to ${rtol}_f= \num{1.45e-9}$ and ${rtol}_q= \num{1.39e-10}$. 
The IBCs $h_0$, $h_r$, and $h_l$ are sampled from their uniform distributions. The source training dataset size is $N^s_{\text{train}} = 1000$.

In the KL-NN surrogate model for the source problem:
\begin{align}
    \hat{h}(x,t|f,Q,h_0,h_l,h_r) =
     \mathcal{KL} \left[\overline{h}^s,  \bm\psi_h^s, \bm{W}^s\bm\xi^s \right],  \nonumber 
\end{align}
we use $N^\eta = 40$ terms, corresponding to ${rtol}_h=\num{7.6e-6}$. The mean $\overline{h}^s$ and covariance $C^s_h$ are computed from Eqs \eqref{eq:ens_mean} and \eqref{eq:ens_cov}, and $\bm\psi_h^s$ is found by solving the eigenvalue problem \eqref{eq:space-time-covar}. The matrix $\bm{W}^s$ is estimated using the OLS method in  \eqref{eq:OLS_solution}. In the parameter vector
${\bm\xi^s}^{\text{T}} = \left[ {\bm{\xi}^s_f}^{\text{T}},  {\bm{\xi}^s_q}^{\text{T}}, h'_0, h'_r, h'_l \right] $, $\bm{\xi}^s_f$ and $\bm{\xi}^s_q$ are obtained using the inverse source KL operators
$
   \bm{\xi}^s_f = \mathcal{KL}^{-1}\left[f,\overline{f}^s, \bm\psi^s_f \right]  
$ 
and
$
   \bm{\xi}^s_q = \mathcal{KL}^{-1}\left[q,\overline{q}^s, \bm\psi^s_q \right]
$.
The input parameters $h'_0$, $h'_r$, and $h'_l$ are found as
\begin{equation}
    h'_0 = h_0 - \overline{h}_0, \quad h'_l = h_l-\overline{h}_l, \quad
    h'_r = h_r - \overline{h}_r. \nonumber 
\end{equation}

The KL-NN surrogate model for the target problem takes the form
\begin{equation}
       \hat{h}(x,t|f,q,h_0,h_l,h_r) =
     \mathcal{KL} \left[ \overline{h}^t,  \bm\psi_h^s,\bm{W}^s\bm\xi^t \right]. \nonumber 
\end{equation}
The target mean function $\overline{h}^t (x,t)$ is found by solving the mean-field PDE:
\begin{equation}\label{eq:mean_field_pde_linear_diffusion}
\frac {\partial \overline{h} }{\partial t} = \frac{\partial}{\partial x} \left[ k(x) \frac{\partial \overline{h}}{\partial x} \right]
+ \overline{f}^t(x,t) + \overline{q}^t(t)\delta(x-x^*), \quad x \in [0, L], \quad t \in [0, T]
\end{equation}
subject to the initial condition
\begin{equation}\label{eq:mean_field_ic_linear_diffusion}
    \overline{h}(x,t=0)= \overline{h}_0 , \quad x \in [0, L],
\end{equation}
and the boundary conditions
\begin{equation}\label{eq:mean_field_bc_linear_diffusion}
    \overline{h}(x=0,t)= \overline{h}_l, \quad  \overline{h}(x=1,t)= \overline{h}_r, \quad t \in [0, T].
\end{equation}
In the input parameter vector ${\bm\xi^t}^{\text{T}} = \left[ {\bm{\xi}^t_f}^{\text{T}},{\bm{\xi}^t_q}^{\text{T}}, h'_0, h'_r, h'_l \right] $,  $\bm{\xi}^t_f$ and $\bm{\xi}^t_q$ are obtained using the inverse TL-KL operators with target mean values and source eigenfunctions:
$
   \bm{\xi}^t_f = \mathcal{KL}^{-1} \left[f;\overline{f}^t, \bm\psi^s_f \right]  
$
and
$
   \bm{\xi}^t_q = \mathcal{KL}^{-1} \left[q;\overline{q}^t, \bm\psi^s_q \right].
$
The TL model is tested for $f$ and $q$ generated as random realizations of the target $F$ and $Q$ distributions.  

We consider three target conditions (T1, T2, and T3) listed in Table \ref{tab:linear_summary}. In target problem T1, the mean function of the target control variables differs from that of the source (the variances and correlation lengths/times are the same under source and target conditions). In target problem T2, the mean target control variables are as in T1 and the correlation lengths/times of the target control variables are scaled by a factor of $\alpha$ compared to the source as  $\left[ l_f^t, \tau_f^t, \tau_q^t \right] = \alpha \left[ l_f^s, \tau_f^s, \tau_q^s \right]$.  In target problem T3, the variances of the target control variables are scaled by a factor of $\beta$ relative to the source as $ \left[ \sigma^{2,t}_f, \sigma^{2,t}_q \right] = \beta \left[ \sigma^{2,s}_f, \sigma^{2,s}_q \right]$. The KL-NN errors for the source and target conditions are computed with respect to the FD reference source and target solutions and listed in Table \ref{tab:linear_summary}. 

\renewcommand{\arraystretch}{1.4}
\begin{table}[!htb]
\centering
\caption{Linear diffusion problem: The summary of source and target (T1, T2, and T3) conditions and the relative errors $\varepsilon$ in the KL-NN surrogate model subject to these conditions.}
\begin{tabular}{cccccc}
\hline
  & \multicolumn{4}{c}{\textbf{Problem setup}} & \textbf{Error} \\
\hline
    \multirow{4}{*}{\textbf{Source}} 
            & 
            \multicolumn{4}{l}{
            $f$: $\overline{f}^s(x, t) = 0.5 \sin\left(2 \pi x \cos\left(10 \pi t\right)\right)$, $\sigma^{2,s}_f =10$, $l_f^s=0.5L$, $\tau_f^s=0.5T$} 
            & \multirow{4}{*}{ \num{2.95e-04}}  \\
            & 
            \multicolumn{4}{l}{
            $q$: $ \overline{q}^s(t) = \sin\left( 2 \pi t/T \right)$, $\sigma^{2,s}_q=1$, $\tau_q^s=0.1T$ 
            } 
            &  \\
            & 
            \multicolumn{4}{l}{
            IC: $h_0^{\text{min}} = 0.975L$, $h_0^{\text{max}} = 1.025L$ 
            } 
            & \\
            & 
            \multicolumn{4}{l}{
            BC: $h_l^{\text{min}} = 1.025L$, $h_l^{\text{max}} = 1.2L$; $h_r^{\text{min}} = 0.8L$, $h_r^{\text{max}} = 0.975L$ 
            } 
            & \\
\hline
    \multirow{11}{*}{\textbf{Target}} 
        & \multicolumn{3}{l}{
            T1:  
        } &  & \multirow{3}{*}{\num{3.12e-04}} \\
        & \multicolumn{3}{l}{ \hspace{0.64cm} $\overline{f}^t(x,t) = 1.2 \left[\exp(x) + t^3 - tx \right]$, $\sigma^{2,t}_f =\sigma^{2,t}_s$, $l_f^t=l_f^s$, $\tau_f^t=\tau_f^s$
        } &  &  \\
        & \multicolumn{3}{l}{ \hspace{0.64cm} $\overline{q}^t(t) = \cos\left( 2 \pi t/T \right)$, $\sigma^{2,t}_q=\sigma^{2,s}_q$, $\tau_q^t=\tau_q^s$
        } &  &  \\
        \cline{2-6}
        & 
         \multicolumn{2}{l}{
            T2: 
        }
        & $\gamma = 1$ & $\alpha = 0.5$ & \num{9.12e-4} \\ 
        \cline{4-6}
        & \multicolumn{2}{l}{
           \hspace{0.64cm} $\overline{f}^t$ and $\overline{q}^t$ are as in T1, \hspace{2cm}
        } & $\gamma = 0.01$ & $\alpha = 0.8$ & \num{3.16e-4} \\
        \cline{4-6}
        & \multicolumn{2}{l}{
           \hspace{0.64cm} $ \left[ \sigma^{2,t}_f, \sigma^{2,t}_q \right] =  \left[ \sigma^{2,s}_f, \sigma^{2,s}_q \right]$,
           \hspace{2cm}
        } & $\gamma = 0$ & $\alpha = 1.2$ & \num{1.63e-04} \\
        \cline{4-6}
        & \multicolumn{2}{l}{
            \hspace{0.64cm} and $\left[ l_f^t, \tau_f^t, \tau_q^t \right] = \alpha \left[ l_f^s, \tau_f^s, \tau_q^s \right]$
            \hspace{2cm}
        } & $\gamma = 0$ & $\alpha = 1.5$ & \num{1.31e-04} \\
        \cline{2-6}
        & \multicolumn{2}{l}{
            T3: 
        }
        & $\gamma = 0$ & $\beta = 0.5$ & \num{1.93e-04} \\ 
        \cline{4-6}
        & \multicolumn{2}{l}{
            \hspace{0.64cm} $\overline{f}^t$ and $\overline{q}^t$ are as in T1,
            \hspace{2cm}
        } & $\gamma = 0$ & $\beta = 0.8$ & \num{2.67e-04} \\
        \cline{4-6}
        & \multicolumn{2}{l}{
           \hspace{0.64cm} $\left[ l_f^t, \tau_f^t, \tau_q^t \right] = \left[ l_f^s, \tau_f^s, \tau_q^s \right]$,
           \hspace{2cm}
        } & $\gamma = 0$ & $\beta = 1.2$ & \num{3.13e-04} \\
        \cline{4-6}
        & \multicolumn{2}{l}{
            \hspace{0.64cm} and $ \left[ \sigma^{2,t}_f, \sigma^{2,t}_q \right] = \beta \left[ \sigma^{2,s}_f, \sigma^{2,s}_q \right]$
            \hspace{2cm}
        } & $\gamma = 0$ & $\beta = 1.5$ & \num{3.43e-04} \\
\hline
\end{tabular}
\label{tab:linear_summary}
\end{table} 

\begin{figure}[hbt!]
\centering
\begin{subfigure}[b]{0.33\textwidth}
    \centering
    \includegraphics[width=\textwidth]{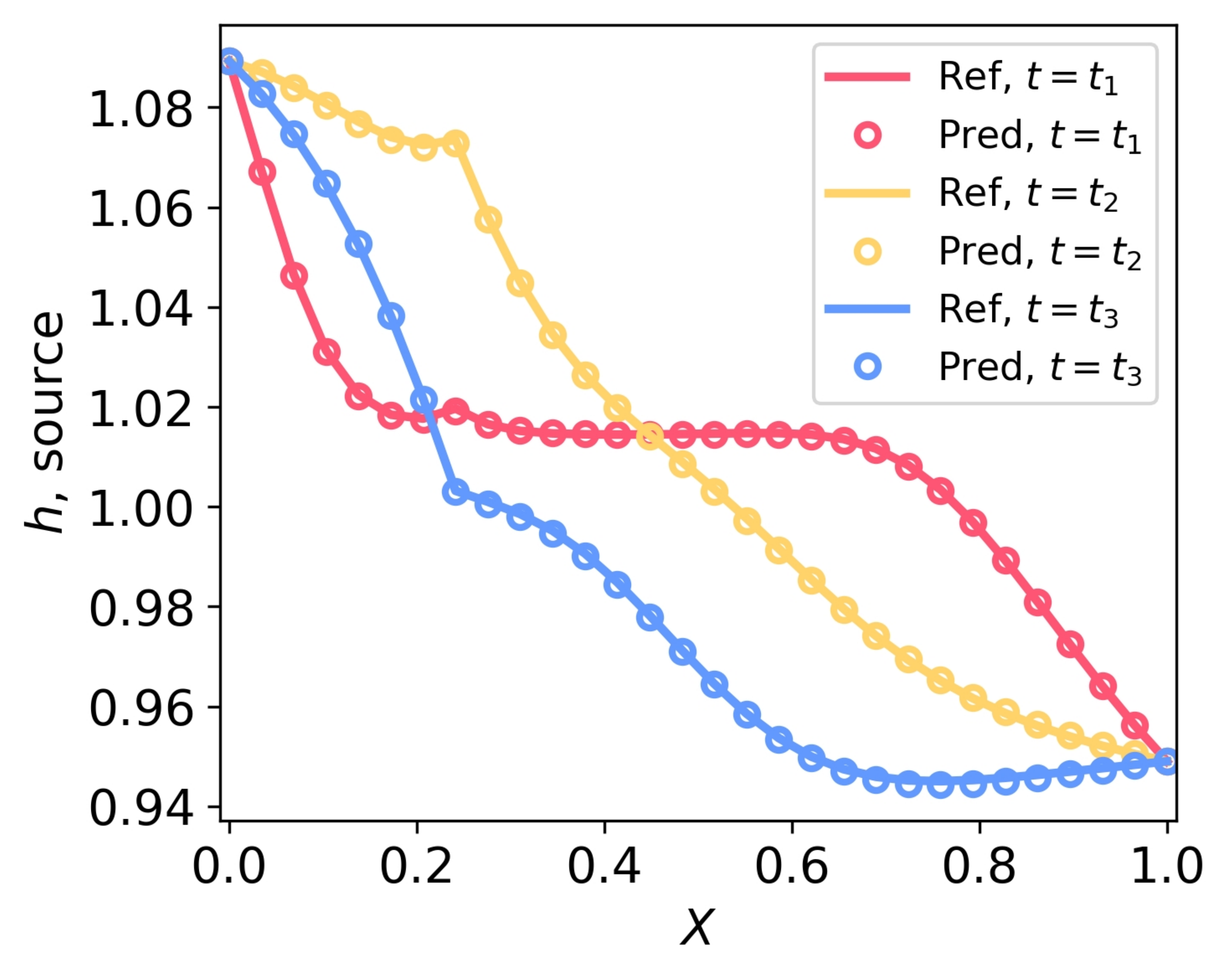}
    \caption{}
\end{subfigure}
\begin{subfigure}[b]{0.33\textwidth}
    \centering
    \includegraphics[width=\textwidth]{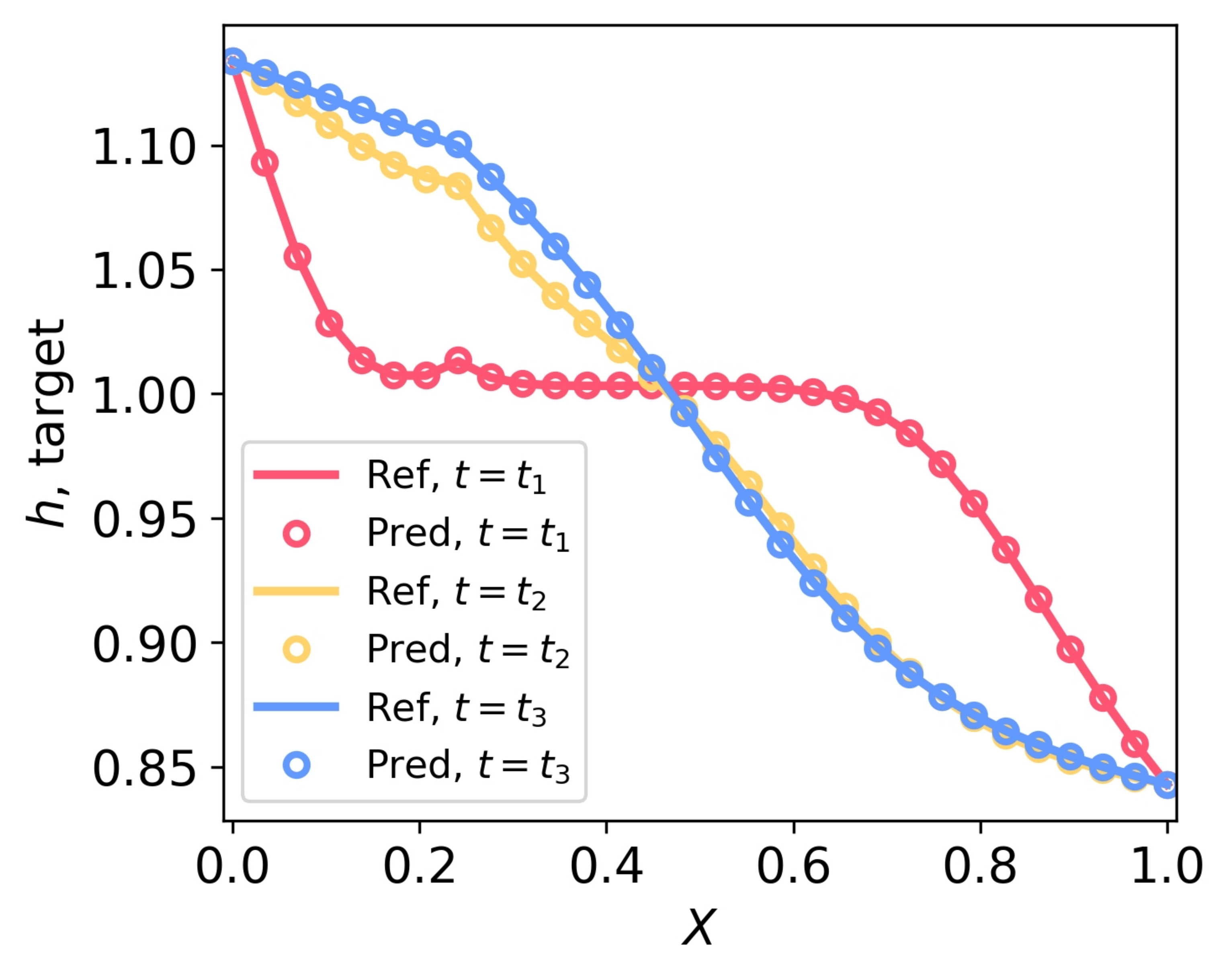}
    \caption{}
\end{subfigure}
\caption{Linear diffusion problem: (a) source and (b) target $h$ solutions versus the reference solutions at three selected times ($t_1 = T/50, t_2 = T/5, t_3 = T$). The reference solutions are shown with a solid line, and KL-NN predictions are marked with open circles.}
\label{fig:linear_OLS_solutions}
\end{figure}

The comparison of the KL-NN and reference solutions for the source and target T1 conditions is given in Figure \ref{fig:linear_OLS_solutions}. 
The KL-DNN and reference solutions are in close agreement for both the source and T1 conditions with the relative $\ell^2$ errors $\varepsilon = 2.95 \times 10^{-4}$ for the source condition and $\varepsilon = 3.12 \times 10^{-4}$ for the target condition. These results agree with our analysis that for linear problems, the proposed one-shot TL learning is exact (i.e., gives similar errors under source and target conditions), assuming that the covariance models of the control variables do not change from the source to target conditions. 

\begin{figure}[hbt!]
    \centering
    \includegraphics[width=0.33\textwidth]{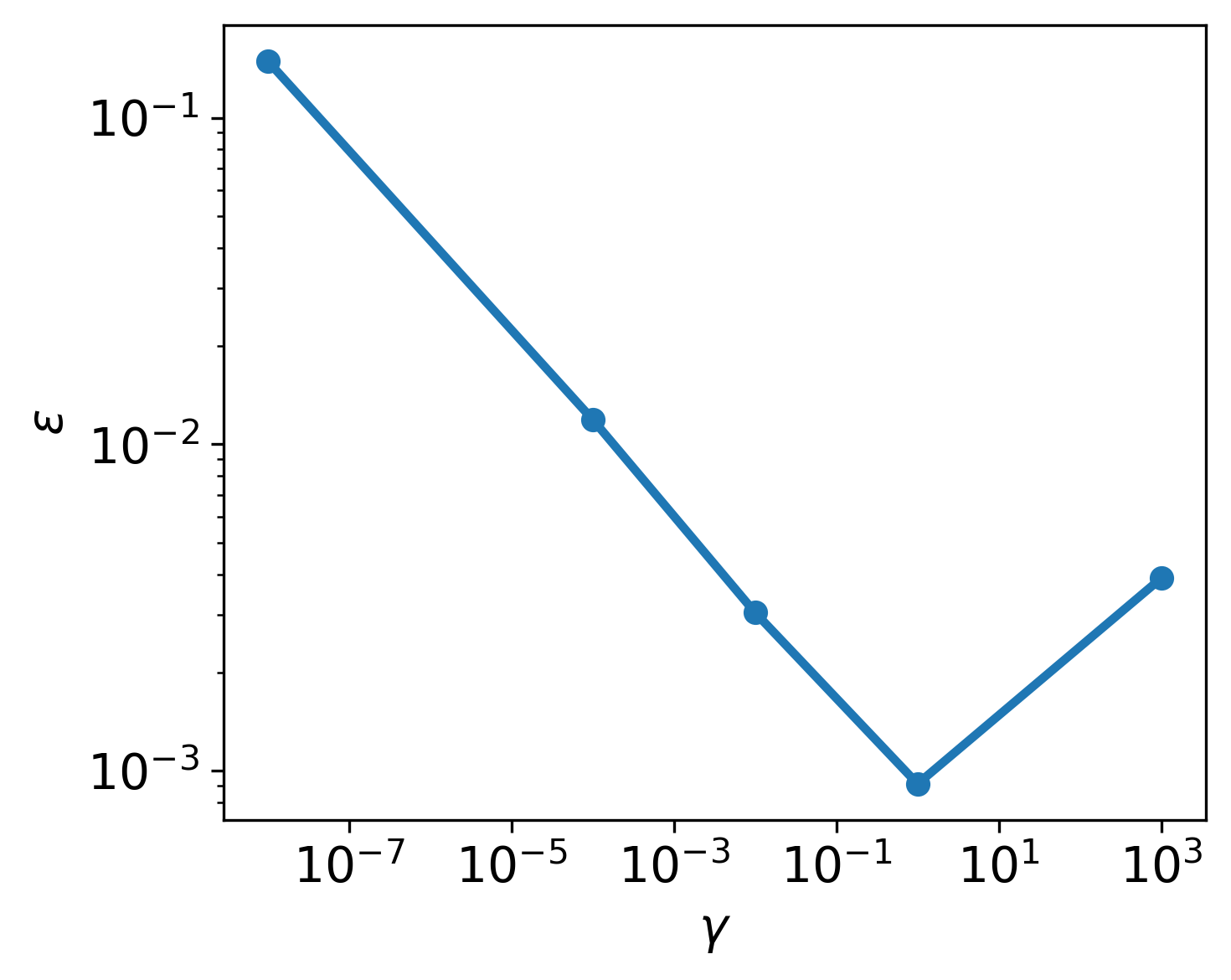}
\caption{Relative error $\varepsilon$ in the T2 target solution with $\alpha=0.5$ as a function of $\gamma$.}
\label{fig:error_vs_gamma}
\end{figure}

In the T2 case, the target KL-NN errors $\varepsilon$ are close to the source $\varepsilon$ for $\alpha >1$. 
However, the target $\varepsilon$ increases to \num{1.95e-3} for $\alpha = 0.8$ and \num{1.19e-2} for $\alpha = 0.5$. 
The target error $\varepsilon$ increases for $\alpha <1$ because the truncated (forward and inverse) KL operators might not contain (source) eigenfunctions with sufficiently high frequencies, which might be present in the target control variables. In turn, this might lead to larger $\varepsilon_{KL}$ errors in the representation of the target control variables and make the inverse KL operators 
$
   \bm{\xi}^t_f = \mathcal{KL}^{-1} \left[f;\overline{f}^t \bm\psi^s_f \right]  
$
and
$
   \bm{\xi}^t_q = \mathcal{KL}^{-1} \left[q;\overline{q}^t, \bm\psi^s_q \right]
$
unstable. The regularization term in the inverse KL operator is critical to provide an accurate TL solution for $\alpha<1$. 
Figure \ref{fig:error_vs_gamma} shows the relative error $\epsilon$ in the target solution as a function of the regularization coefficient $\gamma$ for $\alpha=0.5$. The smallest error is achieved for $\gamma=1$.

For the TL3 target conditions, the KL-NN errors $\varepsilon$ are similar under source and target conditions for all considered $\beta$ values. We note that $\gamma=0$ is used to obtain the target solution. These results indicate that the accuracy of the TL solution does not depend on the variance of the control variable target ranges.  

\subsection{Nonlinear diffusion problem}
Here, we consider the one-dimensional diffusion problem \eqref{eq:nonlinear_pde}-\eqref{eq:nonlinear_pde_bc} and treat $k(x)$ as a control variable. We set $L = 1$, $T = 0.03$, and assume that the range of $y(x) = \log \left[k(x) \right]$ is defined by the Gaussian distribution $\mathcal{GP} \{ 0, C_y(x,x')\}$, with $C_y(x,x') = \sigma^2_y \exp \left[ -\frac{(x-x')^2}{l_y^2}\right]$ and $l_y=0.5L$. The IBCs are assumed to be deterministic but different for the source and target problems. We aim to construct a surrogate model $\hat{h}(x,t|k)$. The source and target IBCs are set to $h_0^s = h_l^s = 1.05L$, $h_r^s = 0.95L$, $h_l^t = 0.95L$, and $h_0^t = h_r^t = 1.05L$. We consider $\sigma^2_y = 0.1$, $0.3$, and $0.6$ to analyze the effect of $\sigma^2_y$ on the accuracy of the TL methods.

The source dataset $D_{\text{train}}^s = \{k^{s, (i)}\rightarrow h^{s, (i)}\}_{i=1}^{N^s_{\text{train}}}$ is constructed by (i) generating $N^s_{\text{train}} =1000$ realizations of $y^{s, (i)}$ using a truncated $y$ KLE with $20$ terms (${rtol}_y=\num{4.9e-8}$), (ii) computing $ k^{(i)}=\ln y^{(i)} $, and (iii) solving the PDE \eqref{eq:nonlinear_pde}-\eqref{eq:nonlinear_pde_bc} for $h^{(i)}$ with the source IBCs on a regular $N_t \times N_x = 250\times 30$ mesh using the finite difference method. The $D_{\text{train}}^s$ dataset is used to compute the KLDs of $k$ and $h$, with $N^\xi_k=20$ (${rtol}_k=\num{7.9e-10}$) and $N^\eta_h = 40$ (${rtol}_h=\num{5.4e-5}$) terms, respectively. The $\bm{\xi}^s_k$ and $\bm{\eta}^s$ are given by the inverse KLD operators
$
   \bm{\xi}^s_k = \mathcal{KL}^{-1} \left[k,\overline{k}^s, \bm\psi^s_k \right]  
$ 
and
$
   \bm{\eta}^s = \mathcal{KL}^{-1} \left[h,\overline{h}^s, \bm\psi^s_h \right]
$.

In this problem, we compare surrogate models with the linear and nonlinear $\bm\eta(\bm\xi_k)$ relationships:
\begin{equation}\label{eq:linear}
\hat{h}(x,t|k) =
     \mathcal{KL} \left[\overline{h}^s, \bm\psi_h^s,\bm{W}^s\bm\xi_k \right],  
\end{equation}
and
\begin{equation}\label{eq:nonlinear}
\hat{h}(x,t|k) =
     \mathcal{KL} \left[\overline{h}^s, \bm\psi_h^s,\mathcal{NN}(\bm\xi_k; \bm\theta^s) \right].  
\end{equation}
The surrogate model for the target IBCs ($h_0^t$, $h_l^t$, and $h_r^t $) is constructed as
\begin{equation}\label{eq:target_kl_linear}
\hat{h}(x,t|k)=
     \mathcal{KL} \left[\overline{h}^t, \bm\phi_h^s,\bm{W}^t\bm\xi_k \right]  
\end{equation}
for the surrogate model \eqref{eq:linear} and 
\begin{equation}\label{eq:target_kl_DNN}
\hat{h}(x,t|k)=
     \mathcal{KL} \left[\overline{h}^t, \bm\psi_h^s,\mathcal{NN}(\bm\xi_k; \tilde{\bm\theta}^s,{\bm W}^t_{N+1}, {\bm b}^t_{N+1} ) \right]  
\end{equation} 
for the surrogate model \eqref{eq:nonlinear}.
In Eq \eqref{eq:target_kl_linear}, we compute $\bm{W}^t$ using the RLS and OLS methods.
In Eq \eqref{eq:target_kl_DNN}, we use 
a fully connected DNN with three hidden layers and 50 neurons per layer, and estimate ${\bm W}^t_{N+1}$ and $ {\bm b}^t_{N+1}$ using Eq \eqref{eq:surrogate_loss_target} (the KL-DNN method)  or Eq \eqref{eq:pi_loss_target} (the PI-KL-DNN method).
In both target surrogate models, the mean $\overline{h}^t (x,t)$ is found as the solution of the zero-moment mean field equation \eqref{eq:nonlinear_mean_sim} subject to the target IBCs. 

\begin{figure}[hbt!]
\centering
\includegraphics[width=0.28\textwidth]{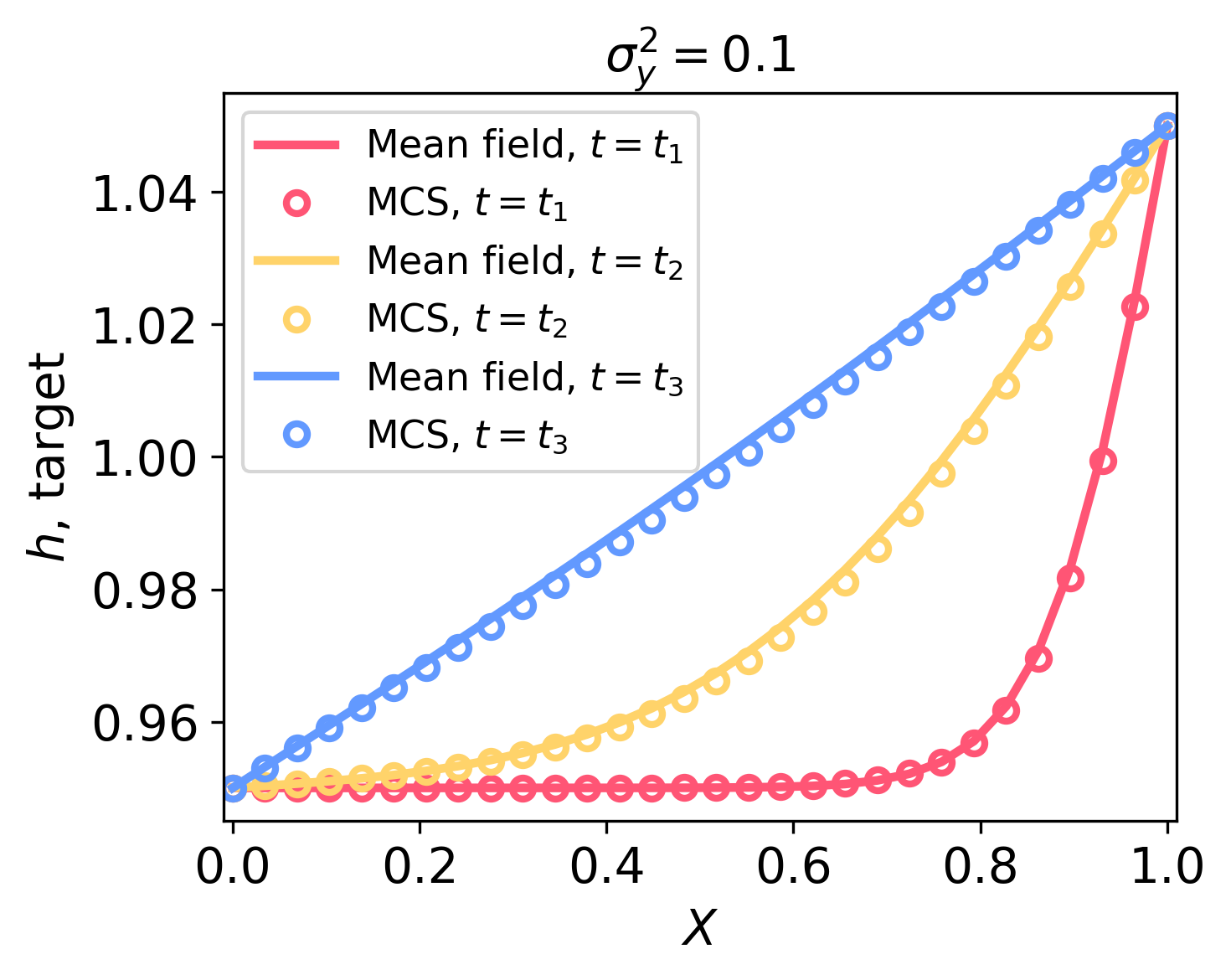}
\includegraphics[width=0.28\textwidth]{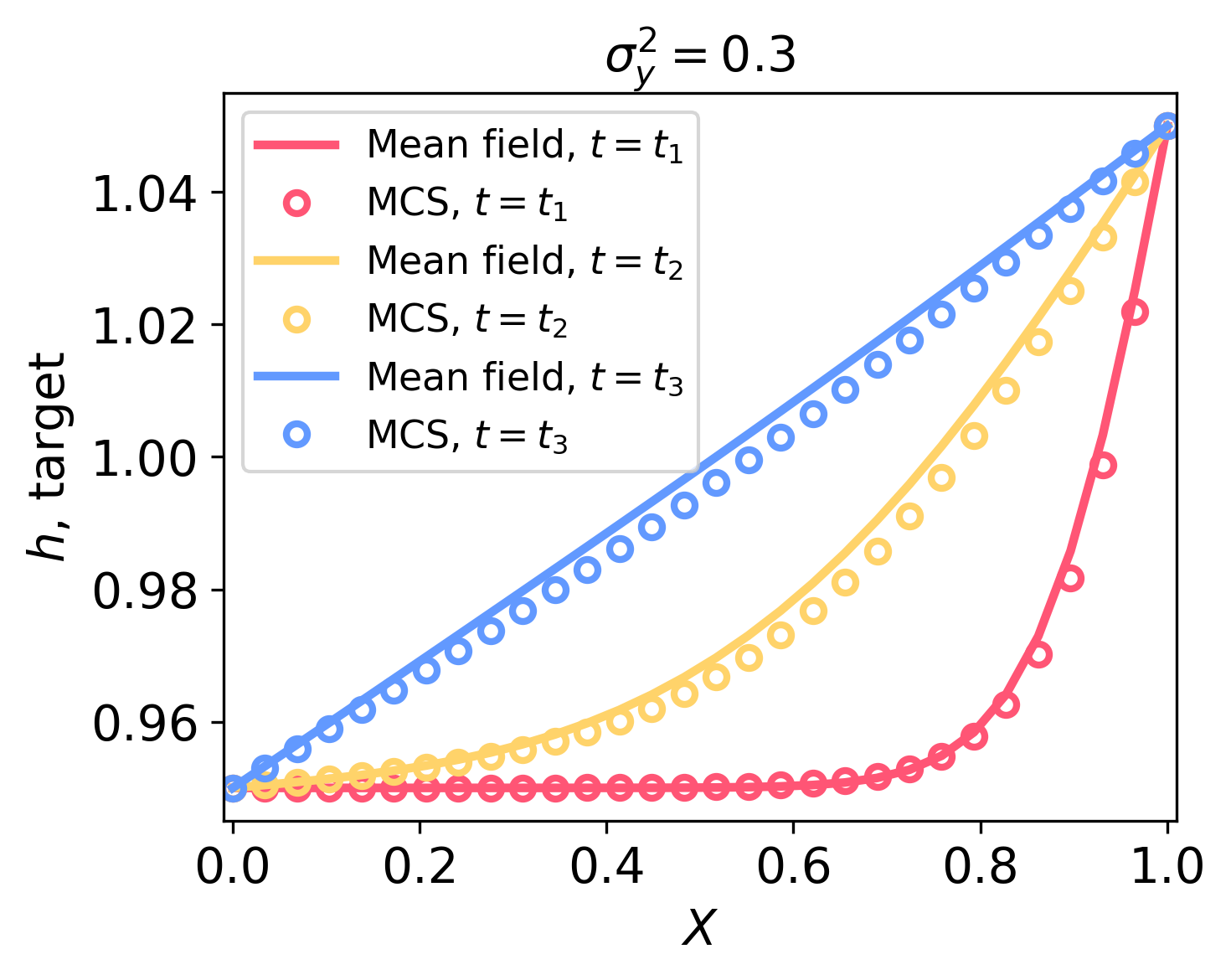}
\includegraphics[width=0.28\textwidth]{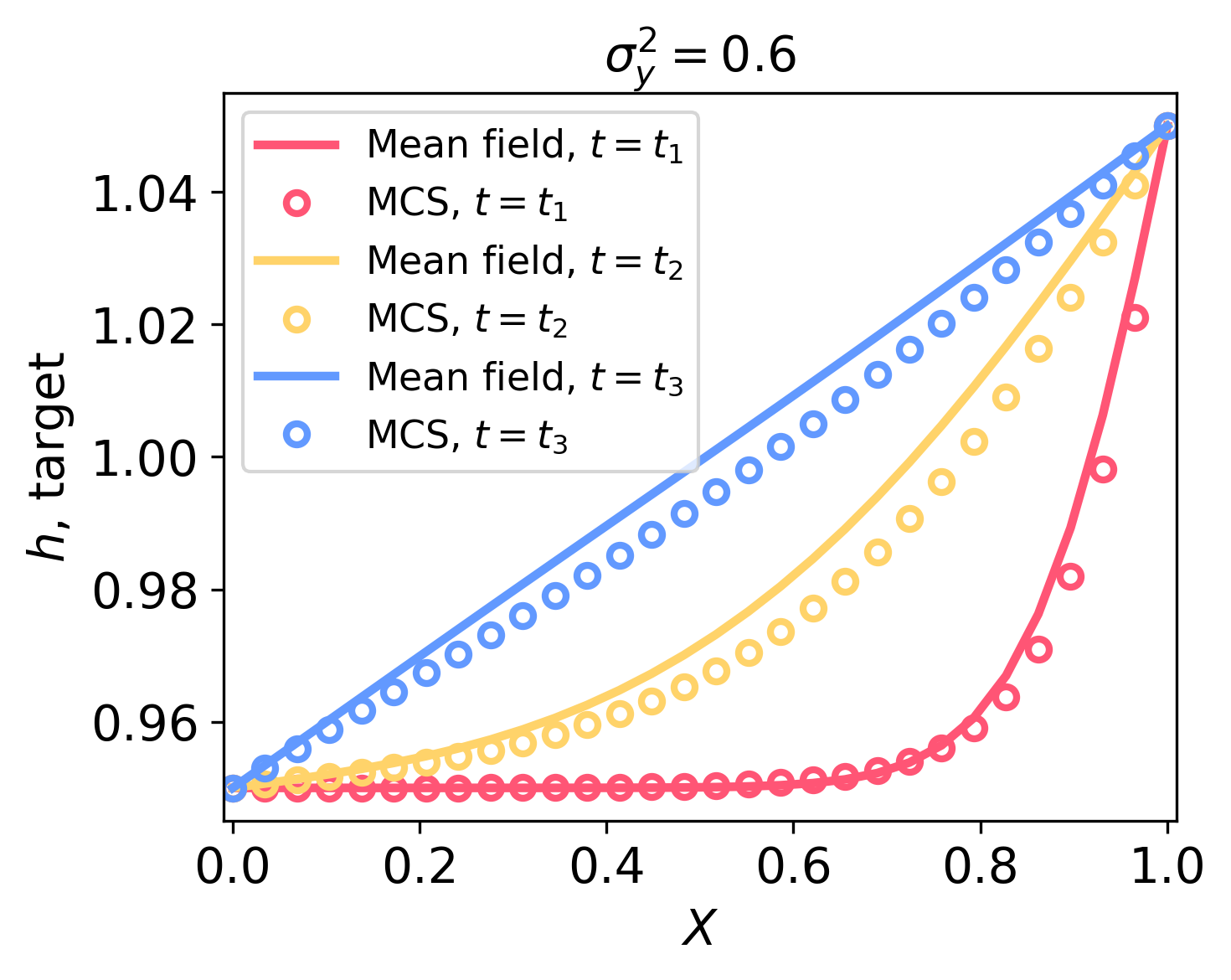}
\caption{Nonlinear diffusion equation: the target mean $\overline{h}^t(x,t)$ solution obtained from MCS (open circles) and the mean-field equation (solid line) for $\sigma^2_y=0.1$ (left), 0.3 (middle), and 0.6 (right).
}
\label{fig:mean_h}
\end{figure}

\begin{figure}[hbt!]
\centering
\begin{subfigure}[b]{0.76\textwidth}
    \centering
    \includegraphics[width=\textwidth]{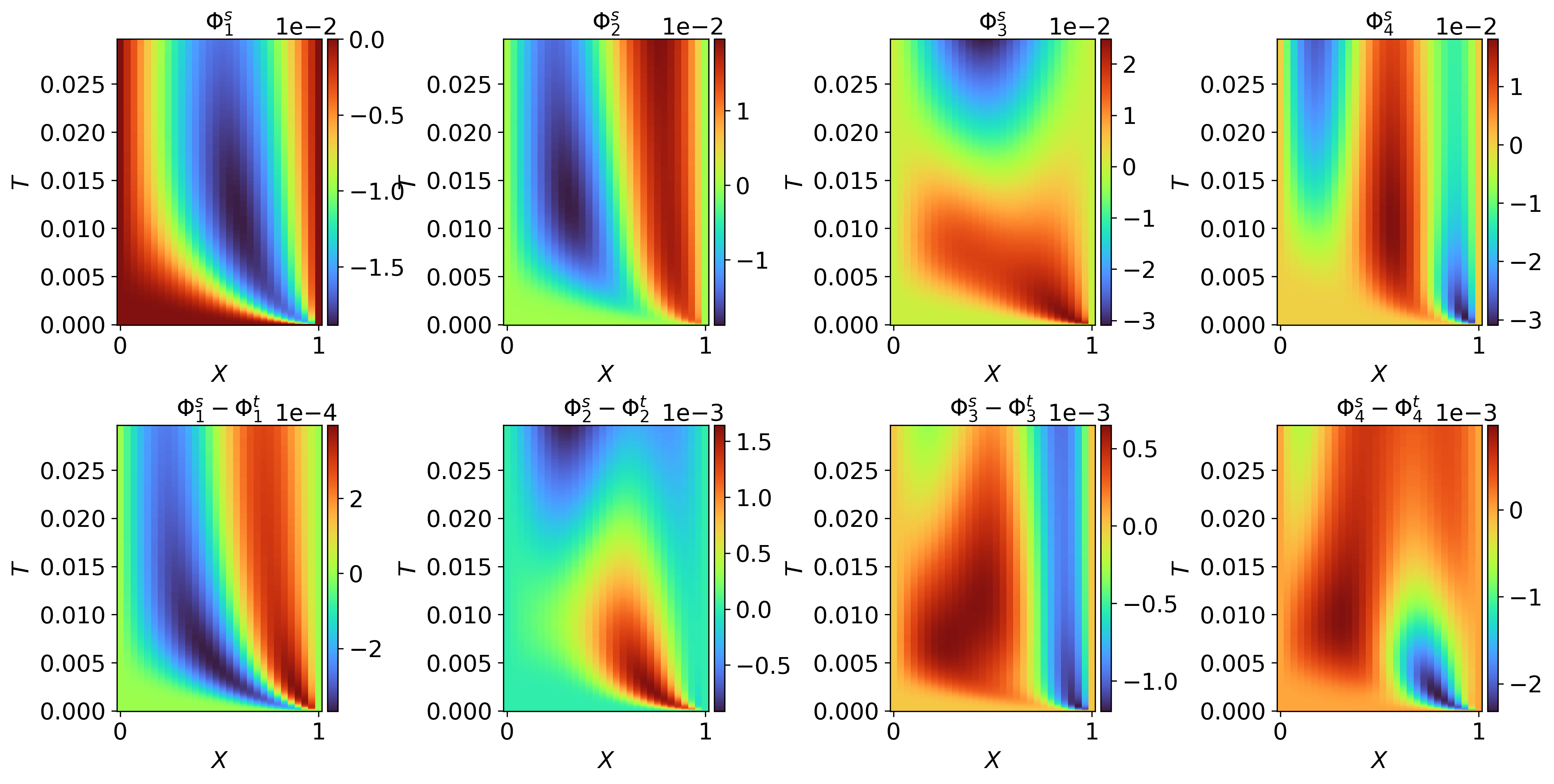}
    \caption{}
\end{subfigure}

\begin{subfigure}[b]{0.76\textwidth}
    \centering
    \includegraphics[width=\textwidth]{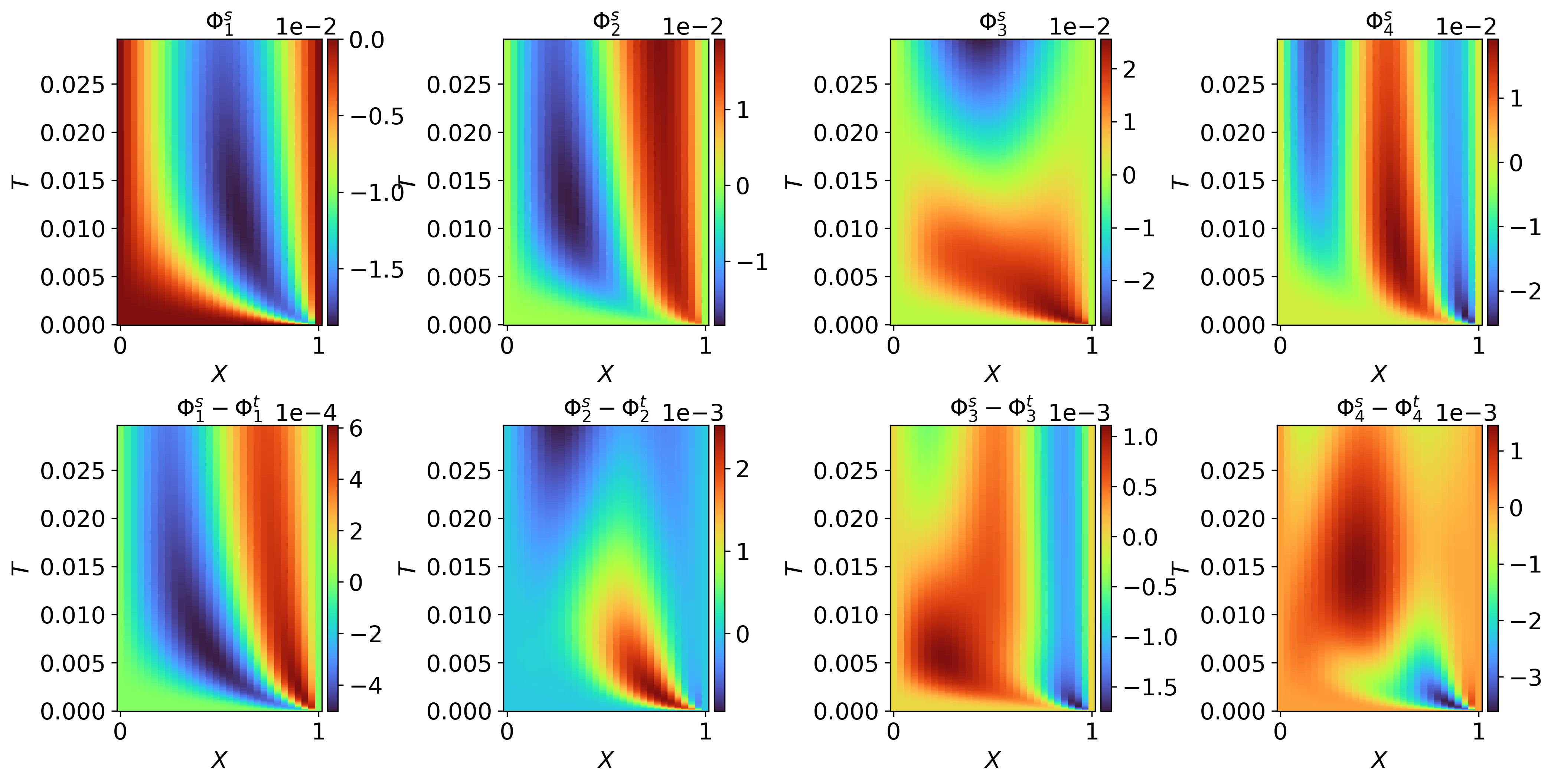}
    \caption{}
\end{subfigure}

\begin{subfigure}[b]{0.76\textwidth}
    \centering
    \includegraphics[width=\textwidth]{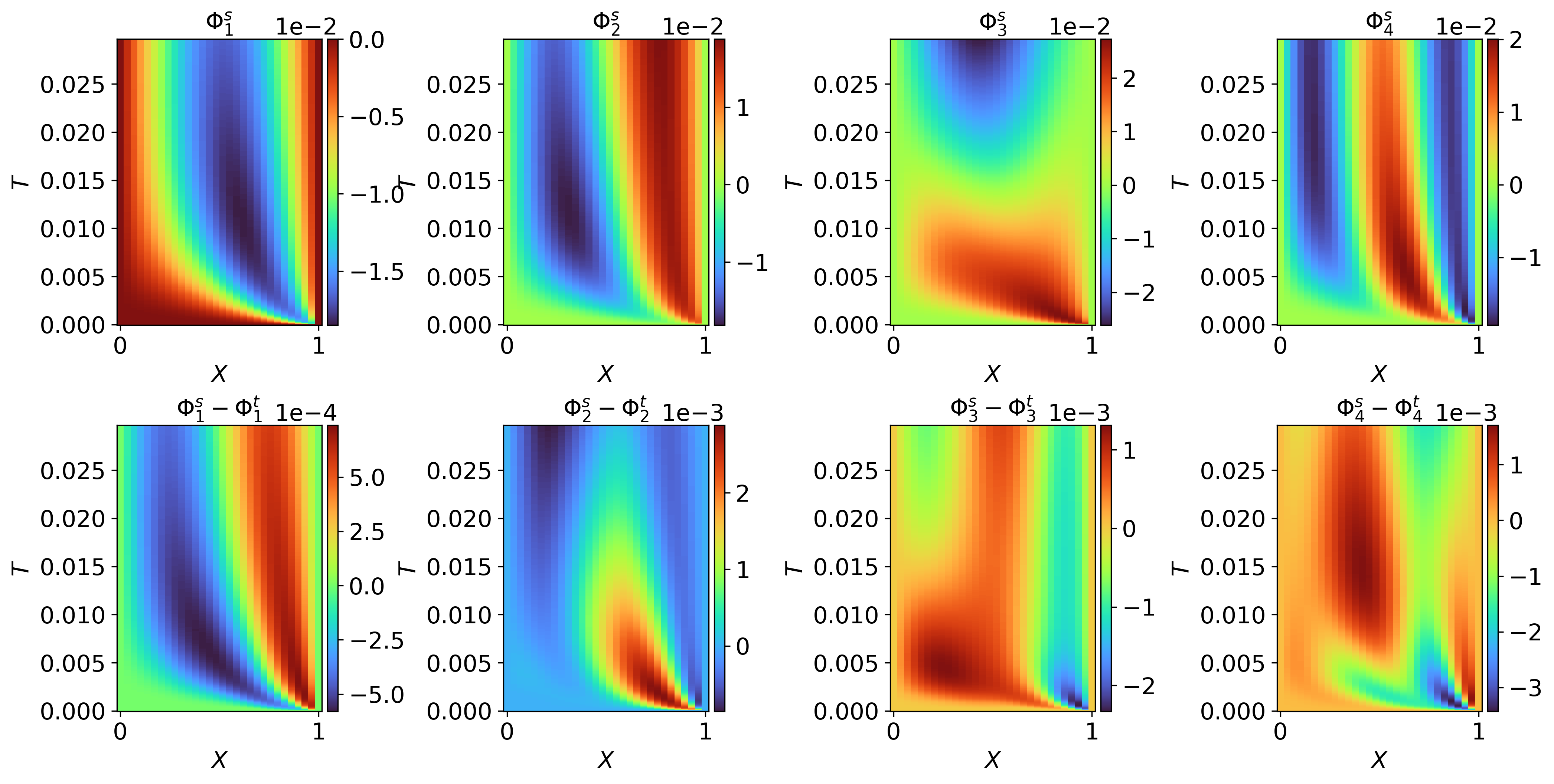}
    \caption{}
\end{subfigure}

\caption{Nonlinear diffusion equation: four leading eigenfunctions of $h$ for the source (top row) and target (bottom) problems computed from MCS for (a) $\sigma^2_y=0.1$, (b) $0.3$, and (c) $0.6$.}
\label{fig:eigenfunctions}
\end{figure}

Figure \ref{fig:mean_h} shows $\overline{h}^t$ computed as the ensemble mean from $1000$ MC samples and the mean-field equation under the target IBCs given for $\sigma^2_y=0.1$, 0.3, and 0.6. We observe a relatively close agreement between the MCS ensemble mean and the mean-field solution. The discrepancies between the two solutions increase with $\sigma^2_y$.  The relative $\ell_2$ distance between the mean-field MCS solutions is \num{1.23e-3} for $\sigma^2_y = 0.1$, \num{2.99e-03} for $\sigma^2_y = 0.3$, and \num{5.23e-03} for $\sigma^2_y = 0.6$.  
Next, we compare the source and target eigenfunctions in the $h$ KLD. It should be noted that orthonormal eigenfunctions are defined up to a sign, and we select source and target eigenfunctions such that they have the same sign. We find that $\phi^s_h(x,t)_i \approx \phi^t_h(x,t)_i$, and the difference between them increases only slightly with $\sigma^2_y$. Figure \ref{fig:eigenfunctions} presents the first four eigenfunctions of $h(x,t)$ for the source problem and the pointwise difference between source and target eigenfunctions for $\sigma^2_y = 0.1, 0.3$, and $0.6$. The absolute pointwise errors averaged over the mesh between source and target eigenfunctions are \num{1.03e-2}, \num{1.05e-2}, and \num{1.08e-2} for $\sigma^2_y = 0.1, 0.3$, and $0.6$, respectively. Based on the results in Figures \ref{fig:mean_h} and \ref{fig:eigenfunctions}, we conclude that the proposed single-shot learning of $\overline{h}^t$ and transferring the eigenfunction from the source problems provide a good approximation of the mean and eigenfunctions in the KLD of $h$ under target conditions.   

\begin{figure}[!htb]
    \centering
    \begin{subfigure}{\textwidth}
        \includegraphics[width=0.33\textwidth]{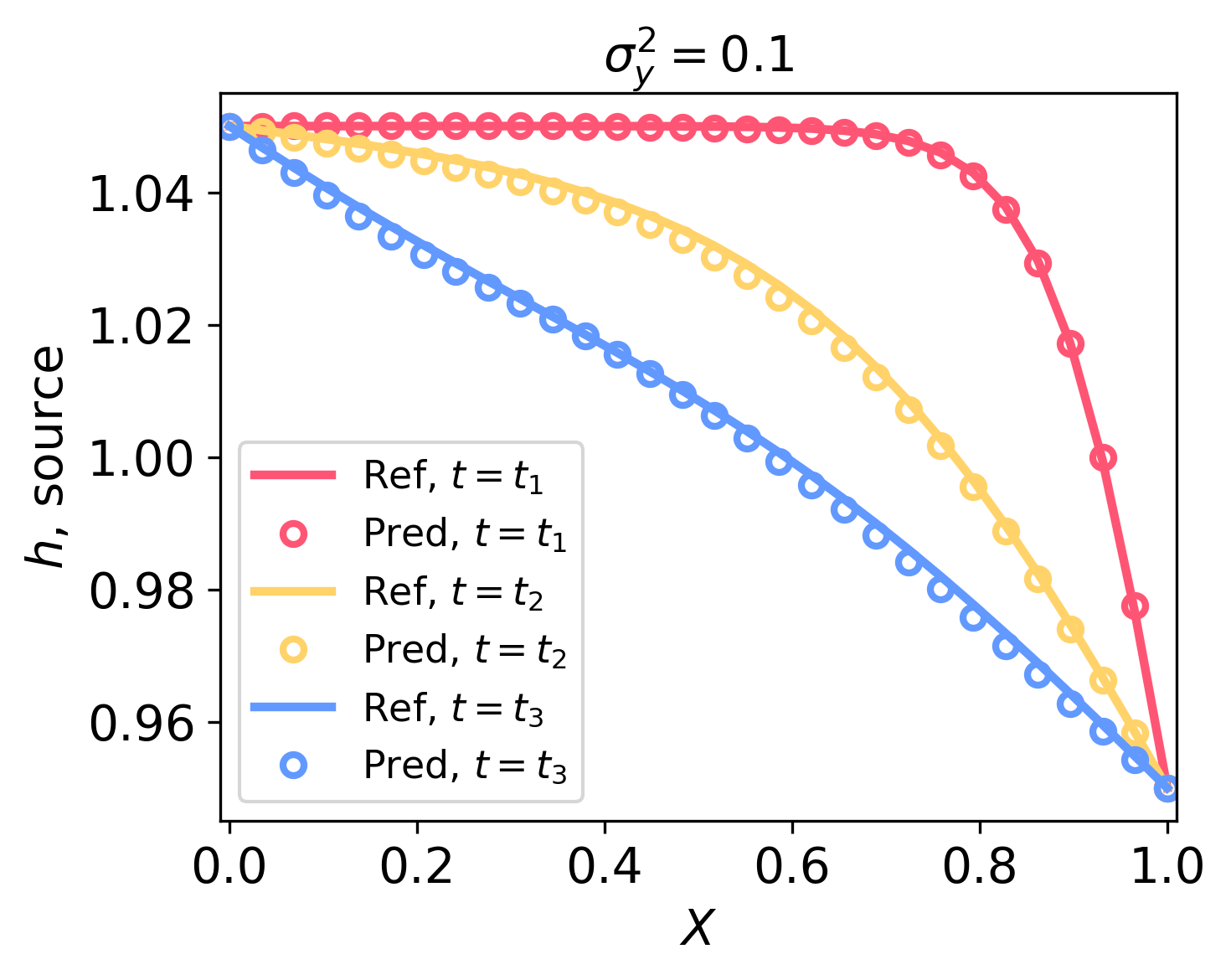}
        \includegraphics[width=0.33\textwidth]{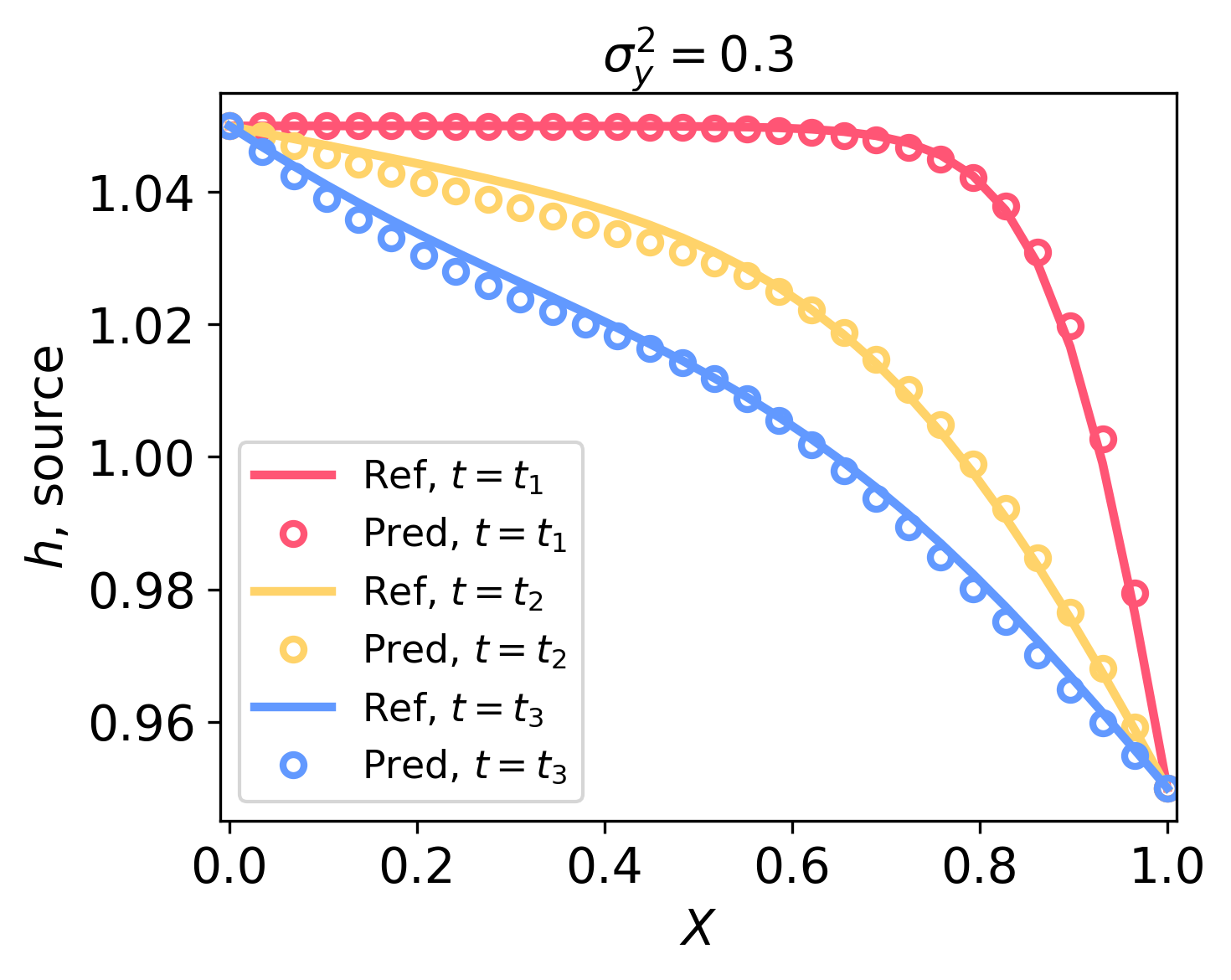}
        \includegraphics[width=0.33\textwidth]{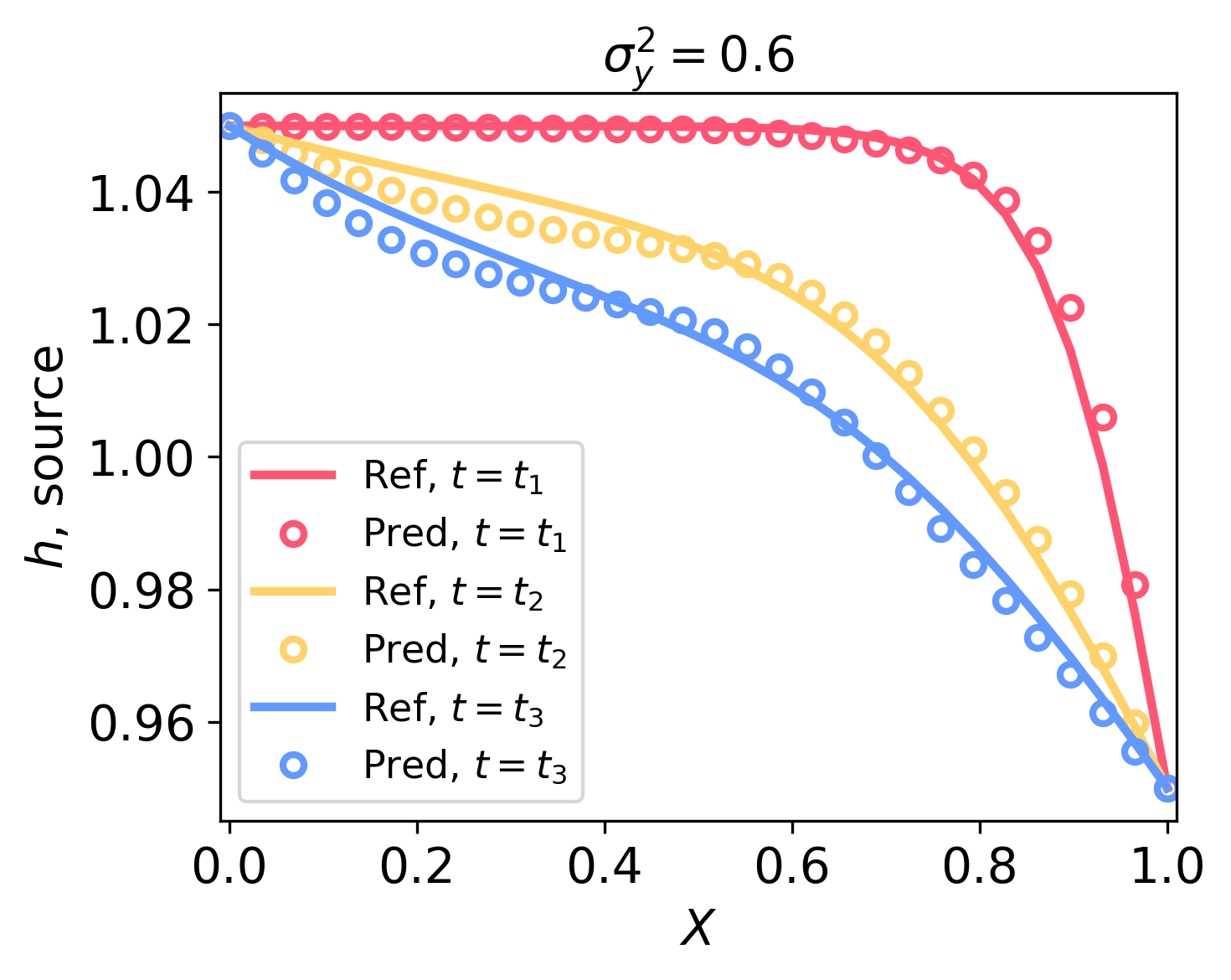}
        \caption{RLS}
    \end{subfigure}

    \begin{subfigure}{\textwidth}
        \includegraphics[width=0.33\textwidth]{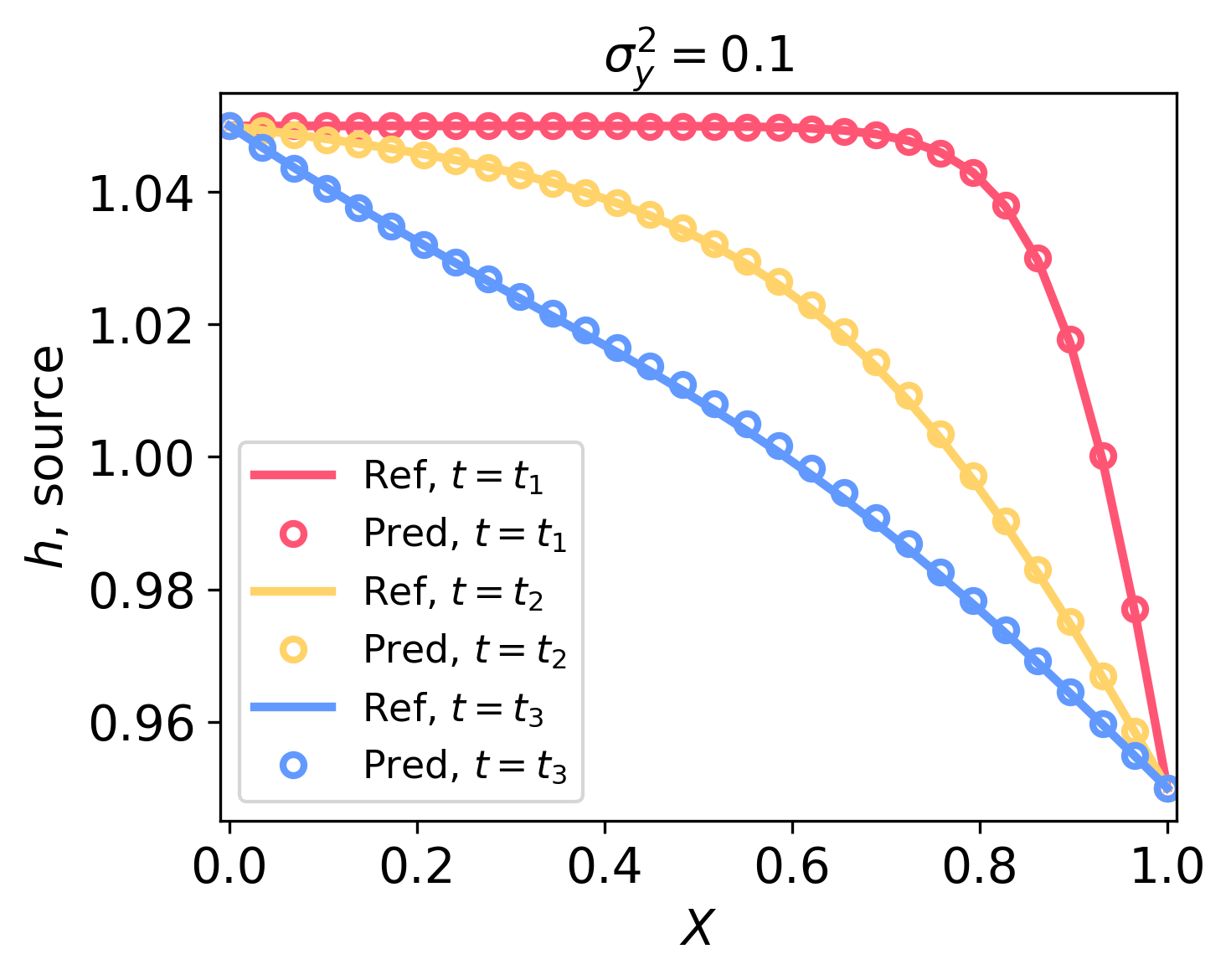}
        \includegraphics[width=0.33\textwidth]{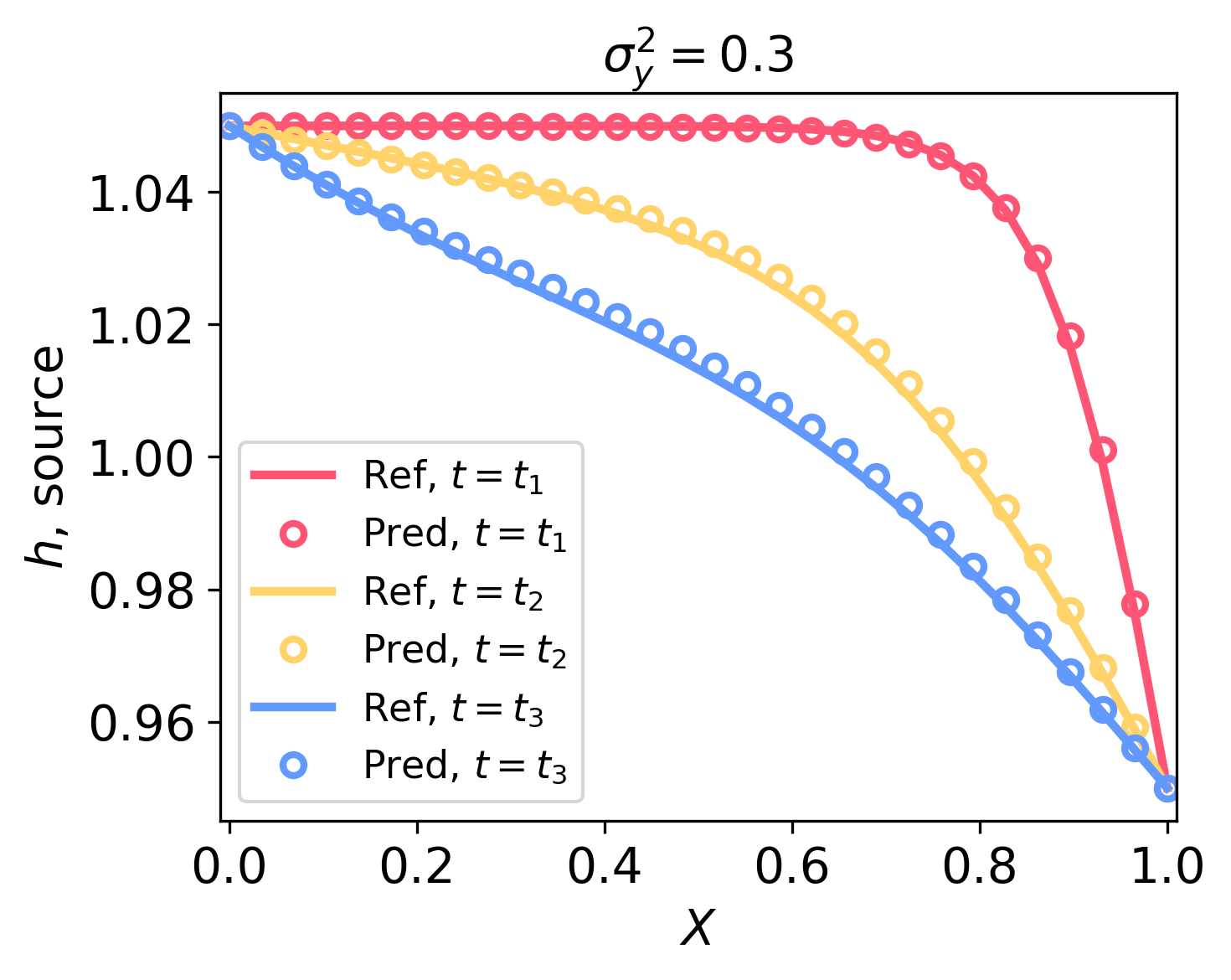}
        \includegraphics[width=0.33\textwidth]{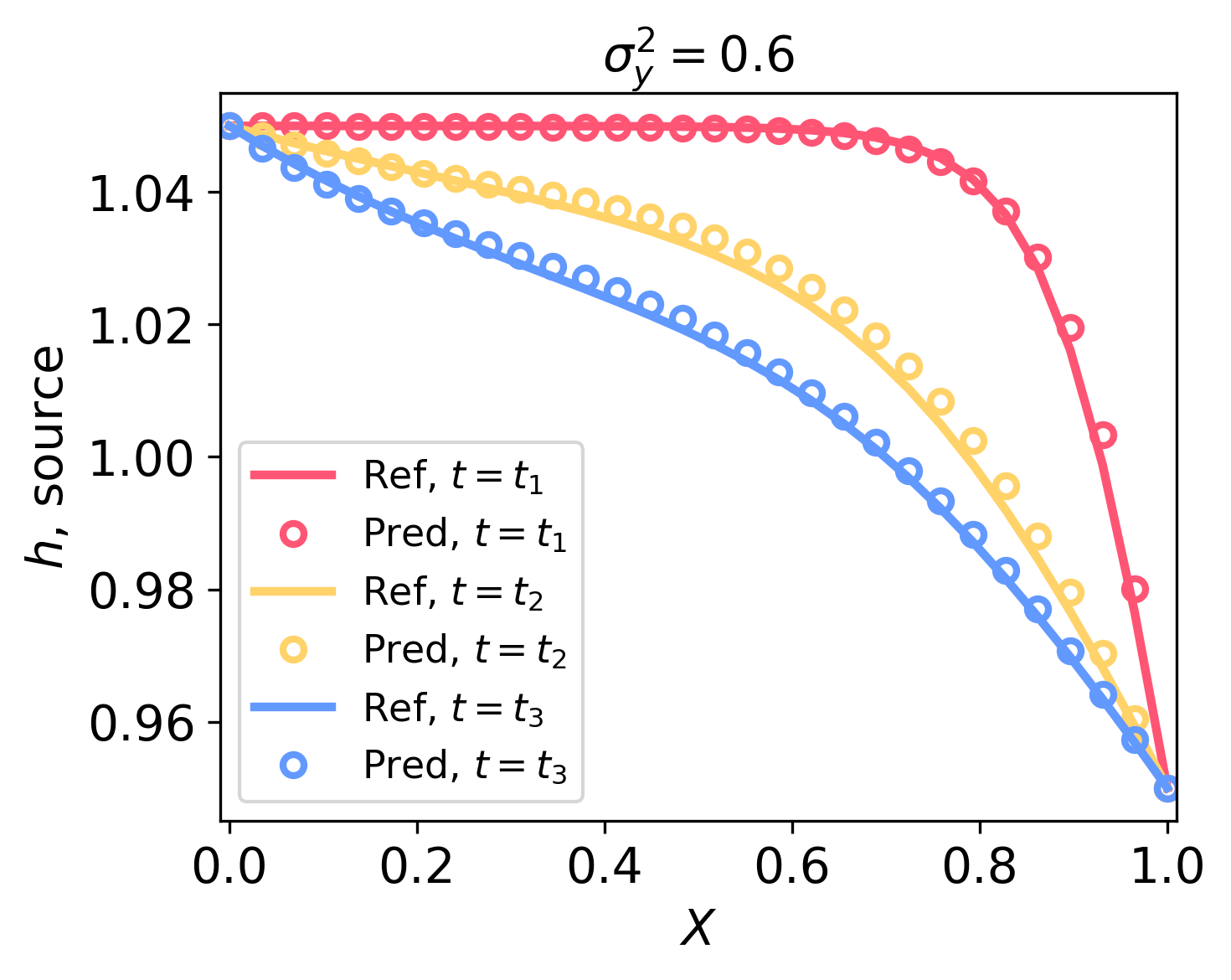}
        \caption{OLS}
    \end{subfigure}

    \begin{subfigure}{\textwidth}
        \includegraphics[width=0.33\textwidth]{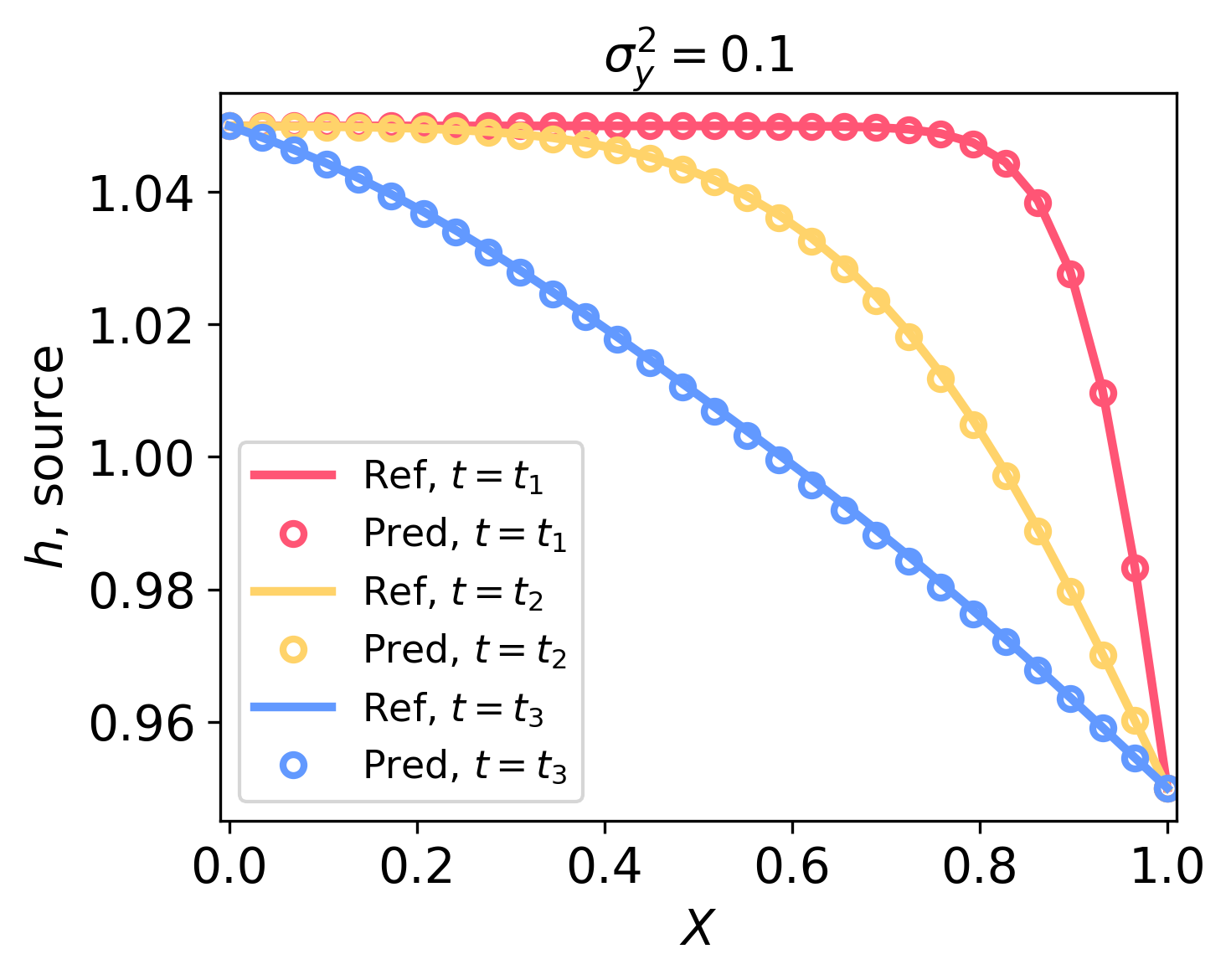}
        \includegraphics[width=0.33\textwidth]{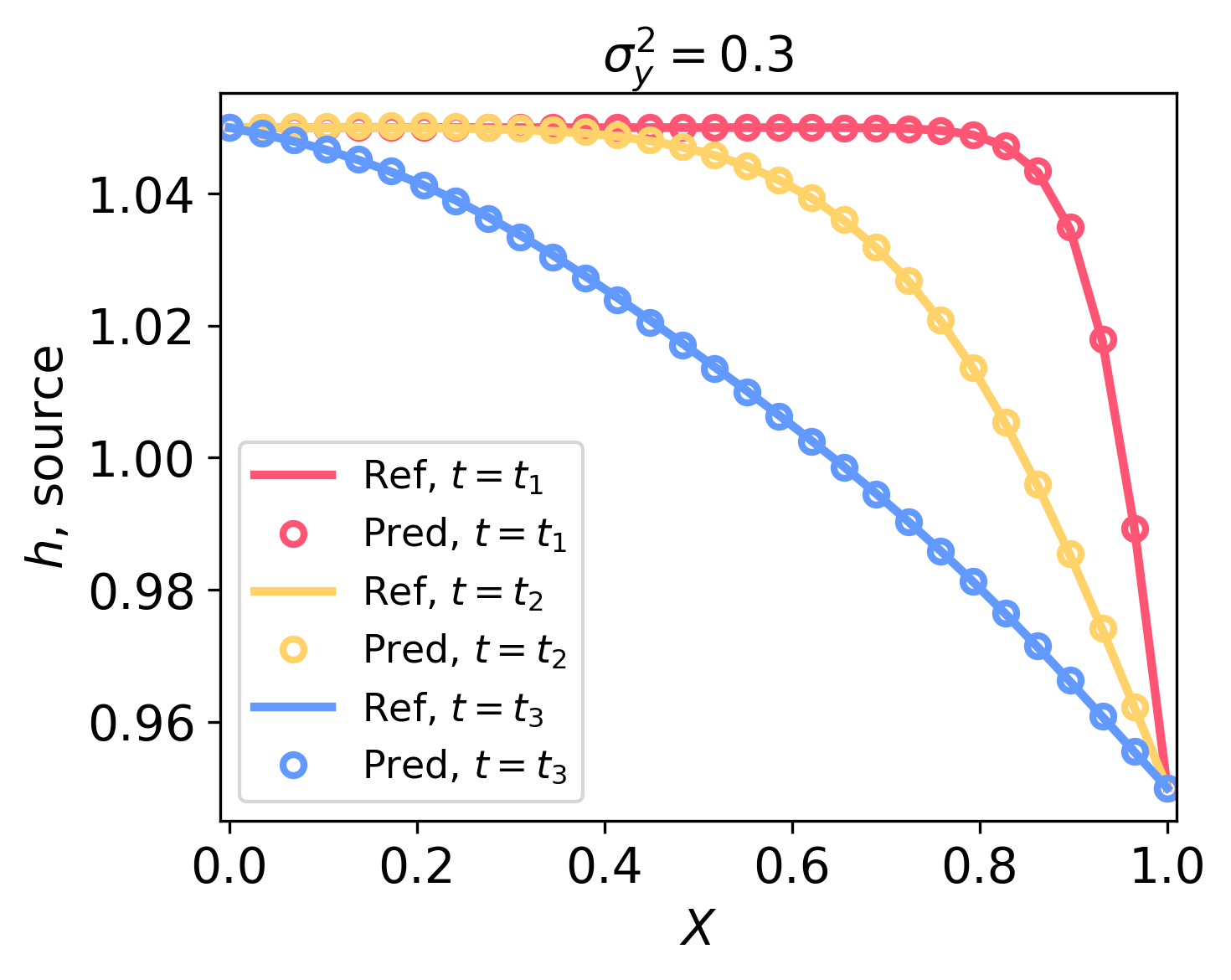}
        \includegraphics[width=0.33\textwidth]{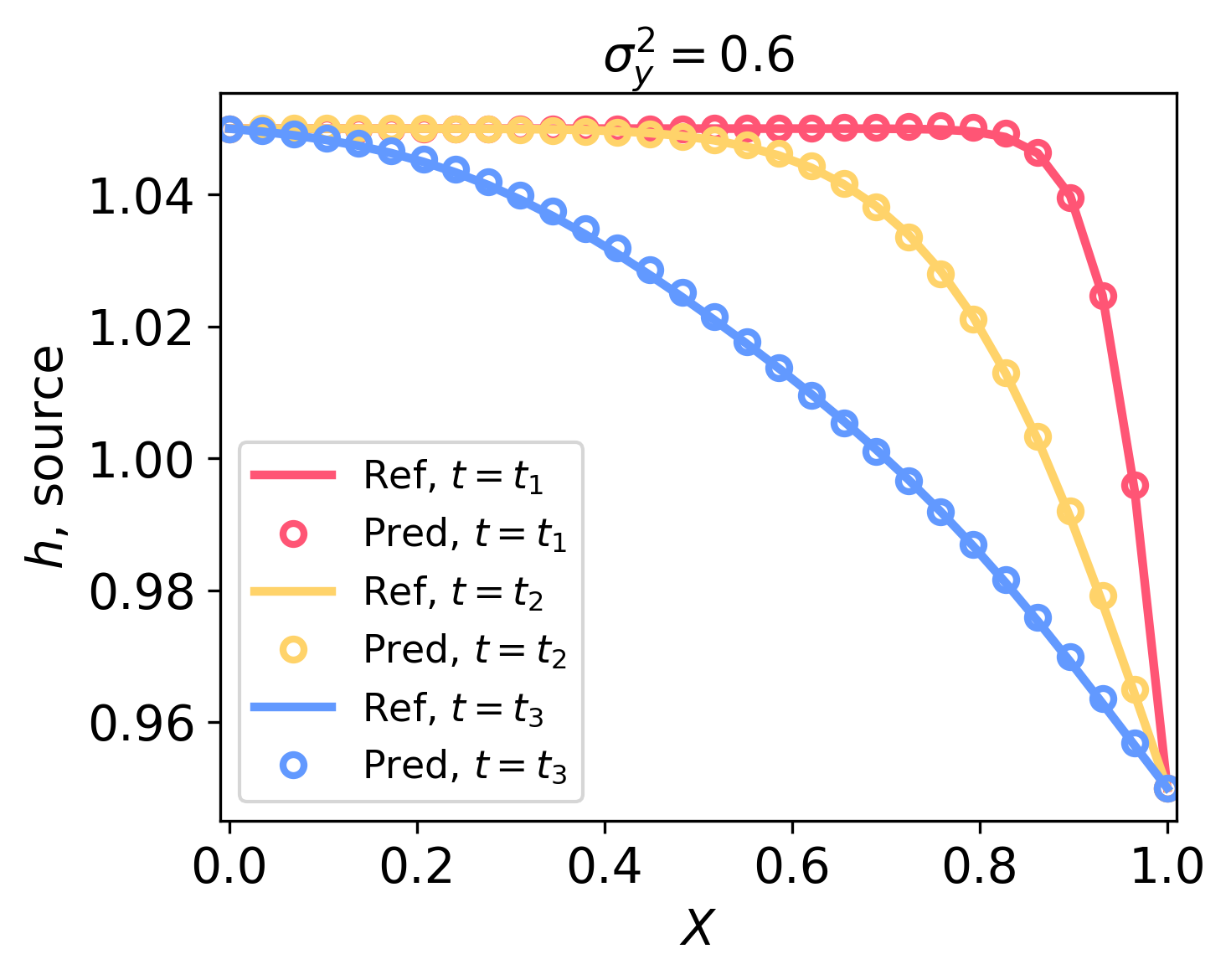}
        \caption{KL-NN}
    \end{subfigure}
    \caption{
    Nonlinear diffusion equation: \emph{source} solutions versus the reference at three different times ($t_1 = T/50, t_2 = T/5, t_3 = T$) for $\sigma^2_y=0.1$ (left column), 0.3 (middle column), and 0.6 (right column). In each subfigure, the (a) linear RLS, (b) OLS, and (c) nonlinear KL-DNN are compared. 
    }
    \label{fig:non_linear_solutions_source}
\end{figure}

\begin{figure}[!htb]
    \centering
    \begin{subfigure}{\textwidth}
        \includegraphics[width=0.33\textwidth]{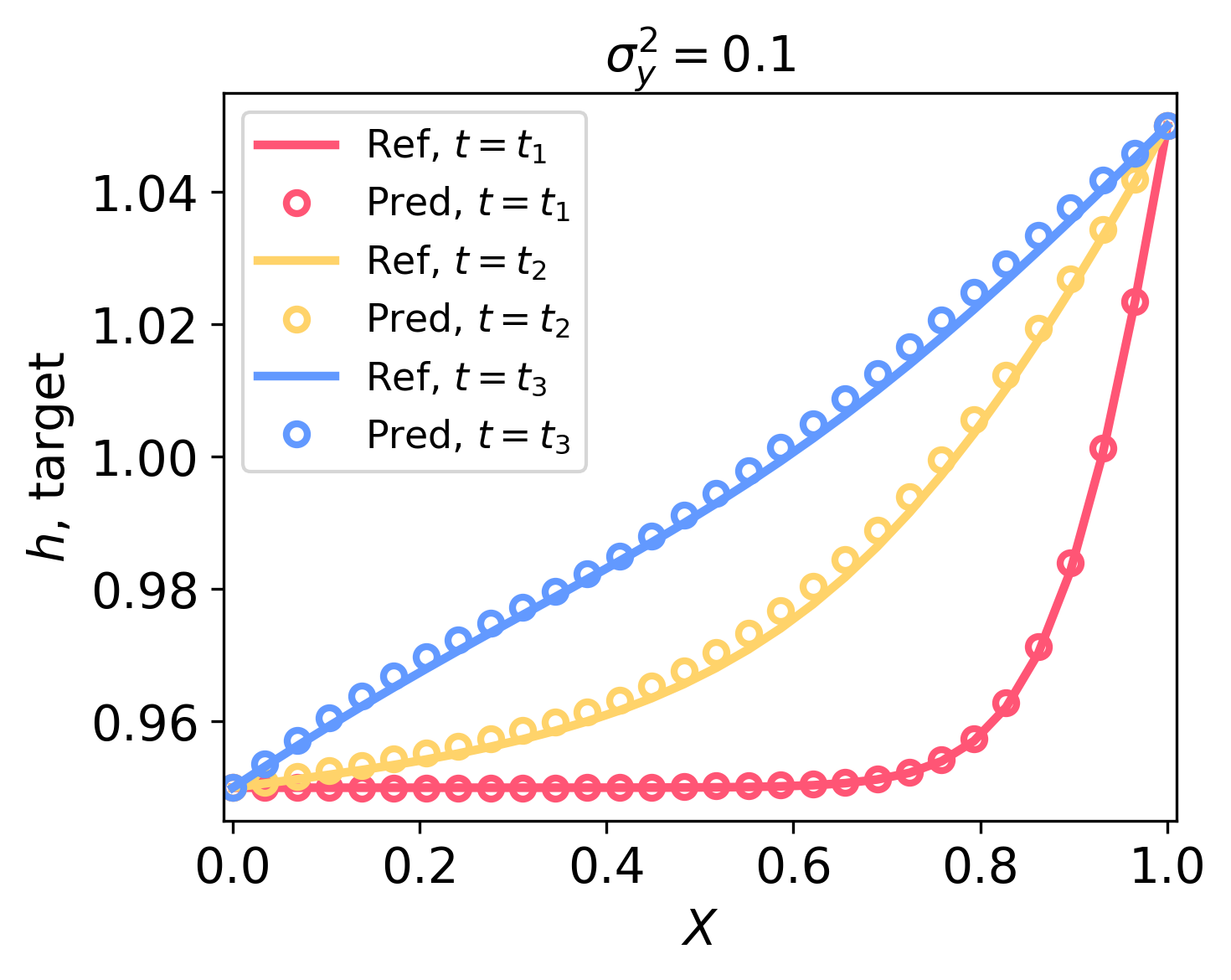}
        \includegraphics[width=0.33\textwidth]{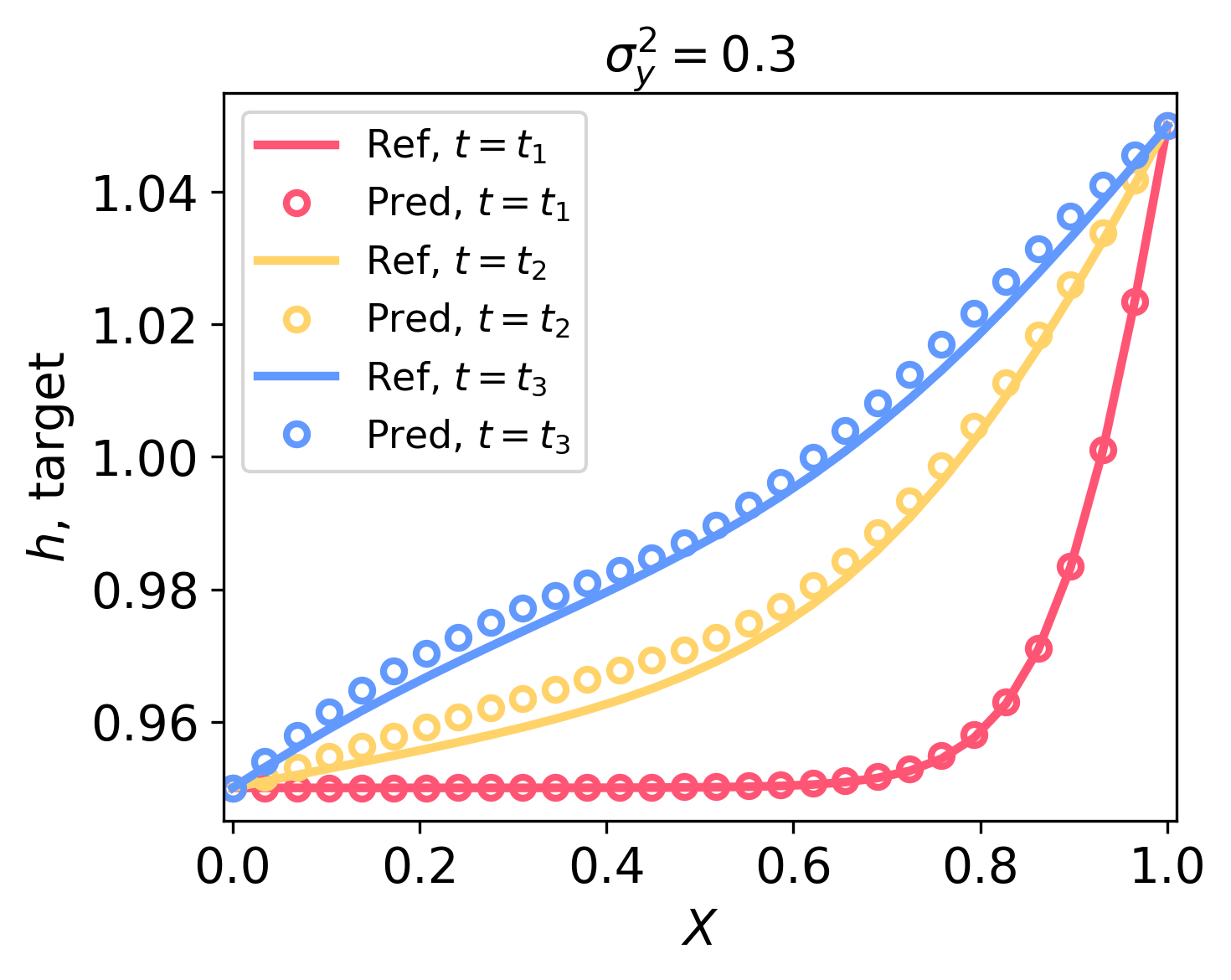}
        \includegraphics[width=0.33\textwidth]{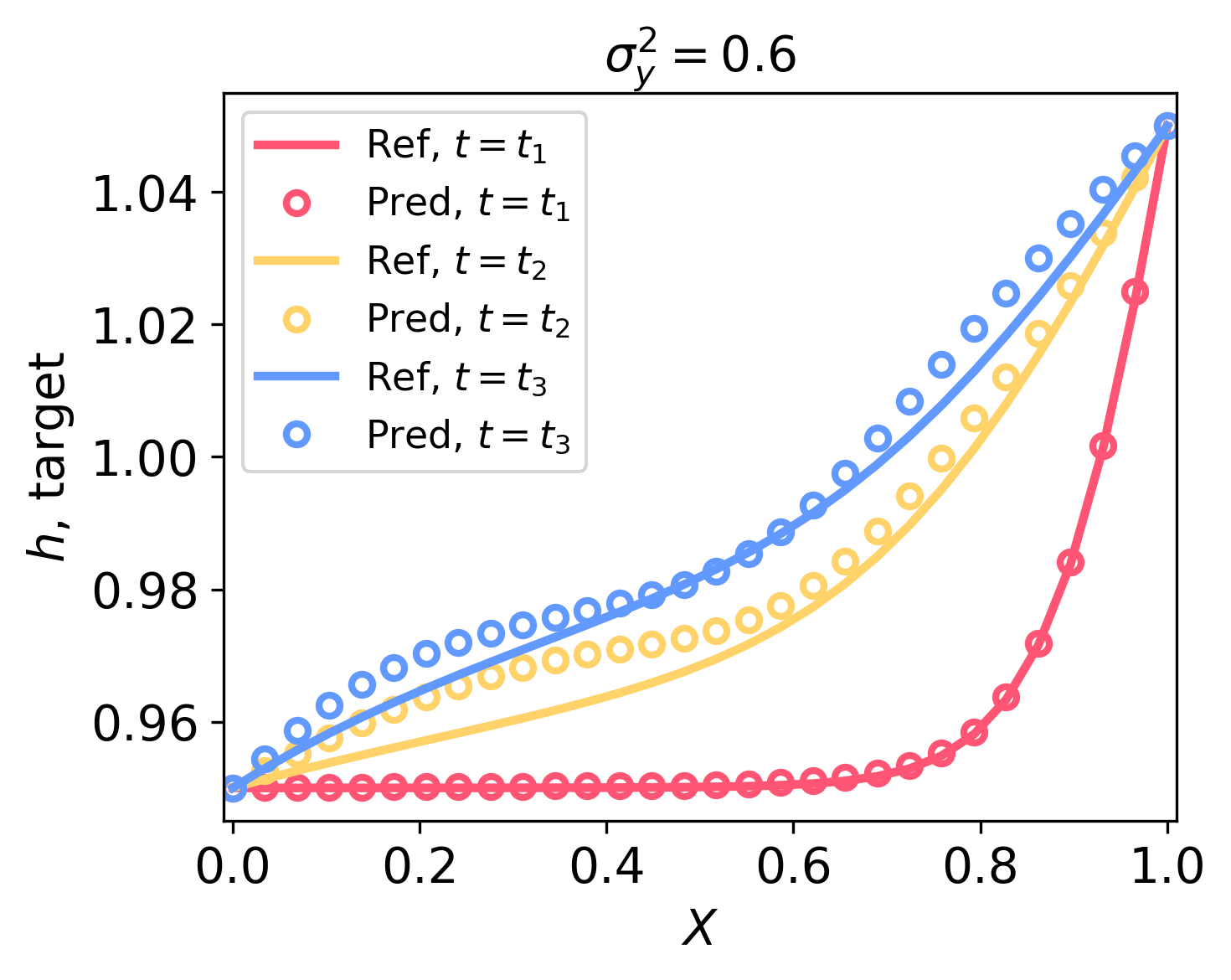}
        \caption{RLS}
    \end{subfigure}

    \begin{subfigure}{\textwidth}
        \includegraphics[width=0.33\textwidth]{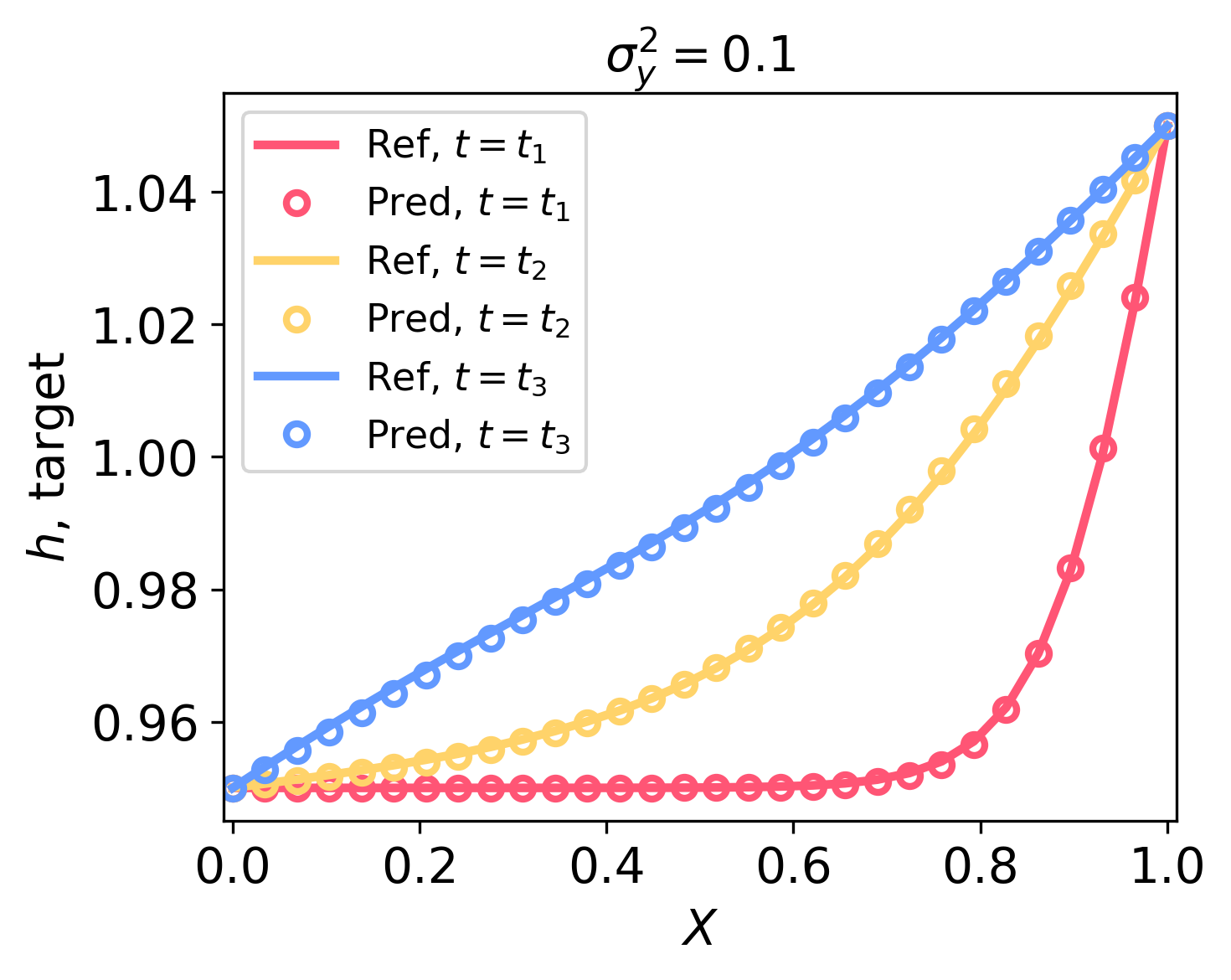}
        \includegraphics[width=0.33\textwidth]{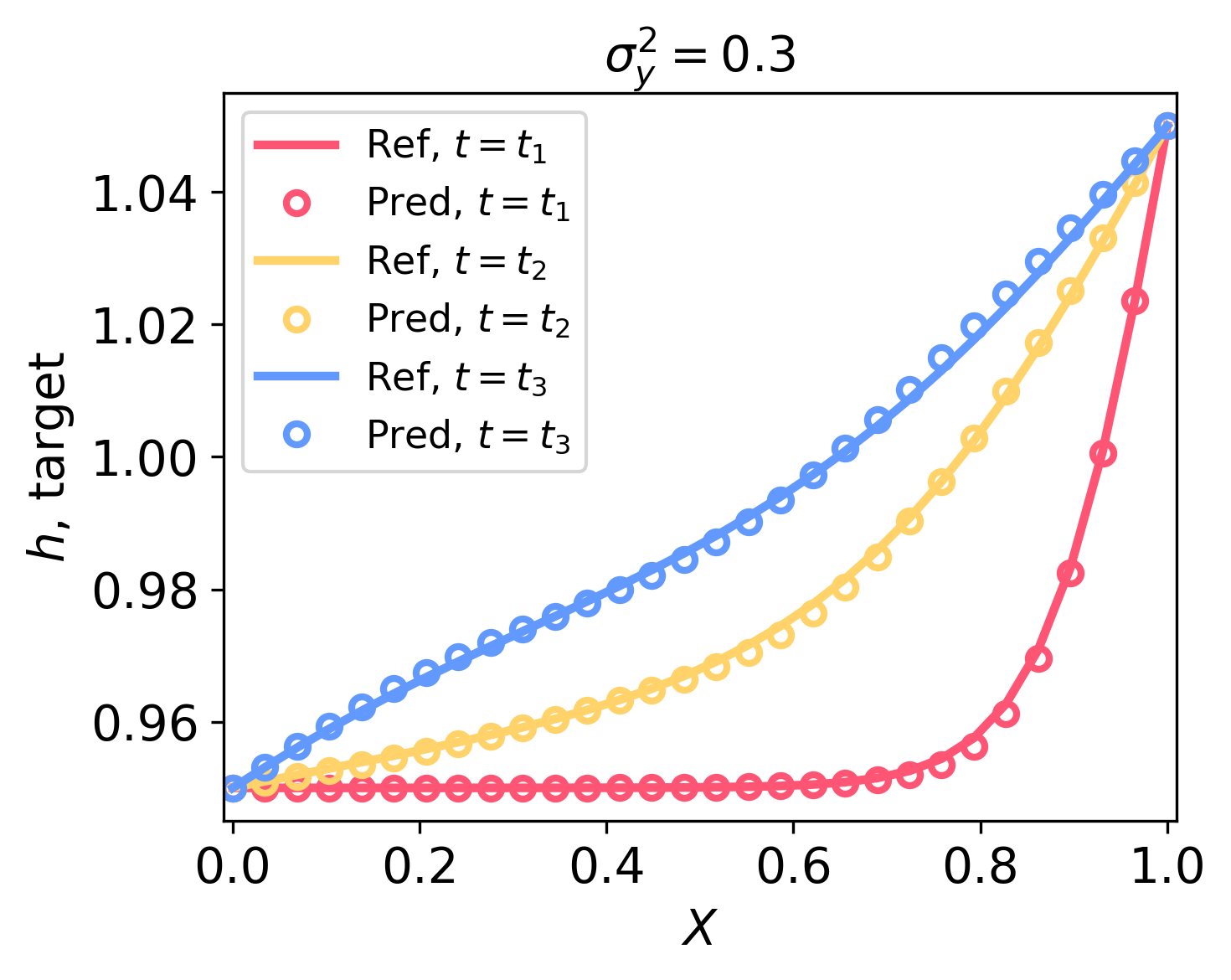}
        \includegraphics[width=0.33\textwidth]{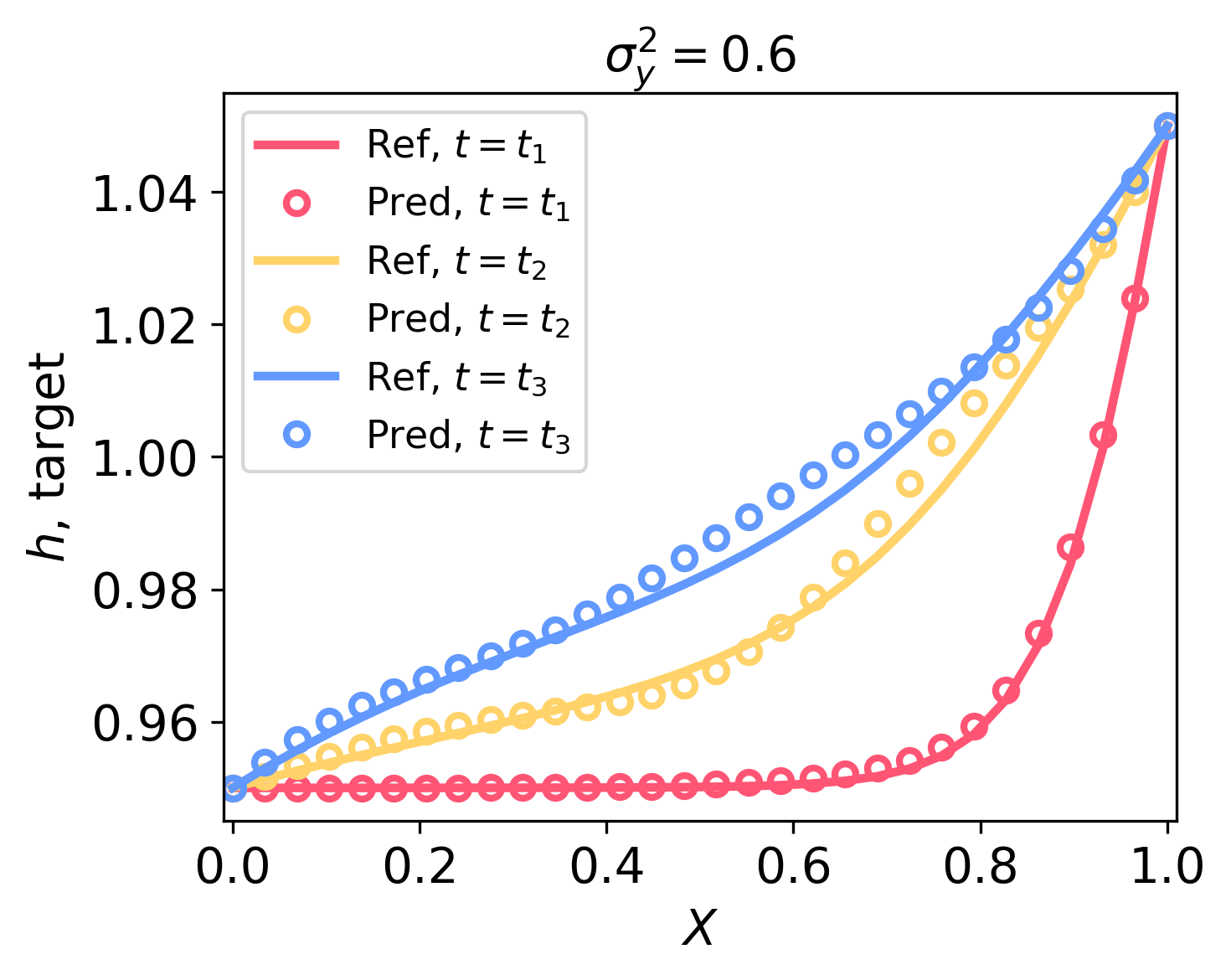}
        \caption{OLS}
    \end{subfigure}

    \begin{subfigure}{\textwidth}
        \includegraphics[width=0.33\textwidth]{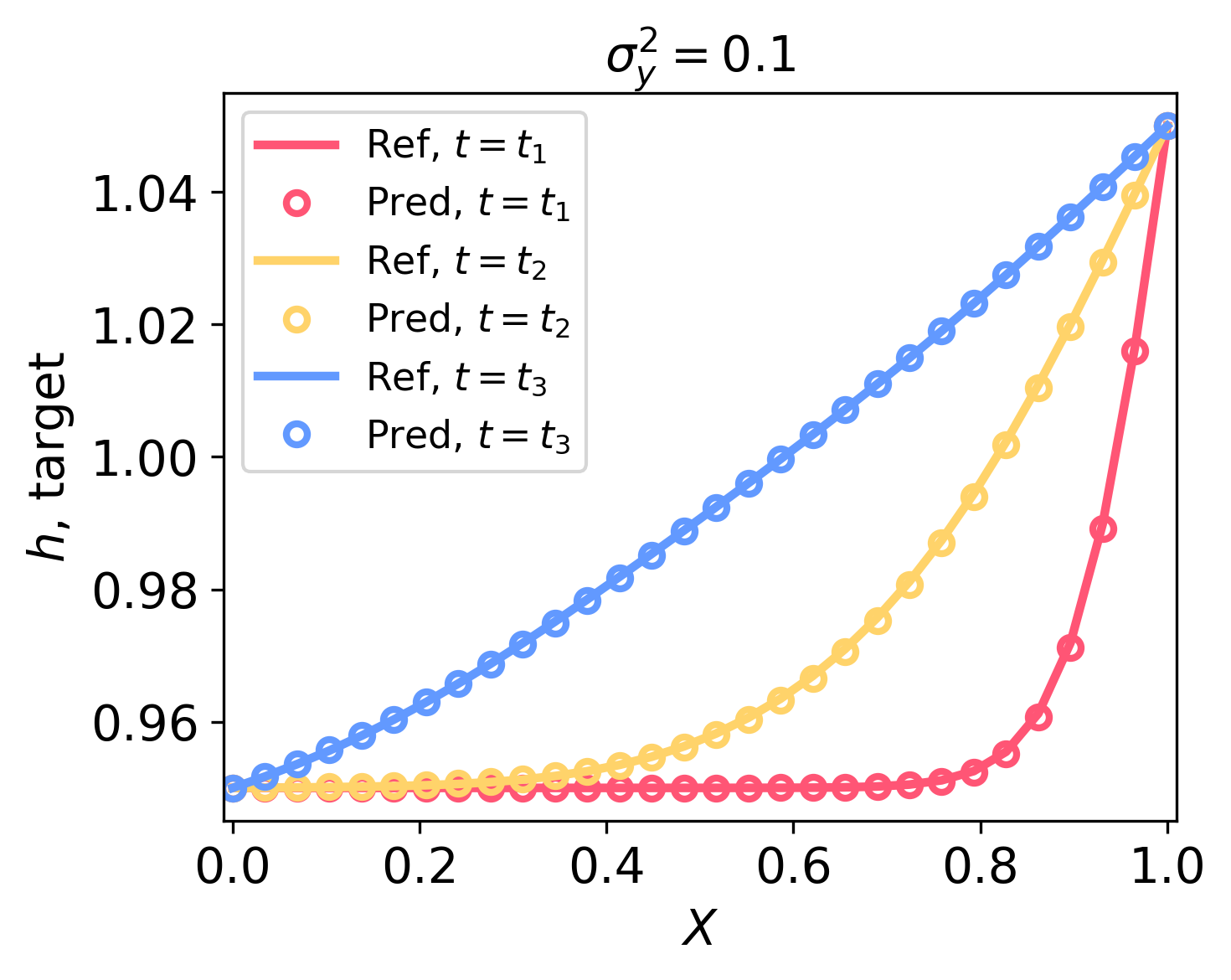}
        \includegraphics[width=0.33\textwidth]{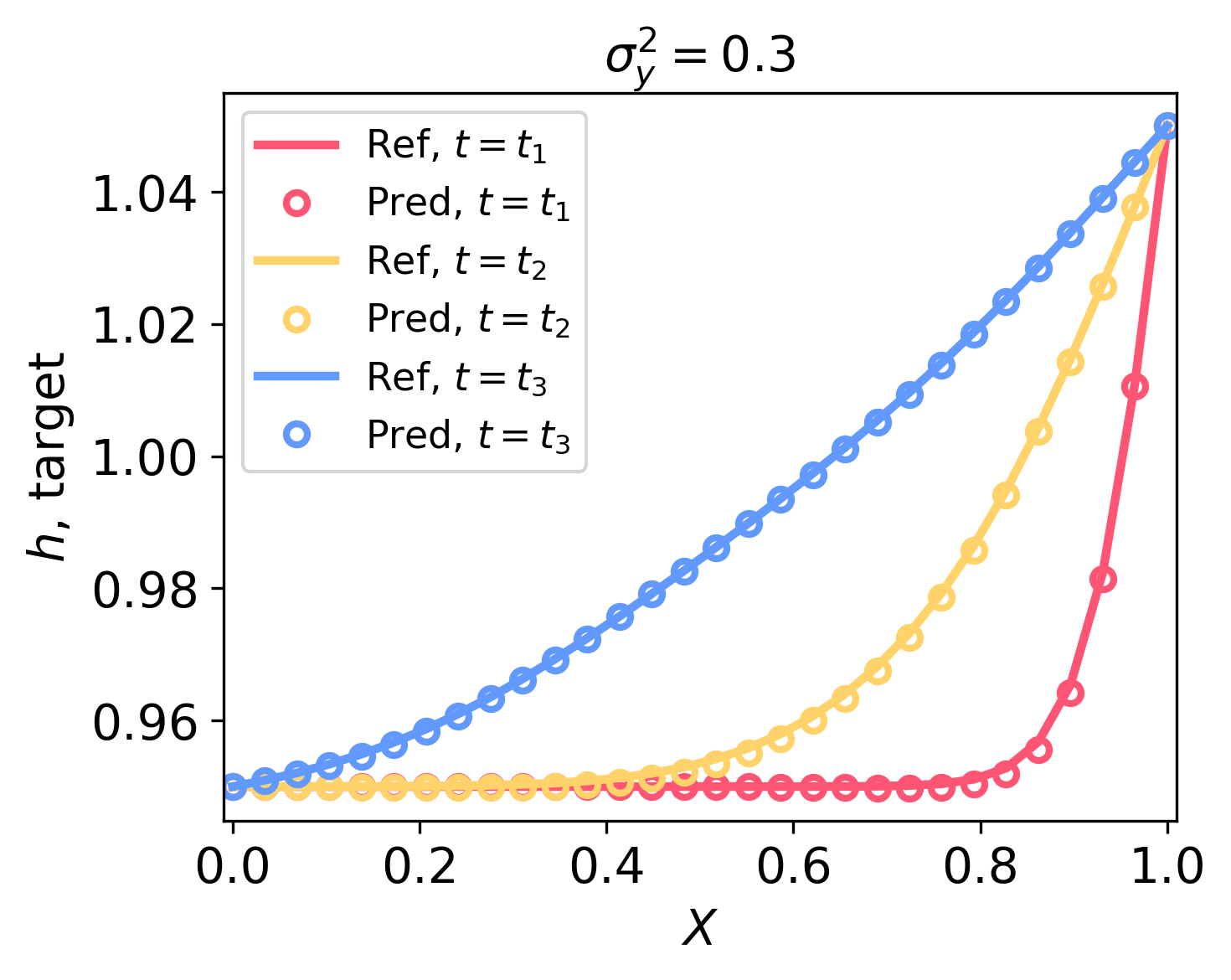}
        \includegraphics[width=0.33\textwidth]{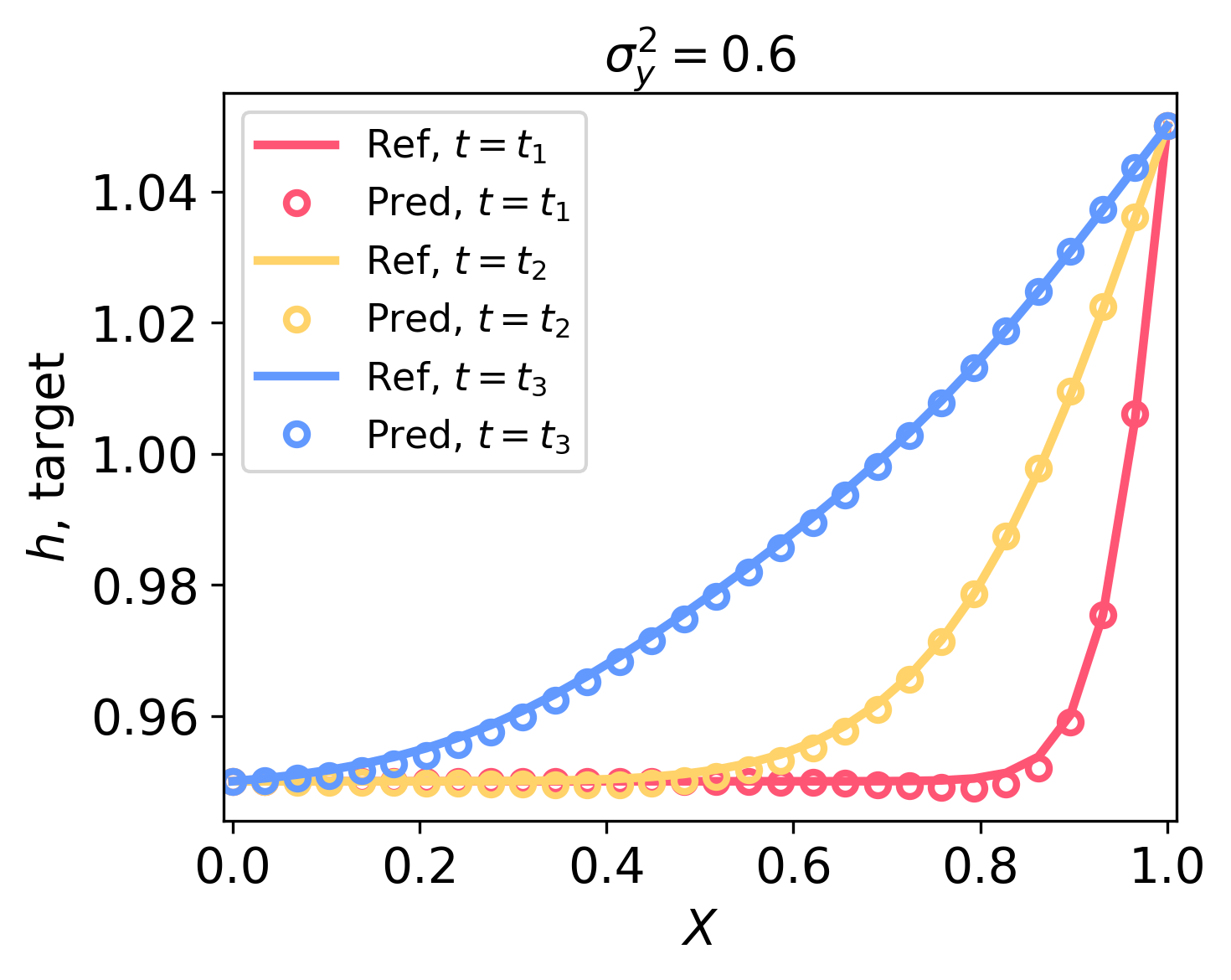}
        \caption{KL-NN}
    \end{subfigure}
    \caption{
    Nonlinear diffusion equation: \emph{target} solutions versus the reference at three different times ($t_1 = T/50, t_2 = T/5, t_3 = T$) for $\sigma^2_y=0.1$ (left column), 0.3 (middle column), and 0.6 (right column). In each subfigure, the (a) linear RLS, (b) OLS, and (c) nonlinear KL-DNN are compared. 
    }
    \label{fig:non_linear_solutions_target}
\end{figure}

Next, we study the accuracy of the proposed surrogate models for calculating the KLD coefficients $\bm\eta$. 
All tested surrogate models and the corresponding errors under source and target conditions are summarized in Table \ref{tab:l2_nonlinear}. 
Figures \ref{fig:non_linear_solutions_source} and \ref{fig:non_linear_solutions_target} compare solutions obtained with linear RLS, OLS, and the nonlinear KL-DNN methods against reference finite-difference solutions for $\sigma^2_y = 0.1, 0.3$, and $0.6$ under the source and target conditions. The linear RLS and OLS, and nonlinear KL-DNN, and PI-KL-DNN solutions are subject to statistical errors (i.e., the errors might be different for different realizations of $\bm\xi_k$ in the training dataset and DNN initialization). Therefore, we obtain 20 solutions for each source and boundary conditions for randomly generated $k$ fields and report average errors. All samples from the source dataset are used to compute $\bm{W}^s$ and $\bm \theta^s$ in the OLS and KL-DNN methods, respectively. For the target conditions, we analyze the $\varepsilon$ error for $N_{\text{train}}^t = 5$, 20, and 80. For the PI-KL-DNN, we use $N_r = 250$ random realizations of $\bm\xi_k$. The weight $\lambda_r = \num{1e-4}$ in the loss function \eqref{eq:pi_loss_target} is chosen based on a grid search over the range $[\num{1e-5}, \num{1e1}]$. 

\begin{table}[!htb]
\centering
\renewcommand{\arraystretch}{1.25}
\caption{Nonlinear diffusion equation: the KL-NN surrogate model errors $\varepsilon$ for $\sigma_y^2=0.1$, $0.3$, and $0.6$ under source and target conditions. Source and target solutions are estimated with the linear RLS and OLS, and nonlinear KL-DNN and PI-KL-DNN models.
}
\begin{tabular}{cccS[table-format=1.2e-1]S[table-format=1.2e-1]S[table-format=1.2e-1]}
    \toprule
    \multicolumn{2}{c}{Method} & & \textbf{$\sigma^2_y = 0.1$} & \textbf{$\sigma^2_y = 0.3$} & \textbf{$\sigma^2_y = 0.6$} \\
    \hline
    \multirow{6}{*}{Linear} &
    \multirow{2}{*}{RLS} 
    & Source & 1.49e-03 & 2.23e-03 & 2.83e-03 \\
    & & Target & 2.20e-03 & 3.96e-03 & 5.75e-03  \\
    \cline{2-6}
    & \multirow{4}{*}{OLS} 
    & Source & 4.82e-04 & 1.13e-03 & 1.72e-03 \\
    & & Target, $N^{\text{t}}_{\text{train}} = 5$ & 2.86e-03 & 6.95e-03 & 1.32e-02 \\
    % & Target, $N^{\text{t}} = 10$ & 3.22e-03 & 7.76e-03 &  1.17e-02 \\
    & & Target, $N^{\text{t}}_{\text{train}} = 20$ & 2.54e-03 & 7.27e-03 &  1.23e-02 \\
    & & Target, $N^{\text{t}}_{\text{train}} = 80$ & 9.59e-04 & 3.09e-03 &  7.02e-03 \\
    \hline
    \multirow{8}{*}{Nonlinear} &
    \multirow{4}{*}{KL-DNN} 
    & Source & 3.21e-04 & 8.20e-04 & 1.46e-03   \\
    & & Target, $N^{\text{t}}_{\text{train}} = 5$ & 3.92e-03 & 6.98e-03 & 9.09e-03  \\
    % & Target, $N^{\text{t}} = 10$ & 1.81e-03 & 2.64e-03 & 5.66e-03 \\
    & & Target, $N^{\text{t}}_{\text{train}} = 20$ & 9.82e-04 & 2.53e-03 & 2.74e-03 \\
    & & Target, $N^{\text{t}}_{\text{train}} = 80$ & 4.38e-04 & 1.07e-03 &  1.44e-03 \\
    \cline{2-6}
    & \multirow{4}{*}{PI-KL-DNN} 
    & Target, $N^{\text{t}}_{\text{train}} = 0$ & 3.40e-04 & 2.28e-03 &  7.91e-03 \\
    & & Target, $N^{\text{t}}_{\text{train}} = 5$ & 3.32e-04 & 2.26e-03 & 6.78e-03  \\
    & & Target, $N^{\text{t}}_{\text{train}} = 20$ & 4.23e-04 & 1.64e-03 & 3.80e-03 \\
    & & Target, $N^{\text{t}}_{\text{train}} = 80$ & 2.74e-04 &  6.31e-04&  1.29e-03 \\
    \bottomrule
\end{tabular}
\label{tab:l2_nonlinear}
\end{table}

For the RLS formulation, the errors in both source and target solutions increase with $\sigma^2_y$. The accuracy of TL decreases as $\sigma^2_y$ increases; the source-to-target error ratio increases from 0.67 for $\sigma^2_y=0.1$ to 0.49 for $\sigma^2_y=0.6$. However, for all considered $\sigma^2_y$, the $\varepsilon$ errors in the source and target solutions stay below $1\%$. 
The linear OLS model is more accurate than the RLS model assuming that a sufficient number of samples are available to compute  $\bm{W}$, and the required number of samples increases with  $\sigma^2_y$. For example, for  $\sigma^2_y \le 0.4$, at least 80 target samples are required to obtain an accurate target OSL solution. However, for $\sigma^2_y = 0.6$, the OLS target solution obtained with 80 samples remains less precise than the RLS solution. 
For the source conditions (where 1000 samples are available), the errors in the OLS solutions are as much as $70\%$ smaller than those in the RLS solutions. 

For the nonlinear KL-DNN and PI-KL-DNN models, the DNN parameters $\bm\theta^s = {\bm W}^s_{1:N+1}, {\bm b}^s_{1:N+1} \}$ are trained jointly using the source dataset with 1000 samples. In the target problem,  parameters in the last linear layer ${\bm W}_{N+1}, {\bm b}_{N+1}$ are retrained using the data-driven method (KL-DNN) for $N_{\text{train}}^t = 5$ to 80 and using the physics-informed method (PI-KL-DNN) for $N_{\text{train}}^t = 0$ to 80 with $N_r = 250$ random realizations. The DNN parameters in other layers $\tilde{\bm\theta}$ are transferred from the source problem. The DNN mapping reduces the source and target solution errors for all $\sigma^2_y$ and $N_{\text{train}}^t$. The PI-KL-DNN method further reduces target solution errors, particularly for small $N_{\text{train}}^t$. Even with $N_{\text{train}}^t = 0$ (one-shot learning), the PI-KL-DNN method achieves target solution errors comparable to the source DNN solutions.

In summary, an accurate one-shot TL is possible for the nonlinear diffusion problem for $\sigma^2_y<1$. Under this condition, the $h$ eigenfunctions can be transferred from the source condition and $\overline{h}$ can be accurately estimated from the closed-form mean-field equation, disregarding the nonlinear term involving $h'$ and $k'$. However, the nonlinearity with respect to $h'$ and $k'$ must be maintained to achieve an accurate TL solution for the target problem, i.e., the nonlinear PI-KL-DNN model $N_{\text{train}}^t=0$ produces significantly lower errors than the linear RLS model.
 Assimilating the labeled target data can further reduce the errors in the TL model, especially for larger $\sigma^2_y$.

%This indicates that when the variance of $y$ is small, the relationship between $\eta$ and $\xi$ is approximately linear. The assumptions that the eigenfunctions can be safely transferred and that the weight matrix $\bm{W}$ can be accurately approximated using $\overline{h}$ and the RLS formulation hold.

\section{Discussion and Conclusions}\label{sec:conclusion}
We analyze the TL of surrogate models as a foundation for constructing an efficient DT. Due to computational cost considerations, surrogate models are trained for specific ranges of control variables and conditions (e.g., IBCs). These ranges can be defined by probability distributions based on prior knowledge and may change over time.  Therefore, an efficient DT should be able to retrain the surrogate model for any distribution of the control variables. 

In this work, we used the KL-NN surrogate model because the KLD representation of the control and state variables in this model is closely related to KLEs of stochastic processes. The parameters in the KL-NN model include the ensemble mean, scaled eigenfunctions, and the NN parameters, where the NN model (shallow linear or deep nonlinear) is constructed by mapping the KLD coefficients of control parameters to those of the state variables. We applied the moment equation approach to formulate and analyze a one-shot TL procedure for the KL-NN surrogate model for the linear and nonlinear diffusion PDEs. 

For linear problems, the ensemble mean of the state variables can be exactly computed from the closed-form mean moment equation. The eigenfunctions and NN parameters are independent of the mean values of the state variables and weakly dependent on the covariance of the control variables. Therefore, the eigenfunction and NN parameter transfer from the source to the target problem does not introduce errors in the surrogate model if the covariance function of the control variables is fixed. 
This TL amounts to one-shot learning, where the single sample, corresponding to the target IBCs and the mean of the target distribution of the control variables, is required to retrain the model for a target condition. 

One-shot TL learning becomes approximate if the covariance of the control variables changes from the source to target conditions because the eigefunctions of $h$ depend on the covariance of the control variables. However, this dependence is weak and the one-shot TL remains accurate as long as the target KLD can accurately represent the target control variable. 
We found that this condition is satisfied except when the target correlation length is much smaller than the source correlation length.  Therefore, when there is uncertainty about the distribution of the control variables or when the distribution is expected to change over time, the source dataset should be developed for the smallest expected correlation length. The correlation length defines the number of terms in the KLE and KLD needed to accurately represent the corresponding stochastic process.

For nonlinear PDEs, all KL-NN parameters depend on the source/target conditions. We analyzed conditions when some of these parameters can be transferred without introducing significant errors for the nonlinear diffusion equation problem describing flow in heterogeneous porous media with the space-dependent diffusion coefficient (hydraulic conductivity) chosen as a control variable. We demonstrated that the eigenfunctions of the state variable can be transferred, and the mean state variable can be estimated from the mean-field equation if the variance of the log diffusion coefficient is less than one and if the changes in the prescribed head boundary conditions from the source to target conditions are smaller than the domain size.  The relationship between the parameters in the KLD expansions of the state and control variables is nonlinear, and, in general, a DNN is needed to approximate this relationship accurately. In the one-shot TL for this diffusion PDE problem, the DNN is trained using the source dataset. For the target problem, only the parameters in the last linear layer are retrained by minimizing the PDE residuals (the PI-KL-DNN), which does not require any labeled target data. We found that the one-shot TL produces errors of the same order of magnitude as in the source solution. The PI-KL-DNN method requires numerically computing spatial derivatives in the PDE residuals. Our results also show the weights in the last linear DNN layer can be accurately estimated from a relatively small labeled target dataset with $N_{\text{train}}^t =  N_k/4$ samples by solving a data-driven linear least-square problem in an efficient few-shot learning procedure.  

Our main conclusions are that the one-shot TL is exact for linear models and is a good approximation for nonlinear problems where the variance of the multiplicative control variables is small. Such control variables usually describe the system properties. Therefore, accurate Bayesian parameter estimation (which reduces the parameter uncertainty) is critical for developing an efficient DT, which can be easily retrained for changing conditions. Also, a DT design often requires a compromise between the need to include the parameters and terms defining system behavior in the control variable list and the computational cost of generating and training the resulting DT, which typically increases with the number of control variables. Our analysis implies that, when possible, the terms introducing nonlinearity and varying in a large range should be excluded from the list of control variables. Instead, the DT could be retrained for different values of such terms using accurate (or exact) one-short learning. Future work could explore the proposed framework application to systems described by equations nonlinear in state variables.

\section{Acknowledgements}
This research was partially supported by the U.S. Department of Energy (DOE) Advanced Scientific
Computing Research program, the United States Geological Survey, 
and the United States National Science Foundation. Pacific Northwest National Laboratory is
operated by Battelle for the DOE under Contract DE-AC05-76RL01830.

\bibliographystyle{unsrt}  
\bibliography{DT_and_TL.bib} 

\begin{thebibliography}{10}

\bibitem{wagg2020digital}
DJ~Wagg, Keith Worden, RJ~Barthorpe, and Paul Gardner.
\newblock Digital twins: state-of-the-art and future directions for modeling
  and simulation in engineering dynamics applications.
\newblock {\em ASCE-ASME Journal of Risk and Uncertainty in Engineering
  Systems, Part B: Mechanical Engineering}, 6(3):030901, 2020.

\bibitem{kaur2020convergence}
Maninder~Jeet Kaur, Ved~P Mishra, and Piyush Maheshwari.
\newblock The convergence of digital twin, iot, and machine learning:
  transforming data into action.
\newblock {\em Digital twin technologies and smart cities}, pages 3--17, 2020.

\bibitem{ritto2021digital}
TG~Ritto and FA~Rochinha.
\newblock Digital twin, physics-based model, and machine learning applied to
  damage detection in structures.
\newblock {\em Mechanical Systems and Signal Processing}, 155:107614, 2021.

\bibitem{kapteyn2020toward}
Michael~G Kapteyn, David~J Knezevic, and Karen Willcox.
\newblock Toward predictive digital twins via component-based reduced-order
  models and interpretable machine learning.
\newblock In {\em AIAA scitech 2020 forum}, page 0418, 2020.

\bibitem{morlot2024hydrological}
Martin Morlot, Riccardo Rigon, and Giuseppe Formetta.
\newblock Hydrological digital twin model of a large anthropized italian alpine
  catchment: The adige river basin.
\newblock {\em Journal of Hydrology}, 629:130587, 2024.

\bibitem{alperen2021hydrological}
Cagri~Inan Alperen, Guillaume Artigue, Bedri Kurtulus, S{\'e}verin Pistre, and
  Anne Johannet.
\newblock A hydrological digital twin by artificial neural networks for flood
  simulation in gardon de sainte-croix basin, france.
\newblock In {\em IOP Conference Series: Earth and Environmental Science},
  volume 906, page 012112. IOP Publishing, 2021.

\bibitem{lu2021learning}
Lu~Lu, Pengzhan Jin, Guofei Pang, Zhongqiang Zhang, and George~Em Karniadakis.
\newblock Learning nonlinear operators via deeponet based on the universal
  approximation theorem of operators.
\newblock {\em Nature machine intelligence}, 3(3):218--229, 2021.

\bibitem{li2020fourier}
Zongyi Li, Nikola Kovachki, Kamyar Azizzadenesheli, Burigede Liu, Kaushik
  Bhattacharya, Andrew Stuart, and Anima Anandkumar.
\newblock Fourier neural operator for parametric partial differential
  equations.
\newblock {\em arXiv preprint arXiv:2010.08895}, 2020.

\bibitem{wang2024bayesian}
Yuanzhe Wang, Yifei Zong, James~L McCreight, Joseph~D Hughes, and Alexandre~M
  Tartakovsky.
\newblock Bayesian reduced-order deep learning surrogate model for dynamic
  systems described by partial differential equations.
\newblock {\em Computer Methods in Applied Mechanics and Engineering},
  429:117147, 2024.

\bibitem{bhattacharya2020model}
Kaushik Bhattacharya, Bamdad Hosseini, Nikola~B Kovachki, and Andrew~M Stuart.
\newblock Model reduction and neural networks for parametric pdes.
\newblock {\em arXiv preprint arXiv:2005.03180}, 2020.

\bibitem{tang2020deep}
Meng Tang, Yimin Liu, and Louis~J Durlofsky.
\newblock A deep-learning-based surrogate model for data assimilation in
  dynamic subsurface flow problems.
\newblock {\em Journal of Computational Physics}, 413:109456, 2020.

\bibitem{yosinski2014transferable}
Jason Yosinski, Jeff Clune, Yoshua Bengio, and Hod Lipson.
\newblock How transferable are features in deep neural networks?
\newblock {\em Advances in neural information processing systems}, 27, 2014.

\bibitem{liu2021deep}
Xu~Liu, Yingguang Li, Qinglu Meng, and Gengxiang Chen.
\newblock Deep transfer learning for conditional shift in regression.
\newblock {\em Knowledge-Based Systems}, 227:107216, 2021.

\bibitem{TartakovskyJCP2024}
Alexandre~M. Tartakovsky and Yifei Zong.
\newblock Physics-informed machine learning method with space-time
  karhunen-lo{\`e}ve expansions for forward and inverse partial differential
  equations.
\newblock {\em Journal of Computational Physics}, 499:112723, 2024.

\bibitem{chakraborty2021transfer}
Souvik Chakraborty.
\newblock Transfer learning based multi-fidelity physics informed deep neural
  network.
\newblock {\em Journal of Computational Physics}, 426:109942, 2021.

\bibitem{liao2021multi}
Peng Liao, Wei Song, Peng Du, and Hang Zhao.
\newblock Multi-fidelity convolutional neural network surrogate model for
  aerodynamic optimization based on transfer learning.
\newblock {\em Physics of Fluids}, 33(12), 2021.

\bibitem{donahue2014decaf}
Jeff Donahue, Yangqing Jia, Oriol Vinyals, Judy Hoffman, Ning Zhang, Eric
  Tzeng, and Trevor Darrell.
\newblock Decaf: A deep convolutional activation feature for generic visual
  recognition.
\newblock In {\em International conference on machine learning}, pages
  647--655. PMLR, 2014.

\bibitem{feng2019structural}
Chuncheng Feng, Hua Zhang, Shuang Wang, Yonglong Li, Haoran Wang, and Fei Yan.
\newblock Structural damage detection using deep convolutional neural network
  and transfer learning.
\newblock {\em KSCE Journal of Civil Engineering}, 23:4493--4502, 2019.

\bibitem{iraniparast2023surface}
Mostafa Iraniparast, Sajad Ranjbar, Mohammad Rahai, and Fereidoon~Moghadas
  Nejad.
\newblock Surface concrete cracks detection and segmentation using transfer
  learning and multi-resolution image processing.
\newblock In {\em Structures}, volume~54, pages 386--398. Elsevier, 2023.

\bibitem{qiao2018few}
Siyuan Qiao, Chenxi Liu, Wei Shen, and Alan~L Yuille.
\newblock Few-shot image recognition by predicting parameters from activations.
\newblock In {\em Proceedings of the IEEE conference on computer vision and
  pattern recognition}, pages 7229--7238, 2018.

\bibitem{yu2020transmatch}
Zhongjie Yu, Lin Chen, Zhongwei Cheng, and Jiebo Luo.
\newblock Transmatch: A transfer-learning scheme for semi-supervised few-shot
  learning.
\newblock In {\em Proceedings of the IEEE/CVF conference on computer vision and
  pattern recognition}, pages 12856--12864, 2020.

\bibitem{loeve1978probability}
M.~Loeve.
\newblock {\em Probability Theory II}.
\newblock F.W.Gehring P.r.Halmos and C.c.Moore. Springer, 1978.

\bibitem{xiu2010numerical}
Dongbin Xiu.
\newblock {\em Numerical methods for stochastic computations: a spectral method
  approach}.
\newblock Princeton university press, 2010.

\bibitem{Shen2023NatureR}
Chaopeng Shen, Alison~P. Appling, Pierre Gentine, Toshiyuki Bandai, Hoshin
  Gupta, Alexandre Tartakovsky, and ...
\newblock Differentiable modelling to unify machine learning and physical
  models for geosciences.
\newblock {\em Nature Reviews Earth {\&} Environment}, 4(8):552--567, July
  2023.

\bibitem{tartakovsky2021physics}
Alexandre~M Tartakovsky, David~A Barajas-Solano, and Qizhi He.
\newblock Physics-informed machine learning with conditional karhunen-lo{\`e}ve
  expansions.
\newblock {\em Journal of Computational Physics}, 426:109904, 2021.

\bibitem{Yeung2022WRR}
Yu-Hong Yeung, David~A. Barajas-Solano, and Alexandre~M. Tartakovsky.
\newblock Physics-informed machine learning method for large-scale data
  assimilation problems.
\newblock {\em Water Resources Research}, 58(5):e2021WR031023, 2022.

\end{thebibliography}

\appendix

\section{Moment equations for cross-covariances}\label{sec:crosscov}
The equation for $C_{yh}(x',x,t)$ is obtained by multiplying Eqs \eqref{eq:component_model_dev_linear}-\eqref{eq:BC_dev_linear} with $y'(x')$ and taking the ensemble average as 
\begin{equation}
\label{eq:cov_hy}
\mathcal{L}(\overline{h'(x,t)y'(x')}, \overline{y'(x)y'(x')},\overline{f'(x,t)y'(x')}) = 0, \quad x \in \Omega, \quad t\in T,
\end{equation}
subject to the initial condition
\begin{equation}
\overline{h'(x,t=0)y'(x')}= \overline{h'_0(x)y'(x')}
\end{equation}
and the boundary condition  
\begin{equation}
\mathcal{B}\overline{h'(x,t)y'(x')}) = \overline{g'(x,t)y'(x')}, \quad x \in \Gamma.
\end{equation}

The equation for $C_{fh}(x',t',x,t) = C_{hf}(x,t,x',t')$ is obtained by multiplying Eqs \eqref{eq:component_model_dev_linear}-\eqref{eq:BC_dev_linear} with $f'(x',t')$ and taking the ensemble average as 
\begin{equation}
\label{eq:cov_hf}
\mathcal{L}(\overline{h'(x,t)f'(x',t')}, \overline{y'(x)f'(x',t')},\overline{f'(x,t)f'(x',t')}) = 0, \quad x \in \Omega, \quad t\in T,
\end{equation}
subject to the initial condition
\begin{equation}
\overline{h'(x,t=0)f'(x',t')}= \overline{h'_0(x)f'(x',t')}
\end{equation}
and the boundary condition  
\begin{equation}
\mathcal{B}\overline{h'(x,t)f'(x',t')}) = \overline{g'(x,t)f'(x',t')}, \quad x \in \Gamma.
\end{equation}

Finally, the equation for $C_{gh}(x',t',x,t) = C_{hg}(x,t,x',t')$ is obtained by multiplying Eqs \eqref{eq:component_model_dev_linear}-\eqref{eq:BC_dev_linear} with $g'(x',t')$ and taking the ensemble average as 
\begin{equation}
\label{eq:cov_hg}
\mathcal{L}(\overline{h'(x,t)g'(x',t')}, \overline{y'(x)g'(x',t')},\overline{f'(x,t)g'(x',t')}) = 0, \quad x \in \Omega, \quad t\in T,
\end{equation}
subject to the initial condition
\begin{equation}
\overline{h'(x,t=0)f'(x',t')}= \overline{h'_0(x)g'(x',t')}
\end{equation}
and the boundary condition  
\begin{equation}
\mathcal{B}\overline{h'(x,t)g'(x',t')}) = \overline{g'(x,t)g'(x',t')}, \quad x \in \Gamma.
\end{equation}
The cross-covariances between the control variables (e.g., $C_{yf}(x,x',t')$) must be specified to close the system of the moment equations. It is common to assume that the control variables are independent, i.e., to set these cross-covariances to zero. We note here that the covariances of the control variables $C_y$, $C_f$, $C_{h_0}$, and $C_g$ are assumed to be known.

\section{Solution to ordinary least square problems}\label{sec:OLS}
In the OLS approach, $\bm{W}$ in the $\bm\eta = \bm{W} \bm\xi$ surrogate matrix is found to minimize the loss function 
$$
L = \sum_{i=1}^{N_{\text{train}}} || \bm\eta^{(i)} - \bm W\bm\xi^{(i)} ||_2^2 = || \bm H - \bm W \bm{\Xi}||_F^2 ,
$$
where $\bm\Xi = [\bm\xi^{(1)}| \bm\xi^{(2)} |\dots| \bm\xi^{(N_{\text{train}})}] \in \mathbb{R}^{N_\xi \times N_{\text{train}}}$, $\bm H = [\eta^{(1)}| \eta^{(2)}| \dots, \eta^{(N_{\text{train}})}] \in \mathbb{R}^{N_\eta \times N_{\text{train}}}$. Here, the Frobenius norm can be expanded as
\begin{align}
|| \bm{H} - \bm{W} \bm{\Xi}||_F^2 &= \text{Tr}\left((\bm{H} - \bm{W} \bm{\Xi})^\top (\bm{H} - \bm{W} \bm{\Xi})\right) \nonumber \\
& = \text{Tr}(\bm{H}^\top \bm{H}) - 2 \text{Tr}(\bm{H}^\top \bm{W} \bm{\Xi}) + \text{Tr}(\bm{W} \bm{\Xi} \bm{\Xi}^\top \bm{W}^\top). \nonumber
\end{align}
We minimize this norm with respect to $\bm{W}$ by solving the following linear matrix equation:
\begin{equation}\label{eq:derivatives}
\frac{\partial}{\partial \bm{W}} \left(\text{Tr}(\bm{H}^\top \bm{H}) - 2 \text{Tr}(\bm{H}^\top \bm{W} \bm{\Xi}) + \text{Tr}(\bm{W} \bm{\Xi} \bm{\Xi}^\top \bm{W}^\top)\right) = 0.
\end{equation}
Noting that $\frac{\partial}{\partial \bm{W}} \text{Tr}(\bm{A} \bm{W} \bm{B}) =  \bm{A}^\top \bm{B}^\top$ and $\frac{\partial}{\partial \bm{W}} \text{Tr}(\bm{A} \bm{W}^\top) =  \bm{A}$, \eqref{eq:derivatives} can be simplified to
$$ 
-2 \bm{H} \bm{\Xi}^\top + 2 \bm{W} \bm{\Xi} \bm{\Xi}^\top = 0.
$$ 
Solving for $\bm{W}$ as
$$ 
\bm{W} \bm{\Xi} \bm{\Xi}^\top = \bm{H} \bm{\Xi}^\top
$$ 
yields the final solution:
$$ 
\bm{W} = \bm{H} \bm{\Xi}^\top (\bm{\Xi} \bm{\Xi}^\top)^{-1}.
$$ 

\section{Residual least square formulation for the linear diffusion equation}\label{sec:linear_diff}
In the linear diffusion equation \eqref{eq:linear_diffusion}, we represent the state variable $h(x,t)$ and control variables $f(x,t)$ and $q(t)$ with KLDs as
\begin{equation}
  \label{eq:space-time-KL_general-ap}
  h(x,t) \approx  \overline{h}(x,t) + \bm\psi_h(x, t) \cdot \bm\eta,
\end{equation}
\begin{equation}
f(x,t) \approx   \overline{f}(x,t) + \bm\psi_f(x, t) \cdot \bm\xi_f,
\end{equation}
and
\begin{equation}
q(t) \approx   \overline{q}(t) + \bm\psi_q(t) \cdot \bm\xi_q. 
\end{equation}

The control variables in the IBCs \eqref{eq:linear_diffusion_ic} and \eqref{eq:linear_diffusion_bc} are represented as:
\begin{equation}
   h_0 = \overline{h}_0 +  h'_0 , \quad h_l= \overline{h}_l + h'_l, \quad h_r = \overline{h}_r +h'_r. 
\end{equation}

Substituting these expansions in \eqref{eq:linear_diffusion}-\eqref{eq:linear_diffusion_bc} yields the following residuals:

\begin{align}\label{eq:linear_PDE_fluc_appendix}
   R(x,t) = \left( \frac{\partial \bm\psi_h}{\partial t} - 
     \frac{\partial}{\partial x} \left[ k  \frac{\partial \bm\psi_h}{\partial x}   \right] \right)
     \cdot \bm\eta  - \bm\psi_f \cdot \bm\xi_f -  \bm\psi_q \cdot \bm\xi_q \delta(x-x^*),
\end{align}
\begin{equation}\label{eq:linear_PDE_fluc_ic_appendix}
   R_0(x) = \bm\psi_h(x, t = 0) \cdot \bm\eta - h'_0,
\end{equation}
and 
\begin{equation}\label{eq:linear_PDE_fluc_bc_appendix}
   R_{bl}(t) = \bm\psi_h (x = 0,t) \cdot \bm\eta - h'_l \quad\text{and}\quad 
   R_{br}(t) = \bm\psi_h (x = L,t) \cdot \bm\eta - h'_r.
\end{equation}

The residual function $R(x,t)$ can be discretized using any numerical method. Consider a finite difference (FD) discretization on a uniform mesh consisting of $N_x + 2$ grid nodes in space (indexed from $0$ to $N_x + 1$), 
and $N_t + 1$ grid nodes in time (indexed from $0$ to $N_t$). In this setup, the initial condition has a time index of 0, the left boundary condition has a space index of 0, and the right boundary condition has a space index of $N_x + 1$. This results in a total of $N = (N_x + 2) \times (N_t + 1)$ nodes, including $N_m = N_x \times N_t$ internal nodes. The space-time residual vector $\hat{\bm r} \in \mathbb{R}^{N_m}$ evaluated within the computational domain is:    
\begin{align}
    \hat{\bm r} &= \bm A \bm \eta - \bm D^I \bm\psi_f \bm \xi_f - \bm \Delta_{x^*}\bm D^I \bm \psi_q \bm \xi_q  \nonumber \\
    & = \bm A \bm \eta - \begin{bmatrix} 
    \bm D^I \bm \psi_f & \bm\Delta_{x^*} \bm D^I \bm  \psi_q 
    \end{bmatrix} 
    \begin{bmatrix} 
    \bm \xi_f \\ 
    \bm \xi_q 
    \end{bmatrix}  \nonumber \\
    & =  \bm A \bm \eta - \bm B \bm \xi, 
\end{align}
where $\bm \psi_f \in \mathbb{R}^{N \times N_f}$ and $\bm \psi_q \in \mathbb{R}^{N \times N_q}$ are discretized eigenfunctions of $f$ and $q$ evaluated at the entire mesh. The matrix $\bm D_I \in \mathbb{R}^{N_m \times N}$ selects internal nodes. On the $i$-th row of $\bm D_I$, the $j$-th column where $j$ corresponds to the global index of the $i$-th internal node is 1, and other elements are 0. $\bm \Delta_{x^*} \in \mathbb{R}^{N_m \times N_m}$ is a diagonal matrix with the diagonal elements $\Delta_{ii} = 1/\delta_x$ for $i=k i^*_x$ ($k = 1, \dots, N_t$ and $\delta_x$ is the spacial grid size) and zero otherwise, where the $x^*$ coordinate of the source term is assumed to coincide with the space node $i^*_x$. The matrix $\bm A \in \mathbb{R}^{N_m \times N_\eta}$ has elements 
$$
A_{ij} = 
\left(
\frac{\partial \psi_h^j(x,t)}{\partial t} - 
     \frac{\partial}{\partial x} \left[ k(x)  \frac{\partial \psi_h^j(x,t)}{\partial x}   \right]
\right)_{x = x_i,t = t_i}.
$$
%We use the FD scheme from \cite{TartakovskyJCP2024} to discretize derivatives in the $\bm{A}$ matrix.
The residuals $R_0(x)$, $R_{bl}(t)$, and $R_{bl}(t)$ are discretized as  
\begin{align}
    \hat{\bm r}_0 = \bm T \bm\psi_h \bm\eta - \bm h'_0, \quad  \hat{\bm r}_{bl} = \bm X^L \bm\psi_h \bm\eta - \bm h'_l , \quad \hat{\bm r}_{br} = \bm X^R \bm\psi_h \bm\eta - \bm h'_r, 
\end{align}
where vectors $\bm {h'}_{0} \in \mathbb{R}^{N_x}$, $\bm {h'}_{l} \in \mathbb{R}^{N_t + 1}$, $\bm {h'}_{r} \in \mathbb{R}^{N_t + 1}$. Matices $\bm T \in \mathbb{R}^{N_x \times N}$, $\bm X^L \in \mathbb{R}^{(N_t + 1) \times N}$, and $\bm X^R \in \mathbb{R}^{(N_t + 1) \times N}$ are defined as
\begin{align}
%  \bm {h'}_{0, i}  &= \begin{cases} 
% h'_0, \quad \text{if} \quad i = 1, 2, \dots, N_x - 1 \\
% 0, \quad \text{otherwise}.
% \end{cases} \nonumber \\
%  \bm {h'}_{l, i}  &= \begin{cases} 
% h'_l, \quad \text{if} \quad i = 0, N_x, \dots, (N_t - 1)\times N_x, \\
% 0, \quad \text{otherwise}.
% \end{cases} \nonumber \\
%  \bm {h'}_{r, i}  &= \begin{cases} 
% h'_r, \quad \text{if} \quad i = N_x - 1, 2 N_x - 1, \dots, N_t \times N_x - 1, \\
% 0, \quad \text{otherwise}.
% \end{cases} \nonumber \\
 \bm T_{i,j}  &= \begin{cases} 
1, \quad \text{if} \quad j = i + 1, i = 0, \dots, N_x - 1  \\
0, \quad \text{otherwise}.
\end{cases} \nonumber \\
 \bm X^L_{i,j} &= \begin{cases} 
1, \quad \text{if} \quad j = i\times N_x, i = 0, 1, \dots, N_t \\
0, \quad \text{otherwise}.
\end{cases} \nonumber \\
 \bm X^R_{i,j} &= \begin{cases} 
1, \quad \text{if} \quad j = (i + 1)\times N_x + 1, i = 0, 1, \dots, N_t \\
0, \quad \text{otherwise}.
 \end{cases} 
\end{align}
Therefore, $\bm T\bm \psi_h $ corresponds to discretized $\psi_h(x, t = 0)$ (row $1$ to $N_x$ in $\bm \psi_h$), $\bm X^L\bm \psi_h $ corresponds to discretized $\bm\psi_h(x = 0, t)$ (every $N_x$ row of $\bm \psi_h$ starting from row $0$), and $\bm X^R\bm \psi_h $ corresponds to discretized $\bm\psi_h(x = L, t)$ (every $N_x$ row of $\bm \psi_h$ starting from row $N_x + 1$). 
For any $\bm\xi$, $\bm h'_0$, $\bm h'_L$, and $\bm h'_R$, the solution of PDE can be obtained by minimizing the residuals as:
\begin{align}
\bm{\eta}^* &= \min_{\bm{\eta}} \left\{ \omega_r ||\bm{\hat{r}} ||_2^2 + \omega_0 || \hat{\bm{r}}_0 ||_2^2  + \omega_b || \hat{\bm r}_{bl} ||_2^2 + \omega_b || \hat{\bm r}_{br} ||_2^2 \right\} \nonumber \\
&= \min_{\bm{\eta}} \left\{ \omega_r \left\| \bm{A} \bm{\eta} - \bm{B} \bm{\xi} \right\|_2^2 + \omega_0 \left\| \bm{T} \bm{\psi}_h \bm{\eta} - \bm{h}'_0 \right\|_2^2 + \omega_b || \bm X^L \bm\psi_h \bm\eta - \bm h'_L ||_2^2  +  \omega_b || \bm X^R \bm\psi_h \bm\eta - \bm h'_R ||_2^2 \right\} \nonumber \\
&= \min_{\bm{\eta}} \left\{ \left\| \bm{\tilde{A}} \bm{\eta} - \bm{\tilde{b}} \right\|_2^2 \right\}
\end{align}
where
\begin{align}
\bm{\tilde{A}} &= 
\begin{bmatrix}
\Omega_r \bm{A} \\
\Omega_0 \bm{T} \bm{\psi}_h \\
\Omega_b \bm X^L \bm\psi_h \\
\Omega_b \bm X^R \bm\psi_h \\
\end{bmatrix}, \quad
\bm{\tilde{b}} = 
\begin{bmatrix}
\Omega_r \bm{B} \bm{\xi} \\
\Omega_0 \bm{h}'_0 \\
\Omega_b \bm h'_L \\
\Omega_b \bm h'_R
\end{bmatrix}
\end{align}
and 
\begin{align}
\Omega_r = \sqrt{\omega_r} \, \bm{I}, \quad \Omega_0 = \sqrt{\omega_0} \, \bm{I}, \quad \Omega_b = \sqrt{\omega_b} \, \bm{I}.
\end{align}
The solution to this minimization problem has the form
$$  
 \bm\eta^*  = (  \bm{\tilde{A}}^{\text{T}}  \bm{\tilde{A}} )^{-1}  \bm{\tilde{A}}^{\text{T}} \bm{\tilde{b}} = \bm W \tilde{\bm\xi},
$$
where  $\tilde{\bm\xi}^T = [\bm\xi^T, {\bm h'_0}^T, {\bm h'_L}^T, {\bm h'_R}^T]^T$.

\end{document}